\documentclass[twoside,11pt]{article}
\usepackage{jair, theapa, rawfonts}
\usepackage{graphicx}
\usepackage{amsmath}
\usepackage{amssymb}
\usepackage{latexsym}
\usepackage{alltt}
\usepackage{fancyvrb}

\usepackage{soul}
\usepackage{color}

\usepackage{rotating}
\usepackage{multirow}
\usepackage{array}
\newcolumntype{b}{>{\global\let\currentrowstyle\relax}}
\newcolumntype{^}{>{\currentrowstyle}}
\newcommand{\rowstyle}[1]{\gdef\currentrowstyle{#1}%
  #1\ignorespaces
}
\ShortHeadings{Combining Ontologies in $E-\mathcal{SHIQ}$}
{Vouros \& Santipantakis}
\firstpageno{1}

\begin{document}

\title{Combining Ontologies with Correspondences and Link Relations: The $E-\mathcal{SHIQ}$ Representation Framework.}

\author{\name George A. Vouros \email georgev@unipi.gr \\
       \addr Department of  Digital Systems, University of Piraeus,\\
       Karaoli \& Dimitriou 80, Piraeus Greece
       \AND
        \name Georgios Santipantakis \email gsant@aegean.gr \\
       \addr Department of Information and Communications Systems Eng., University of the Aegean,\\
       Karlovassi, Samos, Greece
       }


\maketitle

\begin{abstract}

Combining knowledge and beliefs of autonomous peers in distributed settings, is a major challenge. In this paper we consider peers that combine ontologies and  reason jointly with their coupled knowledge. Ontologies are within the $\mathcal{SHIQ}$ fragment of Description Logics. Although there are several representation frameworks for modular Description Logics, each one makes crucial assumptions concerning the subjectivity of peers' knowledge, the relation between the domains over which ontologies are interpreted,  the expressivity of the constructors used for combining knowledge, and the way peers share their knowledge. However in settings where autonomous peers can evolve and extend their knowledge and beliefs independently from others, these assumptions may not hold. In this article, we motivate the need for a representation framework that allows peers to combine their knowledge in various ways, maintaining the subjectivity of their own knowledge and beliefs, and that reason collaboratively,  constructing a tableau that is  distributed among them, jointly. The paper presents the proposed  $E-\mathcal{SHIQ}$ representation framework, the implementation of the $E-\mathcal{SHIQ}$ distributed tableau reasoner, and discusses the efficiency of this reasoner.

\end{abstract}

\section{Introduction}
\label{Introduction}

To combine knowledge and beliefs of autonomous peers in  open and inherently distributed settings, we need special formalisms that take into account  the complementarity and  heterogeneity of knowledge in multiple interconnected contexts (i.e. local theories). Peers may have different beliefs concerning ``bridging'' heterogeneity and coupling their knowledge with the knowledge of others. The subjectivity of beliefs plays an important role in such a setting, as autonomous peers may inherently (i.e. due to restrictions of their task environment) have different views of the knowledge possessed  by others, or they may not agree on the way they may jointly shape knowledge.
 
The expressivity of knowledge representation frameworks for combining knowledge in multiple contexts, and the efficiency of distributed reasoning processes, depend on the language(s) used for expressing local knowledge and on the language used for connecting different contexts. 

In this paper we consider that an ontology in a setting with multiple, distinct, but connected ontologies defines a local logical theory in a specific context. Connections between ontologies express how knowledge can be combined so as peers to jointly exploit their combined/distributed knowledge during reasoning. As already told, these connections may be subjective, and in the most generic case are not known by all peers.  

The Semantic Web architecture is largely based on new languages, among which the Web Ontology Language (OWL) plays a prominent role. Description Logics have deeply influenced the design and standardization of OWL: OWL-Lite and OWL-DL correspond to logics $\mathcal{SHIF(D)}$ and $\mathcal{SHOIN(D)}$. The proliferation of OWL ontologies, many of which have been  developed independently, the  need for fusing semantically annotated, voluminous, and sometimes  streaming data, using different ontologies, makes the effective combination of knowledge and the effective distributed semantic reasoning an emergent need.  

On the other hand, large ontologies need to be dismantled so as to be evolved, engineered and used effectively during reasoning. The process of taking an ontology to possibly interdependent ontology \textit{units} \cite{surveyMod} is called ontology modularization, and specifically, ontology $partitioning$. Each such unit, a $module$, provides a specific $context$ for performing ontology maintenance, evolution and reasoning tasks, at scales and complexity that are smaller than that of the initial ontology. Therefore, in open and inherently distributed settings (for performing either ontology maintenance, evolution or reasoning tasks), several such $ontology$  $modules$ may co-exist in connection with each other. Formally, any axiom that is expressed using terms in the signature of a module and it is entailed by the ontology must be entailed by the module, and vise-versa. The partitioning task requires that the union of all the modules, together with the set of correspondences/relations between modules, is semantically equivalent to the original ontology. This later property imposes hard restrictions to the modularization task: Indeed, it is hard to maintain it, due to limitations on the expressiveness of the language used for specifying correspondences/relations between modules' elements, due to the local (per ontology module) interpretation of constructs, and due to the restrictions imposed by the setting where modules are deployed. For instance, in case the representation framework used does not support inverse relations between units' entities, then the modularization options are rather limited: Properties that are specified to be inverse in the initial ontology, will either be considered as independent (i.e. non-related) when connecting elements in different modules, or shall be used  to relate elements within specific modules, $only$.  Also, in case modules are deployed in a setting where correspondences between modules are considered to be subjective  (i.e. reflect specific beliefs from the point of view of a specific context), then a subsumption relation between two concepts  in modules  $M_i$ and  $M_j$ may be considered to hold from the point of view of  $M_i$, but not for $M_j$, while the combination of knowledge from different modules must support the proper propagation of subsumption relations across modules. Nevertheless, the knowledge representation framework may impose further restrictions to the domains of distinct units: For instance, specific representation frameworks consider that modules are interpreted over mutually disjoint domains, while other frameworks consider overlapping domains, only. These restrictions affect the expressivity of the constructors used for coupling knowledge in different units, thus, the way knowledge is combined, and also the modularization options supported.

Our goal in this work is to provide a rich representation framework for combining and reasoning with distinct ontologies in open, heterogeneous and inherently distributed settings. In these settings we may expect that different ontologies  may be  combined in many different, subtle ways, by means of correspondences and domain-specific relations between concepts and individuals,  while peers retain subjective beliefs on how their knowledge is coupled with that of others. Our aim is to support peers to  reason jointly with knowledge distributed in their  ontologies, by combining local reasoning chunks. Towards this goal we propose the $E-\mathcal{SHIQ}$ representation framework and a distributed tableau algorithm.  

While standard logics may be used to deal with the issues of heterogeneity in the semantic web, special knowledge representation formalisms have been proposed, sometimes called \emph{contextual logics} or \emph{modular ontology languages}.  While these formalisms are presented and discussed in detail in the next sections, the paragraphs that follow introduce them shortly, pointing out their advantages and limitations, and stating the features of $E-\mathcal{SHIQ}$.

Among these languages, $\mathcal{E}$-Connections provide a uniform framework for combining Abstract Description Systems (ADSs) \cite{KutzLWZ04},  a generalization of several families of decidable logics, including Description Logics. An $\mathcal{E}$-Connection language is determined by a set of logics to be combined and a set of constructors that provide the coupling between them. $\mathcal{E}$-Connections provide useful expressivity while assuring transfer of decidability: If reasoning is decidable in each of the logics in combination, then it is decidable in the combined formalism as well. The combination of Description Logics knowledge bases using $\mathcal{E}$-Connections has been proposed as a suitable technique for combining knowledge from distinct ontologies interpreted over \textit{disjoint logical domains} on the Semantic Web \cite{GrauIJCAIWorkshop} \cite{GrauPS2004}, \cite{ParsiaG05}. The component ontologies are connected by means of \textit{link relations}. The constructors provided by the $\mathcal{E}$-Connection language are associated to link relations and are used to describe the relationships between the connected ontologies. Reasoning services for $\mathcal{E}$-Connected ontologies can be provided only by a centralized reasoning engine that receives as input the connected Knowledge Base. 

While $\mathcal{E}$-Connections have been conceived for linking distinct ontologies interpreted over disjoint logical domains by means of link relations, Distributed Descriptions Logics \cite{serafini_borgida_tamilin05dist_modular_reasoning} \cite{BorgidaSerafini03} is a formalism for combining distinct Description Logics knowledge bases by means of constructors specifying inter-ontology concept-to-concept correspondences, called \textit{bridge rules}. These rules establish directional (subjective) subsumption relationships between concepts. An effort towards extending OWL with a suitable means for specifying such inter-ontology mappings, adding the DDL to the language, is C-OWL \cite{conf/semweb/BouquetGHSS03}. While DDL is supported by the DRAGO reasoning system \cite{SerafiniT05} operating in a distributed way in a peer-to-peer setting, no reasoning support is provided for C-OWL. Actually DRAGO does not construct a true distributed tableau that corresponds to a true distributed model. Nevertheless, DDL, as well as C-OWL, has been criticized that it does not model certain crucial properties of subsumption relations (mainly, due to the granularity of correspondences between individuals in different domains). This, in combination to the fact that -according to the original DDL semantics- subjective subsumption relations do not propagate transitively, motivated a  tableau algorithm for restricted DDL, where subsumption propagates between remote ontologies \cite{Homola:2010:ASP:1744779.1744782} \cite{CRPITV90P21-30}. 

In Package-based Description Logics (P-DL) \cite{baojie-crr2006} \cite{BaoVSH09} \cite{Bao2006} a distributed (packaged) ontology is composed of a collection of ontologies called packages. Each term (name of a concept, property or individual) is associated with a home package, while a package can use terms defined in other packages by importing these terms, and their related axioms.  Reusing terms supports modeling inter-package concept subsumption and roles. P-DL supports distributed reasoning with these inter-module constructors, allowing arbitrary references of concepts among ontology modules and the combination of local reasoning chunks in a synchronous peer-to-peer  fashion. P-DL was proposed to compensate with  limitations of the \emph{owl:imports} construct defined in OWL, which is unsatisfactory given that imported axioms do not retain their context.  However, P-DL treat correspondences between distinct ontologies in a rather objective way, allowing inferring semantic relations between entities in two ontologies, even if none of these ontologies imports terms directly from the other. 

Integrated Distributed Description Logics \cite{Zimmermann2008} takes a different paradigm than other contextual frameworks: Usually, cross-ontology assertions (e.g., bridge rules in DDL, links in $\mathcal{E}$-connections, semantic imports in P-DL) define knowledge from the point of view
of one ontology.  On the contrary, IDDL asserts correspondences
from an ÒexternalÓ point of view which encompasses both ontologies in relation.
One consequence of this approach is that correspondences can be manipulated and
reasoned about independently of the ontologies, allowing operations like inversing
or composing ontology alignments, as first class objects.
A reasoning procedure for this formalism has been defined, where a central
system detaining the correspondences can determine global consistency of a network
of ontologies by communicating with local reasoners of arbitrary complexity.
The motivation for separating local and global
reasoning  is to better prevent interactions between contexts, thus making IDDL quite robust
to heterogeneity. 

As already said, this paper describes work that has  been motivated towards proposing a representation framework for combining and reasoning with multiple ontologies in open and heterogeneous settings.   Towards this target, the representation framework $E _{HQ^+}$ $\mathcal{SHIQ}$ (or simply $E-\mathcal{SHIQ}$)  (a) provides constructors associated to link relations and  inter-ontologies' concept-to-concept correspondences, retaining the subjectiveness of specifications, (b) offers expressiveness for combining knowledge in distinct ontologies interpreted over overlapping or disjoint domains, and (c) supports distributed reasoning by combining local reasoning chunks in a  peer-to-peer fashion, inherently supporting subsumption propagation between ontologies. Each reasoning peer with a specific ontology unit holds a part of a distributed tableau, which corresponds to a distributed model.

Correspondences as well as link relations in $E-\mathcal{SHIQ}$ are treated as first-class objects that can be exploited for further reasoning, or for specifying specific information about them.  

This article specifies the representation framework and provides details on the implementation of the distributed reasoner, which is realized as an extension of the Pellet reasoner\footnote{http://clarkparsia.com/pellet/}. The  article provides experimental results using multiple fragments from a specific ontology that is in a highly expressive fragment of  $\mathcal{SHIQ}$. The results show that the reasoning mechanism achieves better performance than its centralized counterpart, although further optimizations may be applied. 

Section 2 provides background knowledge and provides details on combining knowledge using modular ontology representation frameworks. Section 3 provides motivating examples and shows the limitations of existing frameworks, as well as the specific contributions made by $E-\mathcal{SHIQ}$. Section 4 specifies $E-\mathcal{SHIQ}$ and section 5 presents a tableau algorithm for reasoning with multiple ontologies combined via $E-\mathcal{SHIQ}$. Section 6 provides detailed information on the implementation of the $E-\mathcal{SHIQ}$ distributed reasoner and section 7 provides experimental results for reasoning with this reasoner. Finally, section 8 thoroughly discusses the potential of the $E-\mathcal{SHIQ}$ representation framework to serve as a new paradigm for combining multiple ontologies and reasoning with large ontologies, presents future research plans and concludes the paper.

\section{Preliminaries}
\label{preliminaries}

 This section provides preliminary knowledge, also introducing terminology and notation for the next sections. This is necessary to discuss and motivate the choices made, and to present the building blocks of the proposed framework. 

We assume that the language that different ontologies use for their local specifications are at most as expressive as the $\mathcal{SHIQ}$ fragment of DL.  $\mathcal{SHIQ}$ includes  a rich set of constructors shared between different contextual representation frameworks, it is widely used for the development of ontologies, also ensuring decidability of the reasoning procedure, under certain conditions. 

Thus, given a non-empty set of indices $I$, we assume a collection of independent ontology units\footnote{Subsequently we use the terms ontology unit, unit and ontology interchangeably.} indexed using $I$. Each of these units is within a fragment of Description Logics, whose expressivity is at most equivalent to $\mathcal{SHIQ}$.  Given a unit $i \in I$, let $N_{C_i}$, $N_{R_i}$ and $N_{O_i}$ be the mutually disjoint sets of concept, role and individual names respectively. For some $R \in N_{R_i}$, $Inv(R)$ denotes the inverse role of $R$ and ($N_{R_i} \cup \{Inv(R) | R \in N_{R_i}\}$) is the set of $\mathcal{SHIQ}$-roles for the $i-th$ ontology. The set of $\mathcal{SHIQ}$-concepts  is the smallest set constructed by the constructors listed in Table $\ref{shiq}$.

\begin{table*}
\centering
\caption{$\mathcal{SHIQ}$ fragment of Description Logics}
\resizebox{1.0\textwidth}{!}{
\begin{tabular}{|c|c|c|} \hline
Atomic Concept & $C^{\mathcal{I}} \subseteq \Delta$ & \\ 
Universal Concept & $\top^{\mathcal{I}} = \Delta$ & \\ 
Bottom Concept & $\bot^{\mathcal{I}}=\emptyset$ & \\
Atomic Role & $R^{\mathcal{I}} \subseteq \Delta \times \Delta$ & \\
Conjunction & $(C \sqcap D)^{\mathcal{I}}=C^{\mathcal{I}} \cap D^{\mathcal{I}}$ & $\mathcal{S}$\\
Disjunction & $(C \sqcup D)^{\mathcal{I}}=C^{\mathcal{I}} \cup D^{\mathcal{I}}$ & \\
Negation & $(\neg C)^{\mathcal{I}}=\Delta \setminus C^{\mathcal{I}} $ & \\
Existential Restriction & $(\exists R.C)^{\mathcal{I}}=\{x \in \Delta|\exists y \in \Delta, (x,y)\in R^{\mathcal{I}}, y \in C^{\mathcal{I}}\}$ & \\
Value Restriction & $(\forall R.C)^{\mathcal{I}} = \{x\in \Delta | \forall y \in \Delta, (x,y) \in R^{\mathcal{I}} \to y \in C^{\mathcal{I}}\}$ & \\
Transitive Role & $\mathcal{I} \models Trans(R) \leftrightarrow R^{\mathcal{I}}=(R^{\mathcal{I}})^+$ & \\ \hline
Role Hierarchy & $\mathcal{I} \models (P \sqsubseteq R)^\mathcal{I} \leftrightarrow P^{\mathcal{I}} \subseteq R^{\mathcal{I}}$ & $\mathcal{H}$\\ \hline
Inverse Role& $(Inv(R))^{\mathcal{I}} = \{(x,y)|(y,x)\in R^{\mathcal{I}}\}$ & $\mathcal{I}$\\ \hline
Qualified& $(\geq nS.C)^{\mathcal{I}}=\{x \in \Delta,||y, (x,y)\in S^{\mathcal{I}} \wedge y \in C^{\mathcal{I}}||\geq n\}$ & $\mathcal{Q}$\\ 
Number Restrictions& $(\leq nS.C)^{\mathcal{I}}=\{x \in \Delta,||y, (x,y)\in S^{\mathcal{I}} \wedge y \in C^{\mathcal{I}}||\leq n\}$& \\
\hline
\end{tabular}
}\label{shiq}
\end{table*}

A \textit{role box} $\mathcal{R}_i$ is a finite set of role inclusion axioms of the form $R \sqsubseteq S$ (i.e. a role hierarchy), where $R$ and $S$ are in  $N_{R_i}$, and transitivity axioms for roles in  $N_{R_i}$.

Cardinality restrictions can be applied on $R$, given that $R$ is a \textit{simple role}, i.e. a role whose sub-roles w.r.t. the transitive-reflexive closure of the role inclusion relation (denoted by $\sqsubseteq^*_{\mathcal{R}_i}$) are not transitive.

Let $C$ and $D$ possibly complex concepts. $C \sqsubseteq D$ is called a \textit{general concept inclusion} (GCI) axiom. A finite set of GCIs is called a TBox (denoted by $\mathcal{T}_i$). 

The ABox of an ontology unit $i$, denoted as $\mathcal{A}_i$,  is a finite set of  assertions of the form $a:C$, $(a,b):R$, or $a \neq b$, for $a,b \in N_{O_i}$, a possibly inverse role $R$ and a  $\mathcal{SHIQ}$ concept $C$. 

In a distributed setting with multiple ontology units, descriptions and axioms are made distinct by the index of the ontology unit to which they belong, which is used as a prefix: e.g. \textit{i:C}
denotes that the concept $C$ belongs to the $i$-th ontology unit.

An \textit{interpretation} for the $i-th$ ontology $\mathcal{I}_i=\langle {\Delta_i},\cdot^{\mathcal{I}_i}\rangle$ consists of a domain ${\Delta_i}\neq \emptyset$ and the interpretation function $\cdot^{\mathcal{I}_i}$ which maps every $C\in N_{C_i}$ to $C^{\mathcal{I}_i}\subseteq {\Delta_i}$, every $R\in N_{R_i}$ to $R^{\mathcal{I}_i} \subseteq {\Delta_i} \times {\Delta_i}$ and each individual $a \in N_{O_i}$ to an element $a^{\mathcal{I}_i} \in {\Delta_i}$.

An interpretation $\mathcal{I}_i$ \textit{satisfies} a role hierarchy if for any ($R \sqsubseteq S) \in \mathcal{R}_i$ it holds that $R^{\mathcal{I}_i} \subseteq S^{\mathcal{I}_i}$. An interpretation that satisfies all inclusion axioms (the role hierarchy) and all transitivity axioms in $\mathcal{R}_i$, is called a \textit{model} of $\mathcal{R}_i$ . 

An interpretation $\mathcal{I}_i$ satisfies a GCI $C \sqsubseteq D$ if $C^{\mathcal{I}_i} \subseteq D^{\mathcal{I}_i}$. $\mathcal{I}_i$ satisfies a TBox $\mathcal{T}_i$, if it satisfies each GCI in it. In this case $\mathcal{I}_i$ is a $model$ of this TBox. A concept $C$ is \textit{satisfiable} w.r.t. an RBox $\mathcal{R}_i$  and a TBox $\mathcal{T}_i$ if  there is a model $\mathcal{I}_i$ of $\mathcal{T}_i$ and  $\mathcal{R}_i$  with $C^{\mathcal{I}_i} \neq \emptyset $. A concept $D$ subsumes a concept $C$ w.r.t  $\mathcal{T}_i$ and $\mathcal{R}_i$  if $C^{\mathcal{I}_i} \subseteq D^{\mathcal{I}_i}$ holds in every model $\mathcal{I}_i$ of $\mathcal{T}_i$ and $\mathcal{R}_i$ .

An interpretation $\mathcal{I}_i$ maps each individual $a \in N_{O_i}$ to some element $a^{\mathcal{I}_i} \in \Delta_i$. An interpretation $\mathcal{I}_i$ satisfies each a concept assertion $a:C$ iff $a^{\mathcal{I}_i} \in C^{\mathcal{I}_i}$, a role assertion $(a,b):R$ iff $(a^{\mathcal{I}_i},b^{\mathcal{I}_i}) \in R^{\mathcal{I}_i}$ and an inequality $a \overset{.}{\neq} b$ iff $a^{\mathcal{I}_i} \neq b^{\mathcal{I}_i}$. An $ABox$ $\mathcal{A}_i$ is consistent w.r.t. $\mathcal{R}_i$ and $\mathcal{T}_i$ iff there is a model $\mathcal{I}_i$ of $\mathcal{R}_i$ and $\mathcal{T}_i$ that satisfies each assertion in $\mathcal{A}_i$.


\subsection{DDL: Bridge Rules and Individual Correspondences}\label{DDL}
Given the non-empty set of indices $I$ and a collection of ontology units in Description Logics whose expressivity is at most equivalent to $\mathcal{SHIQ}$, a subjective \emph{concept-to-concept bridge rule} from unit $i$ to unit $j$, from the  point of view of $j$ can be (a)  a concept $onto$ concept rule: \textit{i:C} $\overset{\sqsupseteq}{\to}$ \textit{j:G}, or (b)
a concept $into$  concept rule: \textit{i:C} $\overset{\sqsubseteq}{\to}$ \textit{j:G}, where $C \in N_{C_i}$ and $G \in N_{C_j} $\footnote{Other types of mappings between ontology elements  \cite{GhidiniS06},\cite{GhidiniST07} are beyond the scope of this work.}. 

A DDL distributed TBox is a $\mathfrak{T}=\langle \textbf{T},\mathfrak{B} \rangle$, where \textbf{T}=$\{(\mathcal{T}_i,\mathcal{R}_i)\}_{i \in I}$ is a collection of $\mathcal{SHIQ}$ TBox'es ${\mathcal{T}_i}$ and RBox'es ${\mathcal{R}_i}$, and $\mathfrak{B}$=$\{\mathfrak{B}_{ij}\}_{i\neq j \in I}$ is a collection of  bridge rules from ${\mathcal{T}_i}$ to ${\mathcal{T}_j}$. Each ${\mathcal{T}_i}$ is a collection of general inclusion axioms over $N_{C_i}$ and ${\mathcal{R}_i}$ is a collection of role inclusion axioms over $N_{R_i}$.

A DDL distributed ABox $\mathfrak{A}$ includes local ABox'es and subjective individual correspondences $\mathfrak{C}_{ij}$: Given an instance name \textit{i:a} in a local ABox $\mathcal{A}_{i}$ and $\textit{j:b} ^1 , \textit{j:b}^2 , ..., \textit{j:b}^n$ instances names in the ABox $\mathcal{A}_j$, a \textit{partial individual correspondence} is an expression of the form $\textit{i:a} \mapsto \textit{j:b}^k, k=1,...,n$, while a \textit{complete individual correspondence} is an expression of the form $\textit{i:a} \overset {=} \mapsto \{\textit{j:b} ^1 , ..., \textit{j:b} ^n \}$. These individual correspondences are specified to be from the subjective point of view of $j$. These types of correspondences allow bridging ontologies where knowledge is specified at different levels of granularity. However, the generality these  correspondences offer is not without a price: As already pointed out, under the original DDL semantics, the subsumption relation is not  propagated among distant ontologies properly. This issue is further discussed in section 3.

A distributed interpretation $\mathfrak{I}=\langle \{\mathcal{I}_i\}_{i \in I}, \{r_{ij}\}_{i \neq j \in I} \rangle$ of a distributed TBox $\mathfrak{T}$ consists of local interpretations $\mathcal{I}_i$ for each unit $i$ on local domains $\Delta_i$, and a family of domain relations $\{r_{ij}\}_{i,j \in I, i\neq j}$ between these domains. A domain relation $r_{ij}$ from $\Delta_i$ to $\Delta_j$ is a subset of $\Delta_i \times \Delta_j$ and specifies corresponding sets of individuals in the domains of $i$ and $j$ units, from the subjective point of view of $j$. 

DDL introduce non-classical \emph{full hole} (or \emph{empty hole}) interpretations, to prohibit the propagation of inconsistencies between ontology units \cite{serafini_borgida_tamilin05dist_modular_reasoning}. A full (or empty) hole interpretation is an interpretation $\mathcal{I}^\delta = \langle \Delta, \cdotp ^{\mathcal{I}^\delta} \rangle$ (respectively, $\mathcal{I}^\epsilon = \langle \Delta, \cdotp ^{\mathcal{I}^\epsilon} \rangle$), where the function $\cdotp^{\mathcal{I}^\delta}$ maps every concept in a local TBox - including $\bot$ and $\top$ - to the domain $\Delta$ (for empty holes,  the function $\cdotp^{\mathcal{I}^\epsilon}$ maps every concept in a local TBox - including $\bot$ and $\top$ - to the $\emptyset$).

A distributed interpretation $\mathfrak{I}$ satisfies the elements of a distributed TBox $\mathfrak{T}$
 (denoted by $\mathfrak{I} \models_{d} $) according to the following clauses:
 \begin{itemize}
 \item $\mathfrak{I} \models_{d} (\mathcal{T}_i, \mathcal{R}_i)$ if  $\mathcal{I}_i  \models \mathcal{T}_i$ and $\mathcal{I}_i  \models \mathcal{R}_i$
\item $\mathfrak{I} \models_{d} i:C \overset{\sqsubseteq} \rightarrow j: D$ if $r_{ij}(C^{\mathcal{I}_{i}}) \subseteq D^{\mathcal{I}_{j}}$
 \item $\mathfrak{I} \models_{d} i:C \overset{\sqsupseteq} \rightarrow j: D$ if $r_{ij}(C^{\mathcal{I}_{i}}) \supseteq D^{\mathcal{I}_{j}}$
 \item $\mathfrak{I} \models_{d} \mathfrak{B}$ if $\mathfrak{I}$ satisfies all bridge rules in $\mathfrak{B}$ (i.e. in any $\mathfrak{B}_{ij}$, $i \neq j \in I$)
 \item $\mathfrak{I} \models_{d} \mathfrak{T}$ (i.e. $\mathfrak{I}$ is a model of $\mathfrak{T}$) if $\mathfrak{I} \models_{d} \mathfrak{B}$, $\mathfrak{I} \models_{d} \mathcal{T}_i$, and $\mathfrak{I} \models_{d} \mathcal{R}_i$ for each $i \in I$.

 \end{itemize}

Concerning the distributed ABox $\mathfrak{A}$, a distributed interpretation $\mathfrak{I}$, satisfies the elements of $\mathfrak{A}$ if $\mathcal{I}_i \vDash a:C$, $\mathcal{I}_i \vDash (a,b):R$, for all assertions $a:C$, $(a,b):R$ in $\mathcal{A}_i$ and $\mathfrak{I} \vDash _d i:x \overset{=}{\mapsto} j:y$, iff $y^{\mathcal{I}_j} \in r_{ij}(x^{\mathcal{I}_i})$. The distributed interpretation $\mathfrak{I}$ satisfies the distributed ABox $\mathfrak{A}$ if for every $i,j \in I$, $\mathfrak{I} \vDash _d \mathcal{A}_i$ and $\mathfrak{I} \vDash _d \mathfrak{C}_{ij}$.
 



\subsection{  $\mathcal{E}-$connections: i-Concepts and Link Relations}\label{e-connx}

$\mathcal{E}$-connections combine distinct ontology units in any of the  $\mathcal{SHIQ}$, $\mathcal{SHOQ}$, $\mathcal{SHIO}$ fragments of Description Logics, via link relations, assuming a set of  constructors and properties for these relations. In this article we consider only ontologies in the  $\mathcal{SHIQ}$ fragment of Description Logics and consider hierarchically related, transitive link relations. Link relations, as roles, have to be simple if they are restricted with qualified cardinality restrictions. Simplicity of link relations is defined as for $\mathcal{SHIQ}$ roles. 

Link relations connect individuals in the denotation of concepts in different ontology units. 

Given a finite index set $I$ and $i \in I$, the set of $i-roles$ is the set of $\mathcal{SHIQ}$ roles for the $i-th$ ontology. An $i-role$ $axiom$ is  a role inclusion axiom of i-roles $R$ and $S$. 

A \emph{combined role box} is a tuple $\mathcal{R}=(\mathcal{R}_i)_{i \in I}$, where  $\mathcal{R}_i$ is the role hierarchy of the $i-th$ ontology. 

The set of \emph{ij-link relations} relating individuals in the $i$ and $j$ units, $i \neq j \in I$, is denoted by $\mathcal{E}_{ij}$. These sets are not pairwise disjoint, but are disjoint with respect to the sets of concept names.  

An $\textit{ij-}link$ $relation$ $axiom$ is an expression of the form $E^{n}_{ij} \sqsubseteq E^{m}_{ij}$, where the superscript distinguishes link-properties in $\mathcal{E}_{ij}$. An $\textit{ij-}relation$ $box$ 
$\mathcal{R}_{ij}$ includes a finite set of $\textit{ij-}relation$ inclusion axioms.

The combined LBox $\mathcal{L}$ contains all $\mathcal{R}_{ij}, i,j \in I$, and, as it is specified in \cite{ParsiaG05}, and a set of \textit{generalized transitive axioms}. Generalized transitive axioms are of the form $Trans(E;(i_1,i_2),...,(i_{n-1},i_n))$, where $E$ is a link-relation name defined for each pair of ontology units in $\{(i_1,i_2),...,(i_{n-1},i_n)\} \subseteq I \times I$. 

Let $(N_{C_i})_{i \in I}$ be tuples of sets of concept names. The sets of \textit{i-concepts} (i.e. concepts specified in the  $i$-th unit) are inductively defined as the smallest sets constructed by the constructors shown in table \ref{econn_constructors}, where $C$ and $D$ are i-concepts,  $G$ is a j-concept, $E_{ij}$ is an ij-link relation with $i \neq j$ and $R$ is an i-role.

\begin{table}
\centering
\caption{$\mathcal{E}$-connections $i-concepts$}
\resizebox{0.8\textwidth}{!}{
\begin{tabular}{|c|c|} \hline
$C$&$ C^{\mathcal{I}_i}\subseteq \Delta_i$\\ 
$\top_i$&$(\top_i)^{\mathcal{I}_i}= \Delta_i$\\ 
$\bot_i$&$(\bot_i)^{\mathcal{I}_i}= \emptyset$\\ 
$$&$R^{\mathcal{I}_{i}} \subseteq \Delta_i \times \Delta_i$, $(E_{ij})^{\mathcal{I}_{ij}} \subseteq \Delta_i \times \Delta_j$\\ 
$$&$(Inv(R))^{\mathcal{I}_{i}}=\{(x,y) | (y,x) \in R^{\mathcal{I}_{i}} \}$\\
$\neg C$&$(\neg C)^{\mathcal{I}_{i}}=\Delta_i\setminus C^{\mathcal{I}_{i}}$\\
$C \sqcap D$&$(C \sqcap D)^{\mathcal{I}_i}=C^{\mathcal{I}_i} \cap D^{\mathcal{I}_i}$\\
$(C \sqcup D)$&$(C \sqcup D)^{\mathcal{I}_i}=C^{\mathcal{I}_i} \cup D^{\mathcal{I}_i}$\\
$(\exists R.C)$ & $(\exists R.C)^{\mathcal{I}_i}=\{x \in \Delta_i|\exists y \in \Delta_i, (x,y)\in R^{\mathcal{I}_i}, y \in C^{\mathcal{I}_i}\}$\\
$(\forall R.C)$ & $(\forall R.C)^{\mathcal{I}_i} = \{x\in \Delta_i | \forall y \in \Delta_i, (x,y) \in R^{\mathcal{I}_i} \to y \in C^{\mathcal{I}_i}\}$ \\
$(\geq n R.C)$&$(\geq n R.C)^{\mathcal{I}_i}=\{x \in \Delta_i,||y\in \Delta_i, (x,y)\in R^{\mathcal{I}_{i}} \wedge y \in C^{\mathcal{I}_i}||\geq n\}$\\ 
$(\leq n R.C)$&$(\leq n R.C)^{\mathcal{I}_i}=\{x \in \Delta_i,||y\in \Delta_i, (x,y)\in R^{\mathcal{I}_{i}} \wedge y \in C^{\mathcal{I}_i}||\leq n\}$\\ 
$(\exists E_{ij}.G)$&$(\exists E_{ij}.G)^{\mathcal{I}_i}=\{x \in \Delta_i|\exists y \in \Delta_j, (x,y)\in E_{ij}^{\mathcal{I}_{ij}}, y \in G^{\mathcal{I}_j}\}$\\
$(\forall E_{ij}.G)$&$(\forall E_{ij}.G)^{\mathcal{I}_i} = \{x\in \Delta_i | \forall y \in \Delta_j, (x,y) \in E_{ij}^{\mathcal{I}_{ij}} \to y \in G^{\mathcal{I}_j}\}$\\
$(\geq n E_{ij}.G)$&$(\geq n E_{ij}.G)^{\mathcal{I}_i}=\{x \in \Delta_i,||y\in \Delta_j, (x,y)\in E_{ij}^{\mathcal{I}_{ij}} \wedge y \in G^{\mathcal{I}_j}||\geq n\}$\\ 
$(\leq n E_{ij}.G)$&$(\leq n E_{ij}.G)^{\mathcal{I}_i}=\{x \in \Delta_i,||y\in \Delta_j, (x,y)\in E_{ij}^{\mathcal{I}_{ij}} \wedge y \in G^{\mathcal{I}_j}||\leq n\}$\\
\hline
\end{tabular}}
\label{econn_constructors}
\end{table}

 The $\mathcal{E}-$connections definition of a \emph{combined TBox} is a family of TBoxes  $\textbf{T}=\{\mathcal{T}_i\}_{i \in \text{I}}$, where $\mathcal{T}_i$ is a finite set of i-concept inclusion axioms (inclusion axioms between i-concepts). 
A \emph{combined knowledge base} $\Sigma=\langle \textbf{T}, \mathcal{R} , \mathcal{L}\rangle$ is composed by the combined TBox $\textbf{T}$, combined RBox $ \mathcal{R}$, and combined LBox $ \mathcal{L}$. 

Considering different ontology units being connected by link relations only, an interpretation is a structure of the form 
$\mathfrak{I}=\langle (\Delta_i), (\cdotp^{\mathcal{I}_i}), (\cdotp^{\mathcal{I}_{ij}})_{i \neq j} \rangle$,
 where $i, j \in I$, $\Delta_i$ is the i-th non-empty interpretation domain and $\Delta_i \cap \Delta_j = \emptyset$, for $i \neq j$. The  valuation function $\cdotp^{\mathcal{I}_i}$ maps every concept to a subset of $\Delta_i$ and every i-role to a subset of $\Delta_i \times \Delta_i$, while the function  $\cdotp^{\mathcal{I}_{ij}}$ maps every ij-link relation to a subset of $\Delta_i \times \Delta_j$. 

If $C$ and $D$ are i-concepts, the interpretation satisfies the axiom $C \sqsubseteq D$ if $C^{\mathcal{I}_i} \subseteq D^{\mathcal{I}_i}$, and satisfies the combined TBox if it satisfies all axioms in all the component sets. Finally, the interpretation $\mathfrak{I}$ satisfies the combined knowledge base $\Sigma=\langle \textbf{T}, \mathcal{R} , \mathcal{L}\rangle$, denoted by  $\mathfrak{I} \models \Sigma$ iff it satisfies $\textbf{T}, \mathcal{R}$ and $\mathcal{L}$.

\subsection{P-DL: Importing terms and packaging}\label{PDL}

 In P-DL a distributed (packaged) ontology is composed of $packages$. Each package is an ontology unit (i.e. an ontology) that can use terms (concepts, properties, individuals) defined in other packages via $importing$. Each term or axiom is associated with a home package.
 Given a finite index set $I$, each package $P_i$ can import any term $t$ defined in any package $P_j$, with $i \neq j$. This is denoted by $P_j \overset{t} \rightarrow P_i$. 
 
 A packaged ontology $\langle \{P_i\}, \{P_i \rightarrow P_j\}_{i \neq j}\rangle$, has a distributed model $M=\langle \{\mathcal{I}_i\}, \{r_{ij}\}_{i \neq j}\rangle$, where $\mathcal{I}_i=(\Delta_i, \cdotp^{\mathcal{I}_i})$ is the local model for each package $P_i$, and $r_{ij} \subseteq \Delta_i \times \Delta_j$ is the interpretation of the \textit{image domain relation} $\{P_i \rightarrow P_j\}_{i \neq j \in I}$. Specifically, $(x,y) \in r_{ij}$ indicates that $y \in \Delta_j$ is an $image$ of $x \in \Delta_i$.
 
 To assure package transitive reusability and correctness in reasoning, P-DL require that image domain relations are \textit{one-to-one} and that are \textit{compositional consistent}, i.e.  $r_{ij} = r_{ik} \circ r_{kj}$, where $\circ$ denotes relation composition. This means that semantic relations between two terms in two different packages can be inferred even if none of these packages imports terms  directly  from the other package. This imposes a kind of ``objectiveness'' to image domain relations (in contrast to subjective domain relations in DDL). This is a strong assumption, since, as already conjectured, in any inherently distributed and heterogeneous setting domain relations may be subjective for the  peers using these packages.
 
A concept $\textit{i:C}$ is satisfiable w.r.t. a P-DL $\langle \{P_i\}, \{P_j \rightarrow P_i\}_{i \neq j}\rangle$ if there exists a distributed model such that $\textit{i:C}^{\mathcal{I}_i} \neq \emptyset$. Also, the packaged ontology entails the subsumption $\textit{i:C} \sqsubseteq \textit{j:D}$ iff $r_{ij}(C^{\mathcal{I}_i}) \subseteq D^{\mathcal{I}_j}$ holds in every distributed model of the packaged ontology.

\section{Motivating  Examples}
\label{motivatingexamples}

This section provides concrete examples from the different representation frameworks for combining knowledge and reasoning with Description Logics in distributed settings. The aim is to show the specific limitations of these frameworks, motivate the proposed $E-\mathcal{SHIQ}$ framework and point out the specific contributions made by $E-\mathcal{SHIQ}$.

Let us consider a set of indices $I$ and, initially, two ontologies, shown in figure 1: The first specifies knowledge about articles in conference proceedings, and the second about conferences. Considering that these ontologies are interpreted over disjoint logical domains,  $\mathcal{E}$-connections is suitable for combining knowledge from these ontologies. Specifically, we  can consider the combined knowledge base $\Sigma=\langle \textbf{T}, \mathcal{R} , \mathcal{L}\rangle$, with $\textbf{T}=\{\mathcal{T}_1,\mathcal{T}_2\}, \mathcal{R}=(\mathcal{R}_1,\mathcal{R}_2)$ and $\mathcal{L}$ containing $\mathcal{R}_{12},\mathcal{R}_{21}$ and the set $Trans$ of transitivity axioms for link relations and roles. 

\begin{figure}[!htb]
\centering
\includegraphics[scale=.5]{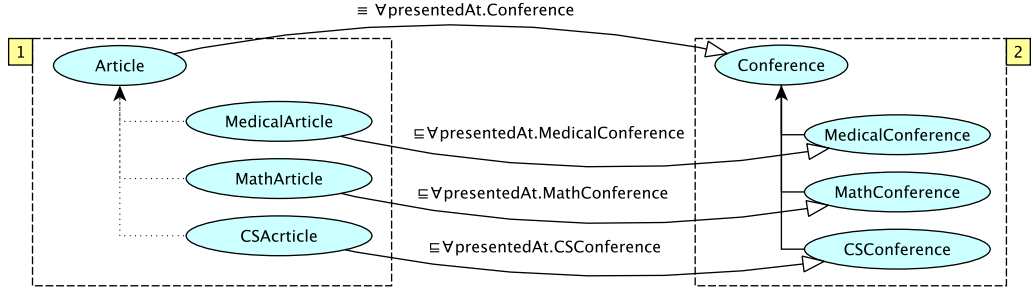}
\caption{Combination of knowledge using link relations. Dotted lines show inclusion relations that are implied from specifications.}
\label{fig1}
\end{figure}

As it is also shown in Figure 1, 
\begin{align}
\mathcal{T}_1= \{ 1:Article & \equiv (\forall presentedAt.2:Conference), \\ 1:MedicalArticle & \sqsubseteq (\forall presentedAt.2:MedicalConference), \\ 1:MathArticle & \sqsubseteq (\forall presentedAt.2:MathConference),\\  1:CSArticle & \sqsubseteq (\forall presentedAt.2:CSConference)\}, 
\\\mathcal{T}_2= \{MedicalConference & \sqsubseteq Conference,\\ MathConference & \sqsubseteq Conference, \\ CSConference & \sqsubseteq Conference\}. 
\end{align}

The property $presentedAt$ is a 12-link relation and  $\mathcal{R}_1=\mathcal{R}_2=\mathcal{R}_{12}=\mathcal{R}_{21}=Trans=\emptyset$.

These specifications imply that $MedicalArticle \sqsubseteq Article$, $MathArticle \sqsubseteq Article$ and $CSArticle \sqsubseteq Article$. 

\begin{figure}[!htb]
\centering
\includegraphics[scale=.5]{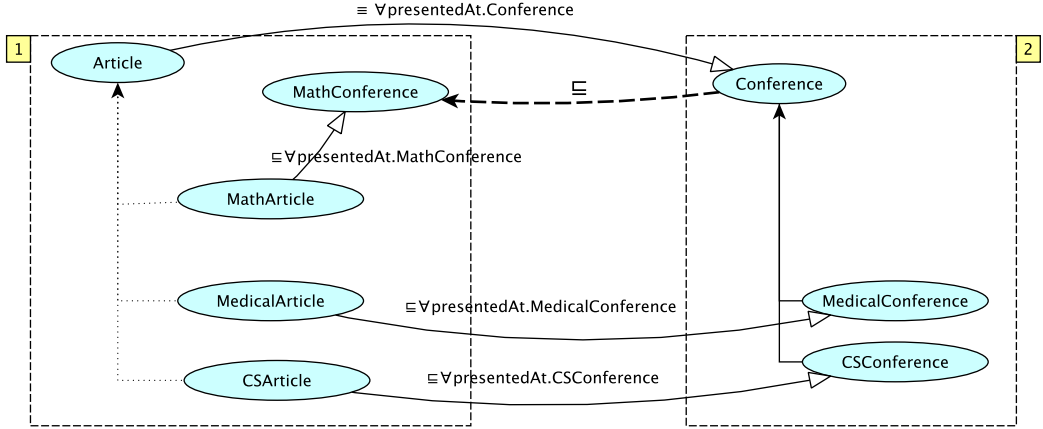}
\caption{The use of $presentedAt$ both as role and a link relation. Concept correspondences are directed to the units that hold them. Dotted lines show implied inclusion relations, and dashed lines show subjective concepts' correspondences.}
\label{fig1b}
\end{figure}

Given that the specifications in the different ontologies may evolve independently from each other, we may consider a setting where the domains of the different ontologies in figure 1 become overlapping: Such a situation is shown in figure 2, where $\mathcal{T}_1$ includes among others the specification $MathArticle \sqsubseteq (\forall presentedAt.MathConference)$, where $MathConference$ is an $1-$concept. $\mathcal{T}_2$  may include or not the concept $MathConference$ and related inclusion axioms (according to Figure 2, this concept does not exist in ontology 2). Now, in order to show that $MathArticle \sqsubseteq Article$, knowledge from both units must be combined: If, for instance, we had a single $\mathcal{SHIQ}$ ontology including all the axioms, then to deduce that $MathArticle \sqsubseteq Article$ we need to specify an additional axiom specifying that $MathConference \sqsubseteq Conference$. 

Nevertheless, 
 there are certain problems in this setting with $\mathcal{E}$-connections: First, $\mathcal{E}$-connections do not support reasoning with ontologies interpreted over overlapping logical domains. Thus, we can not combine knowledge by means of subsumption relations between concepts in different ontologies. Second, $presentedAt$ appears as a local role for ontology 1, as well as a 12-link relation. Specifications concerning the 12-link relation $presentedAt$ must be treaded in conjunction to the specifications for the 1-role  $presentedAt$, so as knowledge from the two ontologies to be coupled. We conjecture that this is so, given that ontology 1 is ``responsible'' for the naming of 1-roles and 12-link relations in any specification. The term $punning$ has been coined by $\mathcal{E}$-connections to refer to the ability of using the same name for roles and link relations in different domains. Then, the interpretation of a property in $\mathcal{E}$-connections is the disjoint union of its interpretation in each of its different ontology units. Punning is necessary in order to specify generalized transitivity axioms defined in section 2, specifying the domains in which relations are transitive. Transitivity specified by means of generalized transitivity axioms imposes the necessity of a ``bird's eye view'' of the combined reasoning process: As far as we know, no distributed or federated reasoner has been implemented for reasoning with $\mathcal{E}$-connected ontology units supporting punning and generalized transitivity axioms. 

To overcome the limitations of $\mathcal{E}$-connections, we may use P-DL. According to P-DL,  packages may also be interpreted in overlapping domains. As far as our example is concerned, terms from the package 2 may be imported to the  package 1. In this case, the specifications in $\mathcal{T}_1$ are as follows: 

\begin{align}
\{ 1:Article & \equiv (\forall 1:presentedAt.2:Conference), \\ 1:MedicalArticle & \sqsubseteq (\forall 1:presentedAt.2:MedicalConference), \\ 1:MathArticle & \sqsubseteq (\forall 1:presentedAt.1:MathConference),\\  1:CSArticle & \sqsubseteq (\forall 1:presentedAt.2:CSConference), \\ 1:MathConference & \sqsubseteq 2:Conference\}. 
\end{align}

In such a setting, a federated tableau algorithm for P-DL can deduce that $MathArticle \sqsubseteq Article$. P-DL supports both, inclusion axioms between concepts in different packages, the import of roles, as well as linking individuals in different packages via (imported) roles. However, there are two issues here: One concerns the fact that neither ``correspondences'' (if we can use this term) between the elements of different ontologies, nor relations between them are first class constructs that can be  manipulated and exploited in any independent way during reasoning. Also, while the subsumption relation relating $1:MathConference$ with $2:Conference$ concerns the 1st package (i.e. it is a  correspondence that package 1 holds), this is handled as any other subsumption relation included in $\mathcal{T}_1$ or $\mathcal{T}_2$: I.e. ontology 2   handles this subsumption relation as if  it was one  included in $\mathcal{T}_2$. This is a strong assumption given that the subjectiveness of correspondences between concepts and individuals in distinct ontologies is not preserved. The P-DL distributed ontology could still imply that $MathArticle \sqsubseteq Article$ in case the axiom $(1:MathConference \sqsubseteq 2:Conference)$ is included in $\mathcal{T}_2$. However, if these correspondences where subjective, then in the later case the conclusion may not hold. Nevertheless, P-DL can not distinguish between these two cases. The treatment of subsumption relations between concepts from different packages in this rather ``objective'' way, allows considering compositional consistent domain relations. We conjecture that subjectiveness is necessary in open and inherently distributed settings with autonomous peers: Each unit may hold correspondences to ontology elements of acquaintances, which may be unknown to other units, or to which other units do not ``agree''.

Retaining the subjectiveness of correspondences, and under the assumption of compositional consistent domain relations, the transitive propagation of subjective subsumption relations between ontologies in the general case (i.e. in the case where $into$ and $onto$ subjective bridge rules between concepts exist) present pitfalls for DDL: This is due to the fact that entailments in one ontology may be affected by correspondences between individuals in distant ontology units. To overcome this limitation, DDL support a relaxed version of compositional consistency for domain relations, requiring transitivity of domain relations in distributed models. Thus, given the specifications in figure 3, DDL  with transitive domain relations can  deduce that $PediatricConference \sqsubseteq HumanActivity$. However,  this relaxed condition does not allow chaining concept $onto$ concept bridge rules.

\begin{figure}[!htb]
\centering
\includegraphics[scale=.5]{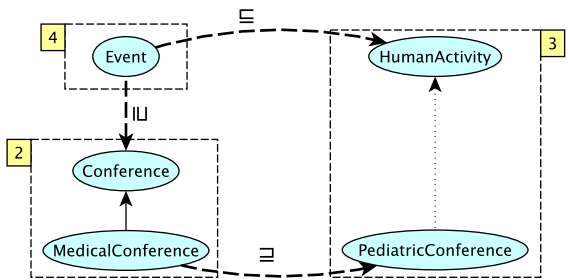}
\caption{Subsumption propagation between ontologies. Dotted lines show implied inclusion relations, and dashed lines show subjective concepts' correspondences. Concept correspondences are directed to the units that hold them. }
\label{fig3}
\end{figure}

Concluding the above, we aim to support federated reasoning while maintaining the subjectiveness of specifications in the presence of concept $onto$ concept and concept $into$ concept correspondences between ontology units, in conjunction to the existence of link relations. Towards this aim we propose $E-\mathcal{SHIQ}$, as a generic framework for combining knowledge between ontologies and for performing reasoning in distributed settings.

\begin{figure}[!htb]
\centering
\includegraphics[scale=.5]{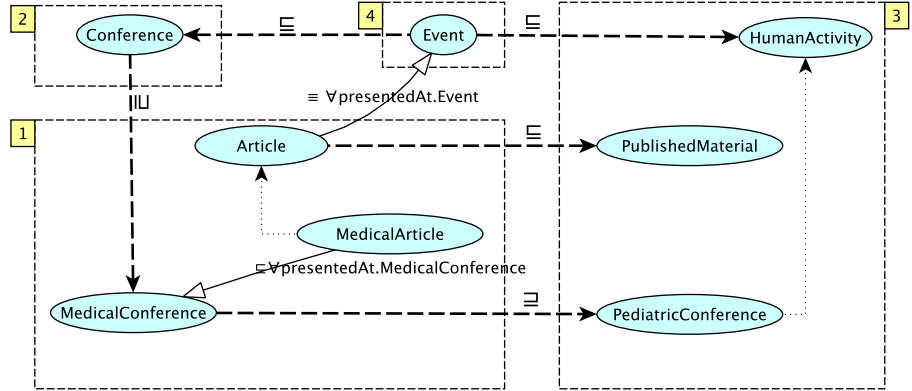}
\caption{Combination of knowledge via subjective correspondences and link relations. Concept correspondences are directed to the units that hold them. Dotted lines show implied inclusion relations, and dashed lines show subjective concepts' correspondences.}
\label{fig4}
\end{figure}

A more elaborated example presenting the kind of specifications and reasoning we aim to support with $E-\mathcal{SHIQ}$ is presented in figure 4. We further present, discuss and elaborate on these examples in the sections that follow. 

To conclude this section, considering  the example  in figure \ref{fig4} we can point out the following: 
\begin{itemize}
\item $E-\mathcal{SHIQ}$ supports subjective concept-to-concept correspondences between concepts in different ontology units. As shown in figures 1,2,3, direction of correspondences point to the ontology that holds the correspondence. Correspondences concerning equivalences between concepts are specified as conjunctions of into and onto correspondences. Also, symmetric (or bi-directional) correspondences specify correspondences to which the involved units ``agree''\footnote{This is a soft form of an agreement: a) correspondences between some units may coincide but involved units do not necessarily know it (i.e. a unit does not know the correspondences of another), b) if one of the involved units drops the correspondence, this action will not affect the symmetric correspondence of the other. Although there are means for units to  reach consensus \cite{Vouros2013}, these are not within the scope of this paper.}: i.e. for each correspondence from the subjective point of view of a unit $i$ to a unit $j$, there is a correspondence stating the same relation between ontology entities from the subjective point of view of unit $j$. 
\item $E-\mathcal{SHIQ}$ in conjunction to subjective concept-to-concept correspondences supports relating individuals in different units via link relations, as well as via subjective individual correspondence relations. While correspondence relations represent equalities between individuals, from the subjective point of view of a specific unit, link relations may relate individuals in different units via domain-specific relations. Link relations may be further restricted via value and cardinality restrictions, and they can be hierarchically related with other link relations from the same unit.
\item $E-\mathcal{SHIQ}$ supports punning by allowing roles and link relations to have the same name, within the same unit. This allows peers to reason with roles and link relations locally. This is the case for $presentedAt$, serving as a role in ontology 1 and as a 14-link relation for ontologies 1 and 4. In case $i \neq j, i,j,k \in I$ i-roles and j-roles or jk-link relations are not considered to be related in any way, even if they have the same name. To take further advantage of punning, $E-\mathcal{SHIQ}$ allows a restricted form of transitive axioms, aiming to support the  computation of the transitivity closure of a role in $i$ by local means: By doing so, the reasoning chunk in unit $i$ has to consider the local ABox of ontology $i$ and the corresponding assertions for relations between individuals in ontologies $i$ and $j$, which are known by $i$.
\item $E-\mathcal{SHIQ}$ inherently supports subsumption propagation between ontologies, supporting reasoning with concept-to-concept correspondences in conjunction to link relations between ontologies.  For instance, the relation $PediatricConference \sqsubseteq HumanActivity$ in figure 4 is entailed by considering all the correpondences between units. Also the relation $MedicalArticle \sqsubseteq Article$ is entailed by considering the specifications involving the role and link relation $presentedAt$ in unit 1, in conjunction to the subjective correspondences $2:Conference \overset \sqsubseteq \rightarrow 1: MedicalConference$ and $4:Event \overset \sqsubseteq \rightarrow 2: Conference$.
\end{itemize}

\section{Combining ontologies using $E-\mathcal{SHIQ}$}
\label{combining}

\subsection{The $E-\mathcal{SHIQ}$ Representation Framework}

\textbf{Definition 1 }($E _{HQ^+}\mathcal{SHIQ}$ Syntax)
Given a non-empty set of indices $I$,  let a collection of ontology units indexed by $I$.  Let $N_{C_i}$, $N_{R_i}$ and $N_{O_i}$ be the sets of concept, role and individual names, respectively. 

For some $R \in N_{R_i}$, $Inv(R)$ denotes the inverse role of $R$ and ($N_{R_i} \cup \{Inv(R) | R \in N_{R_i}\}$) is the set of $\mathcal{SHIQ}$ i-roles for the i-th ontology. An i-role axiom is either a role inclusion axiom or a transitivity axiom, as defined in section 2.2.  Let $\mathcal{R}_i$ be the set of i-role axioms. 

The set of ij-link relations relating individuals in two units $i$ and $j$, $i \neq j \in I$, is defined to be the set $\mathcal{E}_{ij}$. As already pointed out in section 2.2, link relations are nor pairwise disjoint, but are disjoint  with respect to the set of concept names. An $ij-relation$ $box$ $\mathcal{R}_{ij}$ includes a finite number set $ij-link$ $relation$ inclusion axioms, as well as transitivity axioms of the form $Trans(E,(i,j))$, where $E$ is in ($\mathcal{E}_{ij} \cap N_{R_i}$), i.e. it is an ij-link relation and an i-role. As already told, this restricted form of  transitivity axioms preserve the locality of specifications for the ontology unit $i$, and it is a shorthand for the generalized transitivity axiom $Trans(E;(i,i),(i,j))$. 
 
The sets of $i-concepts$ are inductively defined by the constructors in table 2, given that $R$ and $S$ are i-roles, and $E_{ij}$ are ij-link relations. 

Let $i:C$ and $i:D$ possibly complex concepts and $i:C \sqsubseteq i:D$  (or $i:C \sqsubseteq D$) a \textit{general concept inclusion} (GCI) axiom. A finite set of GCI's is a TBox for the ontology unit $i$ and it is denoted by $\mathcal{T}_i$.

Concepts' correspondences may be concept $onto$ concept, or concept $into$ concept: Let $C \in N_{C_i}$, $D \in N_{C_j}$ with $i \neq j \in I$. A concept $onto$ ($into$) concept correspondence from  unit $i$ to unit $j$ from the subjective point of view of $j$ is of the form $i:C \overset{\sqsupseteq} \rightarrow j:D$ (corresp. $i:C \overset{\sqsubseteq} \rightarrow j:D$).

Subsequently, roles and link-relations are referred as ``properties''. If necessary, it is stated explicitly whether a property is a role or a link-relation. Also, when we use indices, e.g. $i,j \in I$, without stating whether these are equal or different, then both cases can be true. Thus, in case $i=j$, then the $\textit{ij-}property$ box is the RBox for the $i$-th unit, i.e. (with an abuse of notation) $\mathcal{R}_{ii}=\mathcal{R}_{i}$. Similarly, we refer to a property $E_{ij}$, denoting either an i-role in case $i=j$, or a link relation in case $i \neq j$. In cases where $i \neq j$, then this is stated explicitly.  
\\

\textbf{Definition 2 }(Distributed Knowledge Base)
A \textit{distributed knowledge base} $\Sigma=\langle \textbf{T}, \textbf{R}, \textbf{C}\rangle$ is composed by the distributed TBox \textbf{T}, the distributed RBox $\textbf{R}$, and a tuple of sets of correspondences $\textbf{C}=(\textbf{C}_{ij})_{i \neq j \in I}$ between ontology units. 
A \textit{distributed TBox} is a tuple of TBoxes  \textbf{T}$=(\mathcal{T}_i)_{i \in I}$, where each $\mathcal{T}_i$ is a finite set of i-concept inclusion axioms. 
A \textit{distributed RBox} is a tuple of $ij$-property boxes  $\textbf{R}=(\mathcal{R}_{ij})_{i,j \in I}$, where each $\mathcal{R}_{ij}$ is a finite set of property inclusion axioms and transitivity axioms. 

A \textit{distributed ABox} (DAB) includes a tuple of local ABox'es $\mathcal{A}_i$ for each ontology $i$, and  sets $\mathcal{A}_{ij}, i\neq j$ with individual correspondences of the form $\textit{j:a} \overset {=} \mapsto \textit{i:b}$, and property assertions of the form 
$(a,b): E_{ij}$, where $E_{ij}$ is an ij-link relation in $\mathcal{E}_{ij}, i\neq j$. Thus, individual correspondences are specified from the subjective point of view of $i$ and, together with assertions concerning linked individuals, these are made locally available to $i$.
\\ 

\textbf{Example} (Distributed Knowledge Base)
To exemplify the constructors provided by  $E-\mathcal{SHIQ}$ and their use for combining knowledge from different ontologies, let us consider the knowledge base shown in figure 4. 

Given that the set of indices is $I=\{1,2,3,4\}$, then, according to definition 2, the distributed knowledge base $\Sigma=\langle \textbf{T}, \textbf{R}, \textbf{C}\rangle$ is composed by the distributed TBox \textbf{T}, the distributed RBox $\textbf{R}$, and a tuple of sets of correspondences $\textbf{C}=(\textbf{C}_{ij})_{i \neq j \in I}$ between ontology units.  
Specifically,  
\begin{itemize}
\item \textbf{T}$=(\mathcal{T}_{i})_{i \in I}$, where \\ $\mathcal{T}_i = \emptyset, i \in \{2,3,4\} $ and \\ $\mathcal{T}_1 = \{MedicalArticle \sqsubseteq (\forall presentedAt.MedicalConference), \\ Article \equiv (\forall presentedAt.4:Event) \}$.
\item $\mathcal{R}=((R_{i})_{i \in I}, (R_{ij})_{i \neq j \in I})$, where \\ $R_{i}=R_{ij} =\emptyset$, $i,j \in I$,
\item \textbf{C}$=(\textbf{C}_{ij})_{i \neq j \in I}$, where  \\ $\textbf{C}_{21}= \{2:Conference \overset \sqsupseteq \rightarrow1:MedicalConference\}$ \\ $\textbf{C}_{13}= \{1:MedicalConference \overset \sqsupseteq \rightarrow 3:PediatricConference, \\ 1:Article \overset \sqsubseteq \rightarrow3:PublishedMaterial\}$, \\ $\textbf{C}_{42}= \{4:Event \overset \sqsupseteq \rightarrow 2:Conference \}$ and \\ $\textbf{C}_{43}= \{4:Event \overset \sqsubseteq \rightarrow 3:HumanActivity\}$
\item $DAB=((\mathcal{A}_i)_{i \in I}, (\mathcal{A}_{ij})_{i \neq j \in I})$, where \\ $\mathcal{A}_i = \emptyset$ and $\mathcal{A}_{ij}= \emptyset$, for any $i,j \in I$
\end{itemize}
\vspace*{2\baselineskip}
Each TBox $\mathcal{T}_{i}$, ${i \in I}$
is locally interpreted by a local, possibly hole interpretation $\mathcal{I}_i$ that consists of a domain $\Delta _i$ and a valuation function $\cdotp^{\mathcal{I}_i}$ 
that maps every concept to a subset of $\Delta_i$.
The ij-property boxes $\mathcal{R}_{ij}$, with $i,j \in I$, 
are interpreted by valuation functions $ \cdotp^{\mathcal{I}_{ij}} $ 
that map every ij-property to a subset of $\Delta_i \times \Delta_j$.  Let $\mathcal{I}_{ij}=\langle \Delta_{i}, \Delta_{j}, \cdotp^{\mathcal{I}_{ij}}  \rangle$, $i,j \in I$. It must be pointed out that in case $i=j$, then (by abusing notation) $\cdotp^{\mathcal{I}_{ij}}=\cdotp^{\mathcal{I}_{i}}$ and $\mathcal{I}_{ij}=\mathcal{I}_{i}=\langle \Delta_{i}, \cdotp^{\mathcal{I}_{i}} \rangle$.
\\ 

\textbf{Definition 3 }(Domain relations) 
Domain relations $r_{ij}, i \neq j \in I$ represent equalities between individuals, from the subjective point of view of $j$. A \textit{domain relation} $r_{ij}, i \neq j$ from $\Delta_i$ to $\Delta_j$ is a subset of $\Delta_i \times \Delta_j$, s.t. in case $d' \in r_{ij}(d_1)$ and $d' \in r_{ij}(d_2)$, then according to the subjective view of $j$, $d_1= d_2$ (denoted by $d_1=_j d_2$). Also, given a subset $D$ of $\Delta ^{\mathcal{I}_i}$, $r_{ij}(D)$ denotes $\cup_{d \in D} r_{ij}(d)$.

Given that domain relations represent equalities, in case $d_1 \in r_{ij}(d)$ and $d_2 \in r_{ij}(d)$, then  $d_1=_j d_2$ ( it must be noticed that $d_1$ and $d_2$ are individuals in the ontology $j$).  
Therefore, $E-\mathcal{SHIQ}$ domain relations are  globally one-to-one relations.
 \\ 
 
\textbf{Definition 4 }(Distributed Interpretation) 
Given the index $I$ and $i,j \in I$, a \textit{distributed interpretation} $\mathfrak{I}$ of a distributed knowledge base $\Sigma$ is the tuple formed by the interpretations $\mathcal{I}_{ij}=\langle \Delta_{i}, \Delta_{j}, \cdotp^{\mathcal{I}_{ij}}  \rangle$, $i,j \in I$, and a set of domain relations $r_{ij}$, in case $i \neq j \in I$. Formally, $\mathfrak{I}=\langle (\mathcal{I}_{ij})_{i,j \in I}, (r_{ij})_{i\neq j \in I} \rangle$. 

A local interpretation $\mathcal{I}_i$ satisfies an i-concept $C$ w.r.t. a distributed knowledge base $\Sigma$, i.e. $\mathcal{I}_i \vDash i:C$ iff $C^{\mathcal{I}_i}\neq \emptyset$.  $\mathcal{I}_i$ satisfies an axiom $C \sqsubseteq D$  between i-concepts ( i.e. $\mathcal{I}_i \vDash i: C \sqsubseteq D$) if $C^{\mathcal{I}_i} \subseteq D^{\mathcal{I}_i}$. Also,  $\mathcal{I}_{ij}$ satisfies an \textit{ij-property inclusion axiom} $R \sqsubseteq S$   ($\mathcal{I}_{ij} \vDash R \sqsubseteq S$) if $R^{\mathcal{I}_{ij}} \subseteq S^{\mathcal{I}_{ij}}$. A transitivity axiom $Trans(E;(i,j))$ is satisfied  by $\mathfrak{I}$ iff $E^{\mathcal{I}_{i}} \cup E^{\mathcal{I}_{ij}}$ is transitive.
 
The distributed interpretation $\mathfrak{I}$ \textit{satisfies} ($\vDash _d$) the elements of a distributed knowledge base, if the following conditions hold:
\begin{center}
\resizebox{1.0\textwidth}{!}{
\begin{tabular}{r l}
 \textbf{1.} & $\mathfrak{I} \vDash _d i:C \sqsubseteq D$, if $\mathcal{I}_i \vDash C \sqsubseteq D$\\
 \textbf{2.} & $\mathfrak{I} \vDash _d \mathcal{T}_i$ if $\mathfrak{I} \vDash i:C \sqsubseteq D$ for all $C \sqsubseteq D$ in $\mathcal{T}_i$\\ 
\textbf{3.} & $\mathfrak{I} \vDash _d \textit{j:C} \overset{\sqsubseteq}{\to{}} \textit{i:D}$, if $r_{ji}(C^{\mathcal{I}_j})\subseteq D^{\mathcal{I}_i}$\\
\textbf{4.} & $\mathfrak{I} \vDash _d \textit{j:C} \overset{\sqsupseteq}{\to{}} \textit{i:D}$, if $r_{ji}(C^{\mathcal{I}_j})\supseteq D^{\mathcal{I}_i}$\\
\textbf{5.} & $\mathfrak{I} \vDash _d \textbf{C}_{ij}$, if $\mathfrak{I}$ satisfies all correspondences in $\textbf{C}_{ij}$\\
 \textbf{6.} & $\mathfrak{I} \vDash _d R \sqsubseteq S$, if $\mathcal{I}_{ij} \vDash R \sqsubseteq S$, where $R \sqsubseteq S$ in $\mathcal{R}_{ij}$\\
 \textbf{7.} & $\mathfrak{I} \vDash _d \mathcal{R}_{ij}$ if $\mathfrak{I} \vDash _d R \sqsubseteq S$ and $\mathfrak{I} \vDash _d Trans(E;(i,j))$ for all inclusion and transitivity axioms in $\mathcal{R}_{ij}$  \\  
 \textbf{8.} & $\mathfrak{I} \vDash _d \Sigma$ if for every $i,j \in I, \mathfrak{I} \vDash _d \mathcal{T}_i$, $\mathfrak{I} \vDash _d \mathcal{R}_{ij}$, $\mathfrak{I} \vDash _d \mathcal{A}_{i}$ and $\mathfrak{I} \vDash _d \textbf{C}_{ij}$\\
\end{tabular}}  
\end{center}
 \normalsize

Finally, it also holds that \\ $\mathfrak{I} \vDash_d \textit{i:a} \overset{=}{\mapsto} \textit{j:b}$ if $b^{\mathcal{I}_j} \in r_{ij}(a^{\mathcal{I}_i}), i \neq j \in I$. 

Notice that in case $i=j$, condition (7) becomes \\ $\mathfrak{I} \vDash _d \mathcal{R}_{i}$ if $\mathfrak{I} \vDash _d R \sqsubseteq S$ and $\mathfrak{I} \vDash _d Trans(E)$ for all inclusion and transitivity axioms in $\mathcal{R}_{i}$.

According to the above, given an individual  \textit{i:x} in $(\Delta^{\mathcal{I}_i} \cap \Delta^{\mathcal{I}_j})$, then, according to $i$ (and/or $j$), there must be a corresponding individual \textit{j:y}, s.t. $\textit{x}^{\mathcal{I}_i}=r_{ji}(\textit{y}^{\mathcal{I}_j})$ (respectively, $\textit{y}^{\mathcal{I}_j}=r_{ij}(\textit{x}^{\mathcal{I}_i})$, from the subjective point of view of $j$). 

Also, as a consequence of the fact  that domain relations represent subjective equalities, and given that individuals must satisfy some constraints imposed by the semantics of specifications, then from the subjective point of view of the unit that holds the individuals correspondences, corresponding individuals must satisfy the same sets of constraints. Thus, while we do not require domain relations to be transitive,  corresponding  individuals must satisfy the same set of constraints (i.e. they must belong to the denotation of the same concepts). We exemplify this in the example that follows definition 5. 
\\

\textbf{Definition 5 }(Distributed entailment and satisfiability) 
$\Sigma \vDash _d X \sqsubseteq Y$ if for every $\mathfrak{I}$, $\mathfrak{I} \vDash _d \Sigma$ implies $\mathfrak{I} \vDash _d X \sqsubseteq Y$, where $X$ and $Y$ are either i-concepts or ij-properties, $i,j \in I$. 
$\Sigma$ is satisfiable if there exists a $\mathfrak{I}$ s.t. $\mathfrak{I} \vDash _d \Sigma$. A concept \textit{i:C} is satisfiable with respect to $\Sigma$ if there is a $\mathfrak{I}$ s.t. $\mathfrak{I} \vDash _d \Sigma$ and $C^{\mathcal{I}_i} \neq \emptyset $. 
\\

\textbf{Example} (Entailments)
Continuing the example shown in figure 4, according to the semantics of specifications, we can prove that from the subjective point of view of unit 3 it holds that $ PediatricConference \sqsubseteq HumanActivity$.  

Let $x$ be an individual, s.t. $x \in PediadricConference^{\mathcal{I}_3} \backslash HumanActivity^{\mathcal{I}_3}$. Given that $x \notin HumanActivity^{\mathcal{I}_3}$, then from the subjective point of view of 3, since $\textit{4:Event} \overset{\sqsubseteq}{\to{}} \textit{3:HumanActivity}$, there can not be any  $y$ s.t. $y  \in Event^{\mathcal{I}_4}$ and $r_{43}(y)=x$. Nevertheless, since $1:MedicalConference \overset \sqsupseteq \rightarrow 3:PediatricConference$, from the subjective point of view of unit 3, there is an $x' \in MedicalConference^{\mathcal{I}_1} $ s.t. $r_{13}(x')=x$. Given that domain relations represent (subjective) equalities, $x'$ must satisfy the constraints for $x$, i.e it must not have any correspondent individual in $Event^{\mathcal{I}_4}$. This is a constraint imposed from the subjective point of view of 3. Given this constraint for $1:x'$, and the fact that according to unit 1  there is an $x''$ in unit 2 s.t. $x'' \in Conference^{\mathcal{I}_2} $ and $r_{21}(x'')=x'$, any such $x''$ (and any of its corresponding individuals in any ontology) must not belong in the denotation of concept \textit{4:Event}. However, according to 2, $r_{42}(Event^{\mathcal{I}_4}) \supseteq Conference^{\mathcal{I}_2}$, implying that $x'' \in r_{42}(Event^{\mathcal{I}_4})$, which according to 3 leads to a contradiction. 
\\

As it can be observed from the above example, given that domain relations represent subjective equalities between corresponding individuals, these individuals must satisfy the same constraints concerning their inclusion in the denotation of specific concepts. Therefore, in our example, unit 3 does not possess any information about $x"$, and about the correspondence between \textit{3:x} and \textit{2:x''}. Nevertheless, unit 3 has set specific constraints that \textit{3:x} and all corresponding individuals must satisfy, in conjunction to other units' local and subjective specifications. The underlying assumption in this type of behaviour is that peers possessing  ontology units are fully collaborative: i.e. peers try to find a $shared$ distributed model, i.e a model that satisfies all the constraints set by all units. This propagation of constraints is specified by new tableau extension rules, for the construction and maintenance of a distributed completion graph during reasoning. This is presented in the next section.  
\\
\section{Reasoning in $E-\mathcal{SHIQ}$ }
\label{reasoning}

\subsection{Overview}

We consider DL reasoners that implement a tableau method for deciding concept satisfiability w.r.t. a knowledge base. In order to check the satisfiability of a concept $C$ w.r.t. a knowledge base, the reasoner constructs a model for $C$, satisfying all the constraints implied by the semantics of the (local and subjective) specifications in the knowledge base. 

A  tableau algorithm is characterized by the following elements:

\begin{itemize}
\item An underlying data structure, called the completion graph
\item A set of expansion rules
\item A blocking condition for ensuring termination
\item A set of conditions to detect contradictions (clashes)
\end{itemize}

The specific properties of the completion graph, the blocking and clash conditions and the expansion rules, depend on the expressivity of the language used. Here we aim at distributed reasoning using ontologies that are combined according to the $E-\mathcal{SHIQ}$ framework. Reasoning in this case combines local reasoning chunks performed over combined $ \mathcal{SHIQ}$ ontologies. The completion graph is distributed among units (actually among reasoning peers exploiting these units), while the expansion rules assure expansion and maintenance of the  completion graph in each of the peers, and the proper maintenance of correspondences among individuals in different units. This is done so as to retain the properties of a tableau. Specifically, the completion graph constructed is a finite directed graph representing a model of the input concept w.r.t. the distributed knowledge base. Subsequently, we use the term $peer$ to specify the active reasoning entity that exploits the knowledge in a specific unit, jointly with other peers. Peers, as units, are distinguished by the  set of units' indices $I$: Peer $i$ exploits the specifications in unit $i$.

Given that in a distributed setting with an $E-\mathcal{SHIQ}$ distributed ontology no unit possesses the combined knowledge, the completion graph is distributed to the different peers. Local chunks of this graph are combined via specific subjective correspondences among graph nodes. Each node and edge in the graph is labeled with a set of concepts and properties (i.e. roles or link relations), respectively. These labels specify the constraints that  individuals in different units must \textit{jointly} satisfy for a model to exist. In the case of $E-\mathcal{SHIQ}$, where knowledge is distributed among units, and where there are specific subjective equality correspondences between individuals in different units, the set of constraints that applies to an individual must also be satisfied by all individuals that are (subjectively) equal to it: In our case, this means that nodes maintained by peer $i$ are not only labeled by i-concepts, but by  concepts of other units, specifying constraints imposed by corresponding individuals in those units. To propagate constraints among peers and combine their local completion graphs, in addition to expansion rules used for $\mathcal{SHIQ}$, the tableau algorithm  for $E-\mathcal{SHIQ}$  uses  rules that project individuals to their corresponding individuals in other units and update the labels of corresponding individuals.

The algorithm repeatedly applies the expansion rules until constraints in a node are determined unsatisfiable (i.e. until a clash occurs), or until a clash-free graph is constructed and no expansion rule can further expand/refine that graph.

Blocking conditions guarantee termination of the tableau algorithm.  A blocking mechanism is correct if a path starting from a blocked node would never yield to a clash if the expansion rules had been further applied. While there may be different blocking strategies, in the case of $\mathcal{SHIQ}$ the double blocking strategy is applied. Also, due to link relations and individual correspondences, we also use a subset blocking strategy.

Concluding the above, in addition to distributed tableau algorithms, the proposed distributed algorithm for $E-\mathcal{SHIQ}$
\begin{itemize}
\item Expands and maintains a true distributed completion graph within and across different peers: No peer has a global, complete view of the overall completion graph. 
\item Labels of nodes include concepts that may have been set by distant peers. This means that the label of a node in the completion graph of peer $i$ may include  concepts from units different that $i$, as well as own ones. Concepts in labels specify constraints that nodes (individuals) must jointly satisfy: I.e. an individual must be in the denotation of any concept in its label.
\item Projection of individuals from one peer to another, according to the semantics of specifications, guarantees propagation of (subjective) constraints and maintenance of a joined view of all the constraints that corresponding individuals in a model must satisfy.
\end{itemize}

\subsection{A Tableau  for $E-\mathcal{SHIQ}$}

We assume all concepts in Negation Normal Form (NNF) \footnote{To transform any concept into NNF,  De Morgan's laws are applied, as well as the duality between atmost and atleast number restrictions. In NNF, negation occurs in front of concept names only.}. Given an i-concept $C$ and the distributed knowledge base $\Sigma= \langle \textbf{T}, \textbf{R}, \textbf{C} \rangle$, the set $sub_i(C,\textbf{R})$ (or simply $sub_i(C)$), is computed over the set of indices $I$, by the following conditions:
\\
\textbf{1.} if $C$ is in $N_{C_i}$ or the negation of a concept name in $N_{C_i}$, then $sub_j(C,\textbf{R})=\{C\}$, $i,j \in I$.\\
\textbf{2.} if $C$ is a complex concept of the form $C_1 \sqcap C_2$ or $C_1 \sqcup C_2$, then \\
$sub_j(C,\textbf{R})=\{C\} \cup sub_j(C_1,\textbf{R}) \cup sub_j(C_2,\textbf{R})$, $i,j \in I$. \\
\textbf{3.} if $C$ is of the form $\exists P .G$, $\forall P. G$, $\leq n P. G$ or $\geq n P. G$ then  $sub_j(C,\textbf{R})=\{C\} \cup sub_j(G,\textbf{R})$, $i,j \in I$, where $P$ is either an i-role or an ij-link relation.

\noindent The $closure$ of the i-concept $C$ over $\Sigma = \langle \textbf{T}, \textbf{R}, \textbf{C} \rangle$, is 

\noindent $cl_i(C,\Sigma)=cl_i(C,\textbf{R}) \cup \bigcup_{j=1,..,n}cl_i(C_{\mathcal{T}_{j}},\textbf{R}) $, where \small $C_{\mathcal{T}_{j}}=\sqcap _{C \sqsubseteq D \in \mathcal{T}_j}(\neg C \sqcup D)$ \normalsize 

\noindent and \small $cl_i(C,\textbf{R})$ \normalsize is the set of concepts in $sub_i(C,\textbf{R})$ under the NNF.

To define the distributed tableau for a concept $k:C$ w.r.t. a distributed knowledge base $\Sigma$, let us consider the non-empty set of individuals $S_k$ in the $k$-th unit. The label $\mathcal{L}_k$ of each individual \textit{k:x} in  $S_k$ can contain any concept in $cl_j(C,\Sigma)$. Thus, according to the above conditions, it may contain any $\textit{j-}concept$, with $j \in I$. 
Doing so, each label $\mathcal{L}_k$ can be seen as the union of disjoint sets $\mathcal{L}_{kj}$, each one including the $\textit{j-}concepts$ in $\mathcal{L}_k$, only. Such a set is called the \textit{j-label part} of $\mathcal{L}_k$.  We denote by $\mathcal{L}_{-k}$ the set $\bigcup_{m \in (I-\{k\})} \mathcal{L}_{km}$.
$\mathcal{L}_{-k}$ includes all the constraints that corresponding individuals of $x$ in units different from $k$ must satisfy. Furthermore, the   $neighborhood$ of unit $k$ is defined to be the set $N(k) =\{m | m \in I$ and $\textbf{C}_{km} \cup \mathcal{E}_{km}  \neq \emptyset\} \cup \{k\}$, i.e. $N(k)$ denotes the set of units to which $k$ is combined.  $N(k)$ specifies also the neighborhood of the peer $k$; i.e. of the peer that exploits the unit $k$.\\

\textbf{Definition 6 }(j-Projection) 
A \textit{j-projection}, $\pi_{kj}$ of an individual \textit{k:x} labeled by $\mathcal{L}_k$  to a unit $j \in N(k)-\{k\}$ is an individual $\textit{j:x'}=\pi_{kj}(\textit{k:x})$ in the $j$-th unit, s.t. $(\mathcal{L}_{k}(x)  \cap \mathcal{L}_{j}(x')) \supseteq \mathcal{L}_{-k}(x) \neq \emptyset$, and $\textit{j:x'} \overset {=} \mapsto \textit{k:x}$.
\\

The rationale behind this definition of projection is as follows:  The projection of an individual from unit $k$ to unit $j$ includes all i-concepts (i.e. constraints concerning membership of this individual), with $i \neq k$, since these concepts may eventually be propagated to the reasoning chunk for unit $i$, which should be able to detect a contradiction (clash) that is due to the coupled (subjective) knowledge.  The super-set relation among labels indicates that the labels of  \textit{k:x} and \textit{j:x'} may include additional concepts due to (subjective) knowledge concerning \textit{j:x'} in unit $j$. This is further elaborated in section 5.4.
\\


\textbf{Definition 7 }(Distributed Tableau) 
A \textit{distributed tableau T} for an i-concept $D$ w.r.t. $\Sigma$ is defined to be a family of tuples $\langle S_{k}, \mathcal{L}_k , (a_{kl}),  (\pi_{kj})_{k \neq j} \rangle$, where $k,l,j \in I$, $S_k$ is the non-empty set of individuals in the $k$-th unit, $\mathcal{L}_k$ maps each individual to a subset of  $\bigcup_{m \in I} cl_m(D,\Sigma)$. $a_{kj}$ maps each kj-property \footnote{It must be noticed that in case $k=j$, then $a_{kj}$ concerns a $k-$role, and it is denoted by $a_k$. },  to a set of pairs of individuals, while $\pi_{kj}$ is the j-projection of individuals from the $k$-th unit. Also, for each $s_k \in S_k$,   $C_{K_{kj}} \in \mathcal{L}_k(s_k)$, where $C_{K_{kj}}= C_{\mathcal{T}_k}\sqcap_{j:F \overset{\sqsupseteq}{\rightarrow}  k:E \in \textbf{C}_{jk}}(\neg k:E \sqcup j:F)\sqcap_{ j:H  \overset{\sqsubseteq}{\rightarrow} k:G \in \textbf{C}_{jk}}(\neg j:H \sqcup k:G)$.

The concept $C_{K_{kj}}, k,j \in I$ allows reducing $\mathcal{T}_k$ and all subjective correspondences of unit $k$ to any unit $j$, to a single formula. This is so, since from the subjective point of view of $k$ each distributed interpretation must satisfy $\mathcal{T}_k$ and  all the correspondences in $\textbf{C}_{jk}$.

The tableau, as it is defined above, is distributed to different peers, where each peer holds a part of the tableau. Local chunks of the tableau are connected via projections and link relations between individuals.
\\

\textbf{Example.} Let us for instance, consider only the 1st, 2nd and the 4th unit in the example depicted in figure 4. A snapshot of the distributed tableau constructed for checking the satisfiability of the concept $(MedicalArticle \sqcap \neg Article)$ is as follows:
\\
$T=\{\langle S_1, \mathcal{L}_1, (a_{1}, a_{12}, a_{14}), (\pi_{12}, \pi_{14}) \rangle,  \langle S_2, \mathcal{L}_2, (a_{2}, a_{21}, a_{24}), (\pi_{21}, \pi_{24}) \rangle, \langle S_4, \mathcal{L}_4, (a_{4}, a_{41}, a_{42}), (\pi_{41}, \pi_{42}) \rangle \}$, where
\begin{itemize}
\item $S_1 = \{x\}$
\item $S_2 = \emptyset$
\item $S_4 = \{y\}$
\item $\mathcal{L}_1(x)=\{ MedicalArticle, \neg Article, (\neg MedicalArticle \sqcup (\forall presentedAt.MedicalConference)), \\ (\neg Article \sqcup (\forall presentedAt.4:Event)), (Article \sqcup \neg (\forall presentedAt.4:Event)), \\ (\neg MedicalConference \sqcup 2:Conference )\}$ = 

$\{ MedicalArticle, \neg Article, (\forall presentedAt.MedicalConference), \\ (\neg Article \sqcup (\forall presentedAt.4:Event)), (\neg (\forall presentedAt.4:Event)), (\neg MedicalConference \sqcup 2:Conference )\}$ =

$\{ MedicalArticle, \neg Article, (\forall presentedAt.MedicalConference), \\ (\neg Article \sqcup (\forall presentedAt.4:Event)), (\exists presentedAt. \neg 4:Event), (\neg MedicalConference \sqcup 2:Conference )\}$



\item $\mathcal{L}_4(y) \supset \{\neg 4:Event, 1:MedicalConference\}$, where $y$ is an $presentedAt$ successor of $x$ in unit 4:

\item $a_{14}(presentedAt)= \{(x,y)\}$

\item for each $i,j \in I$, s.t. $i \neq j$, $i \neq 1$ and $j \neq 4$, $\mathcal{E}_{ij} \cap N_{R_{i}}= \emptyset$. Thus $a_{ij}=\emptyset$.

\item Similarly, for each $i \in I-\{1\}$, it holds that $a_i=\emptyset$.
\end{itemize}

Obviously, peer 1 must further develop the alternatives for node $1:x$, while the label of $4:y$ may be further developed as it will eventually contain more elements than those shown above. 
\\

For all $i,j \in I$, $x \in S_i$, $z, y \in S_j$, $Z, C,C_1,C_2 \in \bigcup_{k \in I}  cl_k(D,\Sigma)$, $E,E_1,E_2 \in \mathcal{E}_{ij}$, and $E(s,C)=\{t \in S_j | (s,t) \in a_{ij}(E)$, and $C \in L_j(t)\}$, the distributed tableau must satisfy the following properties:
\begin{center}
\resizebox{1.0\textwidth}{!}{
\begin{tabular}{l l}
\textbf{1.} & if $C \in \mathcal{L}_i(x)$, then $\neg C \not \in \mathcal{L}_i(x)$\\
\textbf{2.} & if $(C_1 \sqcap C_2) \in \mathcal{L}_i(x)$ then $C_1\in \mathcal{L}_i(x)$ and $C_2 \in \mathcal{L}_i(x)$\\
\textbf{3.} & if $(C_1 \sqcup C_2) \in \mathcal{L}_i(x)$ then $C_1\in \mathcal{L}_i(x)$ or $C_2 \in \mathcal{L}_i(x)$,  \\
\textbf{4.} & if $\forall E . Z \in \mathcal{L}_i(x)$ and $(x,y)\in a_{ij}(E)$ then $Z \in \mathcal{L}_j(y)$\\
\textbf{5.} & if $\exists E . Z \in \mathcal{L}_i(x)$ then there is some $y \in S_j$ s.t. $(x,y)\in a_{ij}(Es)$, $Z \in \mathcal{L}_j(y)$\\
\textbf{6.} & if $(x,y) \in a_{i}(E)$, $(y,z) \in a_{ij}(E)$, and $Trans(E;(i,j)) \in \mathcal{R}_{ij}$ then $(x,z) \in a_{ij}(E)$ \\
\textbf{7.} & if $((x,y) \in a_{ij}(E_1))$ and $(E_1 \sqsubseteq^* E_2)$ then $(x,y) \in a_{ij}(E_2)$\\
\textbf{8.} & if $\leq n E. Z \in \mathcal{L}_i(x)$ then $|E_{ij}(x,Z)| \leq n$\\
\textbf{9.} & if $\geq n E . Z \in \mathcal{L}_i(x)$ then $|E_{ij}(x,Z)| \geq n$\\
\textbf{10.} & if $\{(\geq n E. Z) , (\leq n E. Z)\} \cap \mathcal{L}_i(x)\neq \emptyset$, $(x,y) \in a_{ij}(E)$ then $\{Z, \neg Z\}\cap \mathcal{L}_j(y)\neq \emptyset$\\
\textbf{11.} & if $\mathcal{L}_{-i}(x) \neq \emptyset$,  then for each $j \in I$ with  $j \neq i$ and $\mathcal{L}_{ij}(x) \neq \emptyset$, there is an individual $j:x'$ such that \\&$x'=\pi_{ij}(x)$, i.e. $j:x' \overset {=} \mapsto i:x$ and $(\mathcal{L}_{i}(x)  \cap \mathcal{L}_{j}(x')) \supseteq \mathcal{L}_{-i}(x)$.\\
\end{tabular}}
\end{center}
\normalsize

These properties address the seamless treatment of roles  and link-relations in each unit (given that for $i,j \in I$ it may hold that either $i=j$ or $i \neq j$), and the maintenance of the labels of corresponding individuals in neighbor units, according to definition 6.
\\ \\
\textbf{Theorem} (Satisfiability)
An i-concept $D$ is satisfiable w.r.t. the distributed knowledge base $\Sigma=\langle \textbf{T}, \textbf{R}, \textbf{C} \rangle$, iff it has a distributed tableau w.r.t. $\Sigma$. 

\textbf{Proof}
Given a distributed tableau for the $i$-concept $D$ w.r.t. $\Sigma$ a common model for $D$ and $\Sigma$ is defined to be $\mathfrak{I}=\langle (\mathcal{I}_{ij})_{i,j \in \text{I}}, (r_{ij})_{i\neq j \in \text{I}} \rangle$ with
\begin{itemize}
\item $\Delta_{i}$ $= S_{i}$
\item $A^{\mathcal{I}_i} = \{s | A \in \mathcal{L}_{i}(s) \}$ for all atomic i-concepts $A$,
\item $E^{\mathcal{I}_{ij}} =(a_{ij}(E) \cup a_{i}(E))$ if $E \in \mathcal{E}_{ij} \cap N_{R_i}$, otherwise $E^{\mathcal{I}_{ij}} =(a_{ij}(E))$\footnote{Please notice that cases where $i = j$ and $i \neq j$ can be distinguished.},
\item $r_{ij}(s^{\mathcal{I}_i})=\{d \in S_j | \pi_{ji}(d)=s \}$, $s^{\mathcal{I}_i} \in \Delta_i$
\end{itemize}

Properties 6 and 7 of the distributed tableau assure the correct interpretation of transitive i-roles and link-relations, in conjunction to the interpretation of their sub-roles and sub-link relations. 

Then, the proof that $\mathfrak{I}$ is a model for $D$ w.r.t. $\Sigma$ follows by induction in the structure of i-concepts.
We will not repeat the cases for local roles and link-relations that have already proved in \cite{Grau04tableau}. Additional cases that need to be considered concern the treatment of properties that are specified both as roles and as link relations between units, as well as, the treatment of concepts' correspondences: 

(a)  Let us consider the cases where a property $P$ belongs to $N_{R_i} \cap \mathcal{E}_{ij}$ and either (i) there is an existential or at-least restriction to  the i-role $P$ (denoted by $\langle Restr \rangle P$) and a value restriction to the ij-link relation $P$, or (ii) there is an existential or at-least restriction (also denoted by $\langle Restr \rangle P$)  to  the ij-link relation $P$ and a value restriction to the i-role $P$.

In the case (a.i) above, $\{\langle Restr \rangle P.C_1, \forall P.C_2 \} \subseteq \mathcal{L}_i(x)$, where $C_1$ is an i-concept and $C_2$ is a j-concept. 
According to the properties (8) and (5) of the distributed tableau, there is   a $y$ in $S_i$, s.t. $(x,y) \in a_{i}(P)$, and $i:C_1 \in \mathcal{L}_i(y)$. Also, according to the property (4) of the distributed tableau, since $(x,y) \in a_{i}(P)$, it holds that $j:C_2 \in \mathcal{L}_i(y)$. Therefore the label of $y$ in the i-th unit has an i-concept and a j-concept. According to the correspondence between the model and the distributed tableau, and by induction it holds that $y \in (C_1)^{\mathcal{I}_i} \cap (C_2)^{\mathcal{I}_j}$. Also, given that $\mathcal{L}_{ij}(y) \neq \emptyset$, according to property (11) of the distributed tableau, there is an individual $y'$ in the $j$-th unit, such that $y'=\pi_{ij}(y)$ and $\mathcal{L}_{j}(y') \supseteq \mathcal{L}_{-i}(y) \supset \{j:C_2\}$. Finally, according to the correspondence between the model and the distributed tableau, $y \in r_{ji}(y')$. 

Accordingly, in the case (a.ii) above, $\{\langle Restr \rangle P.C_1, \forall P.C_2 \} \subseteq \mathcal{L}_i(x)$, where $C_1$ is a j-concept and $C_2$ is an i-concept. According to the properties (5) and (8) of the distributed tableau, there is   a $y$ in $S_j$, s.t. $(x,y) \in a_{ij}(P)$, and $j:C_1 \in \mathcal{L}_j(y)$. According to property (4) of the distributed tableau, since $(x,y) \in a_{ij}(P)$, it holds that $i:C_2 \in \mathcal{L}_j(y)$. Therefore the label of $y$ in the $j$ unit has i-concepts and j-concepts. According to the correspondence between the model and the distributed tableau, and by induction it holds that  $y \in (C_1)^{\mathcal{I}_i} \cap (C_2)^{\mathcal{I}_j}$. According to property (11) of the distributed tableau, there is an individual $y'$ in the $i$-th unit, such that $y'=\pi_{ji}(y)$ and $\mathcal{L}_{i}(y')\supseteq \mathcal{L}_{-j}(y) \supseteq \{i:C_2\}$. According to  the correspondence between the model and the distributed tableau, $y \in r_{ij}(y')$. 

(b) It must also be noticed that according to the definition of the distributed tableau, the label of each individual $x$ in unit  $i$ includes all concept correspondences for that unit. Each correspondence from the point of view of $i$ forces a disjunction of the type $i:C_1 \sqcup j:C_2$ in the label of $x$. According to property (3) of the distributed tableau, either $i:C_1$ or $j:C_2$ are in $\mathcal{L}_i(x)$. According to the correspondence between the model and the distributed tableau, $x \in C_1^{\mathcal{I}_i}$ or $x \in C_2^{\mathcal{I}_j}$. The first case where $x \in C_1^{\mathcal{I}_i}$ is addressed locally. In case $x \in C_2^{\mathcal{I}_j}$, according to property (11) of the distributed tableau, there is an individual $x'$ in the $j$-th unit, such that $x'=\pi_{ij}(x)$ and $\mathcal{L}_{j}(x') \supseteq \mathcal{L}_{-i}(x) \supseteq \{j:C_2\}$. According to the correspondence between the model and the distributed tableau, $x \in r_{ji}(x')$. 
\\

For the converse, if $\mathfrak{I}=\langle (\mathcal{I}_{ij})_{i,j \in \text{I}}, (r_{ij})_{i\neq j \in \text{I}} \rangle$ is a model for $D$ w.r.t. $\Sigma$, then the distributed tableau can be defined as follows:
\begin{itemize}
\item $S_i = \Delta_i$
\item $a_{ij}(E) = E^{\mathcal{I}_{ij}} \cup \{ (x,y) | (x,z) \in E^{\mathcal{I}_{i}} \text{ and } z \in r_{ji}(y) \text{, where } E \in \mathcal{E}_{ij} \cap N_{R_i}\}$, 
\item $a_{i}(E) = E^{\mathcal{I}_{i}}$, where $E \in N_{R_i}$, 
\item $\mathcal{L}_i(s) = \{ C \in \bigcup_{k \in N(i)} cl_k(D,\Sigma) |$ s.t. one of the following properties hold: $s \in C^{\mathcal{I}_{i}}$, or  $x \in r_{ik}(s)$ and $m:C \in \mathcal{L}_k(x)$\}.  \\
It must be pointed out that $C$ can be a  concept in any unit $m \in I$.
\item $x= \pi_{ji}(s)$ if  $s \in r_{ij}(x)$ 
\end{itemize}

It is straightforward that the j-projection specified according to the above specification satisfies the definition of projections, given the way labels are constructed (i.e. in case $x \in r_{ij}(s)$, then it holds that  $\pi_{ji}(x) = s$ given that, according to the way labels are constructed, $\mathcal{L}_{-j}(x) \subseteq \mathcal{L}_{i}(s)$, for $j \in N(i)$.

To prove that the above correspondence between the model and the tableau meets all the conditions required by the distributed tableau, we need to consider the additional cases for $E-\mathcal{SHIQ}$ concerning the treatment of properties that are specified both as roles and as link relations between units, as well as, the treatment of inter-unit concept correspondences:  

(a) Considering the case where $P$  belongs to $N_{R_i} \cap \mathcal{E}_{ij}$ and either (i) there is an existential or at-least restriction to  the i-role $P$ (denoted by $\langle Restr \rangle P$) and a value restriction to the link relation $P$, or (ii) there is an existential or at-least restriction (also denoted by $\langle Restr \rangle P$)  to  the link relation $P$ and a value restriction to the role $P$.

We will prove that the property (4) for the distributed tableau holds for cases (a.i) and (a.ii). 

In the case (a.i) it holds that $x \in (\forall P.C_2)^{\mathcal{I}_{i}} \cap (\langle Restr \rangle P.C_1)^{\mathcal{I}_{i}}$ 
where $C_1$  is an i-concept, $C_2$ is an j-concept, $(x,y) \in P^{\mathcal{I}_{i}}$ and $y \in (C_1)^{\mathcal{I}_i}$. 
According to the correspondence specified, $(x,y) \in a_{i}(P)$ and $y \in S_i$. By the semantics, $y \in (C_1)^{\mathcal{I}_i}\cap (C_2)^{\mathcal{I}_j}$ and thus, there is a $y' \in S_j$, s.t. $y \in r_{ji}(y')$ and $y' \in (C_2)^{\mathcal{I}_j}$. By the construction of the labels ($4^{th}$ bullet) $C_2 \in \mathcal{L}_j(y')$ and  $C_2 \in \mathcal{L}_{i}(y)$. Thus, $C_2 \in \mathcal{L}_{-i}(y) \cap \mathcal{L}_j(y')$ and $y' = \pi_{ij}(y)$.

In the case (a.ii), $x \in (\forall P.C_1)^{\mathcal{I}_{i}} \cap (\langle Restr \rangle P.C_2)^{\mathcal{I}_{i}}$ 
where $C_1$  is an i-concept, $C_2$ is a j-concept, $(x,y) \in P^{\mathcal{I}_{ij}}$ and $y \in (C_2)^{\mathcal{I}_j}$. According to the correspondence specified $(x,y) \in a_{ij}(P)$. By the semantics, $y \in (C_1)^{\mathcal{I}_i} \cap (C_2)^{\mathcal{I}_j}$ and thus, there is a $y' \in S_i$, s.t. $y \in r_{ij}(y')$ and $y' \in (C_1)^{\mathcal{I}_i}$. Thus, by the way labels are constructed (third bullet), $C_1 \in \mathcal{L}_i(y')$ and $C_1 \in \mathcal{L}_{j}(y)$. Thus, $C_1 \in \mathcal{L}_{-j}(y) \cap \mathcal{L}_i(y')$ and $y' = \pi_{ji}(y)$.
\\
(b) Property (11): Given an individual $x \in S_i$ such that $\mathcal{L}_{ij}(x) \neq \emptyset$, then by the correspondence specified, $x \in (\sqcap_{C \in \mathcal{L}_{ij}(x)} C)^{\mathcal{I}_j}$. By the semantics, there is a $x' \in \Delta_j$ s.t. $x \in r_{ji}(x')$ and $x' \in (\sqcap_{C \in \mathcal{L}_{ij}(x)} C)^{\mathcal{I}_j}$. Then by the definition of the $j$-projection and the correspondence specified, $x'=\pi_{ij}(x)$. Since domain relations are globally one-to-one, $x'$ cannot be the $j$-projection of any individual different from $i:x$, and in case there is another $j:x''$ individual that satisfies the $j$-projection conditions for $i:x$, then this cannot be different from $j:x'$, according to $i$, i.e. from the subjective point of view of $i$, $x''=_i x'=\pi_{ij}(x)$ $\Box$

\subsection{A Tableau Algorithm for $E-\mathcal{SHIQ}$}

In this section we present a distributed tableau algorithm that, given a concept $X$ in NNF, the concepts $C_{K_{kj}}$, for any $j,k \in I$ (defined in Def. 7), and the property boxes $\mathcal{R}_{kj}$ for each $j,k \in I$, it constructs a \textit{distributed completion graph} deciding about the satisfiability of $X$ w.r.t. the distributed knowledge base $\Sigma$.
\\

\textbf{Definition 8 }(Distributed completion graph)
A \textit{distributed completion graph} for a concept $X$ w.r.t. a distributed knowledge base $\Sigma$, is a directed graph \textbf{G}$=(V,U,\mathcal{L},\neq, \overset {=} \mapsto)$. Each i-node $x \in V$ is labelled with a set $\mathcal{L}_i(x) \subseteq (\bigcup_{m \in N(i)} (cl_m(D,\Sigma) \cup C_{\mathcal{K}_{mj}})$, 
and each edge $\langle \textit{x,y} \rangle \in U$ is labelled with  $\mathcal{L}(\langle \textit{x,y} \rangle) \subseteq (\mathcal{E}_{ij} \cup N_{R_i})$, where $i,j,m \in I$. 
\\

For an edge $\langle \textit{i:x},\textit{j:y} \rangle \in U$, \textit{j:y} is called a $\textit{j-}successor$ of \textit{i:x}, while \textit{i:x} is an $\textit{i-}predecessor$ of \textit{j:y}.  If it holds that $R' \in \mathcal{L}(\langle \textit{i:x},\textit{j:y} \rangle)$ for any property with $R' \sqsubseteq^*_{\mathcal{R}} R$ 
, it holds that \textit{j:y} is an $R$\textit{-j-}successor of $x$, and \textit{i:x} is an $R$\textit{-i-}predecessor of $y$. $Ancestor$ is the transitive closure of predecessor. A node \textit{i:y} is a $neighbor$ ($R-neighbor$) of a node \textit{i:x}, if \textit{i:y} is an i-successor (R\textit{-i-}successor) of \textit{i:x} or \textit{i:x} is an i-successor ($Inv(R)\textit{-i-}successor$) of \textit{i:y}.  Two $i$ nodes can be neighbors only if they are related with an i-role.  

\begin{table}
\centering
\caption{Expansion rules.}
\resizebox{1.0\textwidth}{!}{
\begin{tabular}{|p{1.85cm}|p{19.8cm} |} \hline
\multicolumn{1}{|c}{Rule Name} & \multicolumn{1}{|c|}{Rule}\\ \hline
$\sqcap$\textit{-rule} & if $C_1 \sqcap C_2 \in \mathcal{L}_i (x)$, $x$ is not blocked, $\{C_1,C_2\}\cap \mathcal{L}_i(x)=\emptyset$ then $\mathcal{L}_i (x) \leftarrow \mathcal{L}_i (x) \cup \{C_1,C_2\}$ \\ 
$\sqcup$\textit{-rule} & if $C_1 \sqcup C_2 \in \mathcal{L}_i(x)$, $x$ is not blocked and $\{C_1,C_2\}\cap \mathcal{L}_i(x)=0$ then $\mathcal{L}_i(x) \leftarrow \mathcal{L}_i(x) \cup \{C\}$\\ 

\textit{CE-rule} & if $C_{\mathcal{K}_{ij}} \not \in \mathcal{L}_i(x)$ then $\mathcal{L}_i(x) \leftarrow \mathcal{L}_i(x) \cup \{C_{\mathcal{K}_{ij}}\}$, where 
$C_{\mathcal{K}_{ij}}=\sqcap_{C \sqsubseteq D \in \mathcal{T}_i}(\neg C \sqcup D)\sqcap_{j:F \overset{\sqsubseteq}{\to} i:E \in \mathfrak{B}_{ij}}(\neg i:E \sqcup j:F)\sqcap_{j:H \overset{\sqsupseteq}{\to} i:G \in \mathfrak{B}_{ij}}(\neg j:H \sqcup i:G)$\\

\textit{TRANS-rule} & if $x$ is an $i$-node and has a $P$-ancestor $y_0$ such that $x,y_0$ are connected by a chain $y_0,...,x$ and 
$P$ is transitive for each $(y_0,y_1),...,(y_n,x)$ then add to the graph $\mathcal{L}_i(\langle y_0,x\rangle)=\{P\}$\\ 
$\forall$\textit{-rule} & if $\forall P_{ij} .C \in \mathcal{L}_i(x)$, $x$ is not indirectly blocked and has $P_{ij}$-successor (or in case $i=j$, a $P_{i}$-neighbour) $y$ with $C\not \in \mathcal{L}_j(y)$ then 
$\mathcal{L}_j(y) \leftarrow \mathcal{L}_j(y) \cup \{C\}$\\ 
$\exists$\textit{-rule} & if $\exists P_{ij} .C \in \mathcal{L}_i(x)$, $x$ is not blocked and has no $P_{ij}$-successor (or in case $i=j$, a $P_{i}$-neighbour) $y$ with $\{C\}\in \mathcal{L}_j(y)$ then 
create a new node with $\mathcal{L}(\langle x,y \rangle)=P_{ij}$ and $\mathcal{L}_j(y)=\{C\}$\\ 
$\geq$\textit{-rule} & if $\geq n S_{ij} .C \in \mathcal{L}_i(x)$, $x$ is not blocked and there are not $n$ distinct $S_{ij}$-successors (or in case $i=j$, a $S_{i}$-neighbours) $y_1,...,y_n$ with $C \in \mathcal{L}_j(y_k)$ then create $n$ new distinct nodes with $\mathcal{L}(\langle x,y_k \rangle)=\{S_{ij}\}, \mathcal{L}_j(y_k)=\{C\}$, $1 \leq k \leq n$ \\
$\leq$\textit{-rule} & if $\leq n S_{ij} .C \in \mathcal{L}_i(x)$, $x$ is not indirectly blocked and has $n+1$ $S_{ij}$-successors (or in case $i=j$, $S_{i}$-neighbours) $y_0,...,y_n$ with $C \in \mathcal{L}_j(y_k)$ for each $0 \leq k \leq n$ and there exists $k\neq l$ s.t. $y_k \neq y_l$ \text{then  set} $\mathcal{L}_j(y_k) \text{ equal to } \mathcal{L}_j(y_k)\cup \mathcal{L}_j(y_l)$, and add $z \neq y_k$ for each $z \neq y_l$.
Remove $y_l$, all corresponding individuals of $y_l$, and all the edges leading to $y_l$. \\
\textit{choose-rule} & if $\{\geq n S_{ij} .C, \leq n S_{ij} . C\}\cap \mathcal{L}_i(x)\neq 0$, $x$ is not blocked and $y$ is a $S_{ij}$-successor (or in case $i=j$, a $S_{i}$-neighbour) of $x$ 
$\{C, \neg C\}\cap \mathcal{L}_j(y)= \emptyset$ then $\mathcal{L}_j(y)=\mathcal{L}_j(y) \cup \{X\}$ for some $X \in \{C, \neg C\}$\\
$\pi$\textit{-rule} & if $\mathcal{L}_{ij}(x) \neq \emptyset, i \neq j,$  $x$ is not blocked and there is no node $j:x'$ such that $j:x' \overset {=} \mapsto i:x$ then, create a node $j:x'$ such that$j:x' \overset {=} \mapsto i:x$ and $(\mathcal{L}_{i}(x)  \cap \mathcal{L}_{j}(x')) = \mathcal{L}_{-i}(x)$.\\
$\pi$\textit{-update-rule} & if $\mathcal{L}_{ij}(x) \neq \emptyset, i \neq j, x$ is not blocked and there is a node $j:x'$ such that $j:x' \overset {=} \mapsto i:x$ and $(\mathcal{L}_{-i}(x)  \cap \mathcal{L}_{j}(x')) \subset\mathcal{L}_{-i}(x)$ \textit{then set} $\mathcal{L}_{j}(x')$ \textit{equal to} $\mathcal{L}_{j}(x') \cup (\mathcal{L}_{-i}(x)-\mathcal{L}_{j}(x'))$.\\
\hline
\end{tabular}}
\label{expansionRules}
\end{table}

The distributed completion graph is initialized with $n=|I|$ nodes $x_k, k \in I$, with $\mathcal{L}_i(x_k)=\{X\}$, if $k=i$, and $\mathcal{L}_i(x_k)=\emptyset$, otherwise. The algorithm expands the graph by repeatedly applying the expansion rules presented in Table \ref{expansionRules}. 
The rules may generate new nodes locally or to graphs of neighbor peers, or merge existing nodes. In any case nodes' labels must be maintained. Having said that, we must notice that generative expansion rules (i.e. $\exists-rule$ and $\geq-rule$) may create new successor nodes for an ij-link relation, $i \neq j \in I$,  either to the completion graph of $i$ or to the graph of a neighbor $j$. Specifically, applying a generative expansion rule to node \textit{i:x} concerning a ij-link relation E, the result may be either (a) a new node in the completion graph of $j$, e.g. $y$, that will be later projected to   $i$, given that $\mathcal{L}_{ji}(x) \neq \emptyset$, or alternatively, (b) a new node $y$ in the completion graph of  $i$, that may be projected to $j$, given that the $ij$-part $\mathcal{L}_{ij}(y)$ of its label is not empty:These are alternative strategies that a tableau algorithm may follow, balancing between the number of messages exchanged between local reasoning chunks for the maintenance of the corresponding nodes' labels, and effectiveness.  

A node is not expanded if it is blocked or in case it contains a clash.
A $clash$ in a node is a contradiction in one of the following forms:
(a)  $\{ \textit{i:C},i:\neg C\} \subseteq \mathcal{L}(x)$ for a concept name  \textit{i:C} and $x \in V$; or  
 (b) $(\leq n R.C)\in \mathcal{L}(x)$, $x\in V$,  and there are at least ($n+1$) $R$-successors $y_0,...,y_n$ of $x$, such that $C \in \mathcal{L}(y_i)$ and $y_i\neq y_j$, for each $0\leq i < j \leq n$. 
It must be pointed out that a clash may occur in any of the peers participating in the construction of the completion. 
The reasoning chunk in peer $i$ can detect clashes of the form (a) in case $C$ is  j-concept,  as well as of form (b), without relying to $j$. 

$Blocking$ guarantees termination of the tableau algorithm. A node $y$ $directly$ $blocks$ $x$ in the following cases: 

(1) Both are i-nodes, $x$ has ancestors $x',y,y'$, none of its ancestors are blocked, and it holds that: 
(i) $x$ is successor of $x'$, 
$y$ successor of $y'$ and 
(ii) $\mathcal{L}_i(x)=\mathcal{L}_i(y),\mathcal{L}_i(x')=\mathcal{L}_i(y')$,
(iii) $\mathcal{L}(\langle x,x' \rangle)=\mathcal{L}(\langle y,y' \rangle)$.


(2) $i:x$ is a projection of $j:x'$ and there is a node $i:y$ s.t $\mathcal{L}_i(x) \subseteq \mathcal{L}_i(y)$. This is a subset blocking condition that is due to a projection from unit $j$ to unit $i$.
 
A node is blocked if either it is blocked directly, or in case one of its predecessors is blocked ($indirect$ $blocking$).

The algorithm returns that a concept $C$ is $satisfiable$ with respect to $\Sigma$, if it results to a \textit{clash-free}  (i.e. none of its nodes contains a clash) and \textit{complete} (i.e. none of the expansion rules is applicable)  completion graph. Otherwise  $C$ is considered \textit{unsatisfiable}  w.r.t. $\Sigma$.

The propagation of the subsumption relation between ontology units is done via the completion of the $\pi-rule$ and $\pi-update-rule$, as it applies to the local chunks  of the distributed completion graph. To show this, let us consider the example in figure 3.
\\

\textbf{Example} (Subsumption propagation). As shown in figure 3,  it holds that  $I=\{2,3,4\}$ and $\Sigma = \langle \textbf{T}, \textbf{R}, \textbf{C} \rangle$, where
\begin{itemize}
\item \textbf{T}=$(\mathcal{T}_2, \mathcal{T}_3, \mathcal{T}_4)$, 
\begin{itemize}
\item $\mathcal{T}_2=\{MedicalConference \sqsubseteq Conference\}$, 
\item $\mathcal{T}_3=\emptyset$, 
\item $\mathcal{T}_4=\emptyset$
\end{itemize}
\item $\mathcal{R}=(R_2, R_3, R_4, (R_{ij})_{i \neq j \in I}$, where each of these sets is equal to $\emptyset$.
\item It also holds that \textbf{C}=$(C_{ij})_{(i \neq j \in I)}$, where
\begin{itemize}
\item $C_{23}=\{2:MedicalConference \overset \sqsupseteq \rightarrow 3:PediatricConference\}$, 
\item $C_{43}=\{4:Event \overset \sqsubseteq \rightarrow 3:HumanActivity\}$, 
\item $C_{42}=\{4:Event \overset \sqsupseteq \rightarrow 2:Conference\}$, 
\item all other sets of correspondences among units are equal to $\emptyset$.
\end{itemize}
\end{itemize}
We need to prove that $PediatricConference \sqsubseteq HumanActivity$. Thus, we must prove that the concept $PediatricConference \sqcap \neg HumanActivity$ is unsatisfiable w.r.t. the distributed knowledge base $\Sigma$. 

The distributed completion graph is initialized with three nodes ${x_i}, i \in \{2,3,4\}$. Considering unit 3,
initially it holds that 

$\mathcal{L}_3(x)= \{PediatricConference, \neg HumanActivity\}$. 

The completion of the $CE-rule$ results to 

$\mathcal{L}_3(x)= \{PediatricConference, \neg HumanActivity, (\neg 4:Event \sqcup 3:HumanActivity), \\ (3: \neg PediatricConference \sqcup 2:MedicalConference)\}$. 

This finally (after the exploration of alternatives in unit 3) results to: 

$\mathcal{L}_3(x)= \{PediatricConference, \neg HumanActivity, \neg 4:Event, 2:MedicalConference\}$.

The completion of the projection $\pi-rule$ from unit 3 to unit 2 will result in the generation of a new node $y$ in 2, s.t. $2:y \overset = \rightarrow 3:x$, i.e. $y=\pi_{32}(x)$ s.t.

$\mathcal{L}_2(y)= \mathcal{L}_{-3}(x)= \{\neg 4:Event, 2:MedicalConference\}$.
  
The completion of the $CE-rule$ for $y$ in 2 results to 

$\mathcal{L}_2(y)= \{\neg 4:Event, 2:MedicalConference, (\neg MedicalConference \sqcup Conference), \\ (\neg Conference \sqcup 4:Event)\}$.

After the exploration of alternatives for the first disjunction in the label $\mathcal{L}_2(y)$, the label becomes,

$\mathcal{L}_2(y)= \{\neg 4:Event, 2:MedicalConference, Conference, (\neg Conference \sqcup 4:Event)\}$.

All the alternative branches for this node result to clashes. Thus, the concept \\ $PediatricConference \sqcap \neg HumanActivity$ is unsatisfiable w.r.t. the distributed knowledge base $\Sigma$. 
\\ 

\textbf{Theorem} (Distributed Satisfiability) 
A concept $X$ is satisfiable w.r.t. the distributed knowledge base $\Sigma$ iff the application of expansion rules create a clash-free and complete distributed completion graph. 

The worst-case complexity of the algorithm is $2NexpTime$ w.r.t the sum of the size of $X$ and $\Sigma$.

\textbf{Proof}
\begin{enumerate}
\item \textbf{Termination}{

Given that each unit $i \in I$, where $I$ is a finite set of indices, has a finite number of units in its neighborhood $N(i)$, the proofs for the following statements are similar to those given for $\mathcal{E}$-connections:
\begin{itemize}
\item \textit{The number of possible concepts in a node label and the number of properties in an edge label are bounded}
\item \textit{The length of a given path is finite, i.e. the depth of the distributed completion graph is bounded}
\item \textit{Concepts are never deleted from node labels}
\item \textit{The out-degree of the distributed completion is finite}
\item \textit{The worst-case complexity of the algorithm is 2NexpTime w.r.t. the sum of the size of X, the distributed TBox and the distributed RBox}
\end{itemize}

The proof for the statement:

\noindent\textit{``Nodes may be deleted (merged or identified with other nodes), but the properties of the distributed tableau are not affected from this fact''}

 is the same to that given for $\mathcal{SHIQ}$ and $\mathcal{E}$-connections, considering also that the $\pi$-rule applies only once in any case.
}
\item{\textbf{Soundness}: If the algorithm yields to a complete and clash-free distributed completion graph for an input $X$ w.r.t. $\Sigma$, then $X$ has a distributed tableau w.r.t $\Sigma$.

Let $T_{comp}$ be a complete and clash-free completion graph. We define a path as a sequence of pairs of nodes in $T_{comp}$ as follows:  $p=[x_0/x'_0,..., x_n/x'_n ]$. For such a path we define $Tail(p) = x_n$ and $Tail'(p)=x'_n$. By $[p|x_{n+1}/x'_{n+1}]$ we denote the path $p=[x_0/x'_0,..., x_n/x'_n, x_{n+1}/x'_{n+1} ]$. 

The sets of paths $Path(T_{comp})$ in the completion tree  is defined by induction:

\begin{itemize}
\item{For the root nodes $x_j$, $j=1,...n$ of $T_{comp}$, $[x_j/x_j] \in Path(T_{comp})$}
\item{for a path $p \in path(T_{comp})$, and a node $z$ in $T_{comp}$
\begin{itemize}
\item{if $z$ is a (role or link) successor of $Tail(p)$, $z$ is not blocked, then $[p|z/z] \in Path(T_{comp})$}
\item{if for some i-node $y$ in $T_{comp}$, $y$ is a (role or link) successor of $Tail(p)$, $z$ blocks $y$, then $[p|z/y] \in Path(T_{comp})$}
\item{if  $Tail(p)$ is not blocked, $y \overset {=} \mapsto Tail(p)$ and $y$ is not blocked, then $[p|y/y] \in Path(T_{comp})$}
\item{if  $Tail(p)$ is not blocked, $y \overset {=} \mapsto Tail(p)$ and $z$ blocks $y$, then $[p|z/y] \in Path(T_{comp})$}
\end{itemize}} 
\end{itemize}

To make the proof more succinct, in the following we do not distinguish the last two cases. Indeed, we consider that in case $Tail(p)$ is not blocked, and $y \overset {=} \mapsto Tail(p)$, then $[p|z/y] \in Path(T_{comp})$, where it holds that either $z=y$, or $z$ blocks $y$. In any case it holds that   $\mathcal{L}_i(y) \subseteq \mathcal{L}_i(z)$, where $i \in I$ is the unit of $z$ and $y$.

Due to the construction of a path, if $p \in Path(T_{comp})$ with $p=[p'|x/x']$, $x$ is never blocked. It also holds that in case $Tail(p)=x$ is an i-node and $x'$ has no link predecessors, $x'$ is pairwise blocked iff $x'\neq x$, and due to the definition of indirect blocking $x'$ is never indirectly blocked. It holds that $\mathcal{L}_i(x)=\mathcal{L}_i(x')$. 

The distributed tableau $T=\{{\langle S_i, \mathcal{L}_i, (a_{ij}), (\pi_{ij})\rangle}\}, i,j \in I$, for a concept $X$ and a distributed knowledge base $\Sigma$ is defined as follows:

\begin{itemize}
\item{$S_i$=\{$p | p \in$  Path($T_{comp}$) and $Tail(p)$ is an i-node \}}
\item{$\mathcal{L}_i(p) = \mathcal{L}_i(Tail(p))$ }
\item{$a_{ij}(E) = \{(p,q) \in S_i \times S_j $ such that $x'$ is an j-node, $q=[p|x/x']$ and it holds either that (a) $x'$ is an E-$j$-successor of $Tail(p)$, or (b) $p=[p'|z/y]$ and $j:x' \overset {=} \mapsto i:y$, $y$ is an E-$i$-successor of Tail(p) and either $z=y$ or $z$ blocks $y$ via subset blocking\}}.
\item{$\pi_{ij}(p) =q$, s.t  $p \in S_i, q \in S_j, q=[p|x/x]$ and $j:x \overset {=} \mapsto i:Tail(p)$}
\end{itemize}

It has to be noticed that, as it is defined, $\pi_{ij}$ satisfies the definition for projections: Indeed, given that $j:Tail(q) \overset {=} \mapsto i:Tail(p)$, according to the completion of the $\pi-rule$ and the $\pi-update-rule$ it holds that $\mathcal{L}_{-i}(Tail(p)) \subseteq \mathcal{L}_{j}(Tail(q))$. Therefore, $Tail(q)=\pi_{ij}(Tail(p))$, and according to the mapping above, $\mathcal{L}_{-i}(p) \subseteq \mathcal{L}_{j}(q)$, and thus, $q=\pi_{ij}(p)$. Also, it must be noticed that $i$ considers that $q$ is the one and only j-projection of $p$, given that any other $q'=[p|z/y]$ with $q'= \pi_{ij}(p)$ cannot be different from $q$ from the subjective point of view of $i$.  

It has to be proved that T satisfies all the conditions required to a distributed tableau.

It is easy to check that according to the translation above (from $T_{comp}$ nodes to paths), the initialization conditions hold:
\begin{itemize}
\item{There is an individual $s_i \in S_i$ s.t $X \in \mathcal{L}_i(s_i)$: This condition is trivial, since  $T_{comp}$ is initialized with an i-node $x$ in the completion s.t. $X \in \mathcal{L}_i(x)$. Hence $p=[x/x]$ is a path in $T_{comp}$, and ${X} \in \mathcal{L}_i(p)$ using the translation.}
\item{There are $n$ individuals $p_j, j=1,...,n$, such that $p_j \in S_j$: In the initialization step of the algorithm, $n$ individuals $x_j, j=1...,n$ are generated in $T_{comp}$. Then $p_j=[x_j/x_j] \in Path(T_{comp})$ and since $Tail(p_j)=x_j$ is a j-node in $T_{comp}$, then $p_j \in S_j$.}
\item{$\forall p \in S_i, i=1,...,n$ it holds that  $C_{K_{ij}} \in \mathcal{L}_i(p)$: Using the translation we have that if $p \in S_i$, then it corresponds to a path in $T_{comp}$ s.t. $Tail(p)$ is an i-node. It then holds that $C_{K_{ij}} \in \mathcal{L}_i(Tail(p))$ as a consequence of the completeness of the $CE$-rule. Therefore, $C_{K_{ij}} \in \mathcal{L}_i(p)$.}
\end{itemize}

We show that T satisfies the remaining properties required to a distributed tableau. We concentrate on ij-link relations  in $\mathcal{E}_{ij}$ ($i \neq j$), and concepts' correspondences.

\begin{itemize}

\item{Let $\exists E.C \in \mathcal{L}_i(p)$, where $p \in S_i$ and $E \in \mathcal{E}_{ij}$ and $C$ a j-concept. It has to be proved that there exists some $q \in S_j$, s.t. $(p,q) \in a_{ij}(E)$ and $C \in \mathcal{L}_j(q)$.  

Using the translation, $\exists E.C \in \mathcal{L}_i(Tail(p))$. As $Tail(p)$ cannot be blocked by definition of paths, the $\exists$-rule ensures that there exists an E-$j$-successor $x$ of $Tail(p)$ in $T_{comp}$, with $C \in \mathcal{L}_j(x)$. There are two possibilities: 
	\begin{itemize}
		\item{First, if $q=[p|x/x] \in Path$, then using the translation, $C \in  \mathcal{L}_j(Tail(q))= \mathcal{L}_j(q)$.}
		\item{Second, if x is blocked by z, then $q=[p|z/x] \in Path$. Since $C \in \mathcal{L}_j(x)$, and by the blocking condition $\mathcal{L}_j(x) \subseteq \mathcal{L}_j(z)$, then  $C \in \mathcal{L}_j(z)$. As $z=Tail(q)$, then  $C \in \mathcal{L}_j(Tail(q))$, and  $C \in \mathcal{L}_j(q)$. }
	\end{itemize} 
In both cases it holds that since $x$ is an E-$j$-successor of $Tail(p)$, then $(p,q) \in a_{ij}$. }

\item{Let $\forall E.C \in \mathcal{L}_i(p)$ and $(p,q) \in a_{ij}(E)$ where $E \in \mathcal{E}_{ij}$ and $C$ a j-concept. It has to be proved that  $C \in \mathcal{L}_j(q)$. 

Using the translation, it holds that $\forall E.C \in \mathcal{L}_i(Tail(p))$. We distinguish the following cases:
	
	\begin{itemize}
		\item{Let $q=[p|z/x]$ and $x$ be a j-node, $p=[p'|y/y]$, $y$ an E-$i$-successor of $Tail(p')$, s.t. $j:x \overset {=} \mapsto i:y $, and either \textit{j:z=j:x} or \textit{j:z} blocks \textit{j:x} .  By definition of paths, $y$ is not blocked. Thus, $Tail(p)$ is not blocked, and the $\forall-rule$ ensures that $C \in \mathcal{L}_i(y)$. Also, the $\pi-rule$ and the $\pi-update-rule$ ensure that $C \in \mathcal{L}_j(x)$ and that $\mathcal{L}_{-i}(y) \subseteq \mathcal{L}_{j}(x)$, thus $x=\pi_{ij}(y)$. Since $z=Tail(q)$, and in any case $\mathcal{L}_j(x) \subseteq \mathcal{L}_j(z)$, it holds that $C \in \mathcal{L}_j(Tail(q))$, thus $C \in \mathcal{L}_j(q)$.}
		\item{If $q=[p|x/x]$ and $x$ is an E-$j$-successor of $Tail(p)$, then by the definition of Paths, $Tail(p)$ is not blocked, and the $\forall-rule$ ensures that $C \in \mathcal{L}_j(x)$. Since $x=Tail(q)$, it holds that $C \in \mathcal{L}_j(Tail(q))$, thus $C \in \mathcal{L}_j(q)$}.
		\item{If $q=[p|x/y]$ and $y$ is an E-$j$-successor of $Tail(p)$ and $x$ blocks $y$, $Tail(p)$ by the definition of paths is not blocked, and the $\forall-rule$ ensures that $C \in \mathcal{L}_j(y)$. Since, by the blocking condition, $ \mathcal{L}_j(y) \subseteq \mathcal{L}_j(x)$, it holds that $C \in \mathcal{L}_j(x)$.  As $x=Tail(q)$, $C \in \mathcal{L}_j(Tail(q))$, thus $C \in \mathcal{L}_j(q)$.}
	\end{itemize} }

\item{Let $(\geq n E.C) \in \mathcal{L}_i(p), E \in \mathcal{E}_{ij}$ and $C$ a j-concept. It has to be proved that $\sharp E^{T}(p,C)= \sharp \{q \in S_j | (p,q) \in a_{ij}(E)$, and $C \in L_j(q)\} \geq n$. 

Using the translation, $(\geq n E.C) \in \mathcal{L}_i(Tail(p))$. Due to the definition of paths, $Tail(p)$ can not be blocked. By completeness of the $\geq - rule$ there are $n$ different individuals $y_k, k=1,...n$ in $T_{comp}$ such that each $y_k$ is an E-$j$-successor of $Tail(p)$ and $C \in \mathcal{L}_j(y_k)$ and $y_k \neq y_l \forall k,l, s.t. 1 \leq k < l \leq n$. It has to be proved that for each of these individuals there exists a path $q_k$ such that $(p,q_k) \in a_{ij}(E)$ and $C \in \mathcal{L}_j(q_k)$ and $q_k \neq q_l \forall k,l, 1 \leq k < l \leq n$. Let $x=Tail(p)$. 

For each $y_k$ there are three possibilities:
	\begin{itemize}
		\item{$q_k=[p|z_k/y_k]$ s.t. $p=[p'|y'_k/y'_k]$ and $j:y_k \overset {=} \mapsto i:y'_k$, $y'_k$ is an E-$i$-successor of Tail(p'), and either \textit{j:z}$_k$=\textit{j:y}$_k$ or \textit{j:z}$_k$ blocks \textit{j:y}$_k$.  By definition of paths $Tail(p)$ is not blocked, and the $\geq-rule$ ensures that $C \in \mathcal{L}_i(y'_k)$. Also, the $\pi-rule$ and the $\pi-update-rule$ ensure that $C \in \mathcal{L}_j(y_k)$  and that $\mathcal{L}_{-i}(y'_k) \subseteq\mathcal{L}_{j}(y_k)$, thus $y_k=\pi_{ij}(y'_k)$. 
		Since $z_k=Tail(q)$, and in any case $\mathcal{L}_j(y_k) \subseteq \mathcal{L}_j(z_k)$, it holds that $C \in \mathcal{L}_j(Tail(q))$, thus $C \in \mathcal{L}_j(q)$.}
		\item{$q_k=[p|y_k/y_k]$ and $y_k$ is an E-$j$-successor of $Tail(p)$. As $y_k$ is an E-successor of $Tail(p)$ and by the definition of Paths, $Tail(p)$ is not blocked, then the $\geq-rule$ ensures that $C \in \mathcal{L}_j(y_k)$. As $y_k=Tail(q_k)$, $C \in \mathcal{L}_j(Tail(q_k))$, thus $C \in \mathcal{L}_j(q_k)$}
		\item{$q_k=[p|z/y_k]$ and $y_k$ is an E-$j$-successor of $Tail(p)$ and $z$ blocks $y_k$. $Tail(p)$, by the definition of Paths, is not blocked, and the $\geq-rule$ ensures that $C \in \mathcal{L}_j(y_k)$. Since, by the blocking condition, $ \mathcal{L}_j(y_k) \subseteq \mathcal{L}_j(z)$, $C \in \mathcal{L}_j(z)$.  As $z=Tail(q)$, $C \in \mathcal{L}_j(Tail(q))$, thus $C \in \mathcal{L}_j(q)$.}
	\end{itemize} 
}

Obviously, for any of these cases it holds that $(p,q_k) \in a_{ij}(E)$ and $q_k \neq q_l \forall k,l, 1 \leq k < l \leq n$.

\item{Let $\{(\geq n E.C), (\leq n E.C)\} \cap \mathcal{L}_i(p) \neq \emptyset$, $E \in \mathcal{E}_{ij}$, $C$ a j-concept and $(p,q) \in a_{ij}(E)$. It is proved that $\{C, \neg C \} \cap \mathcal{L}_j(q) \neq \emptyset$. 

Using the translation $\{(\geq n E.C), (\leq n E.C) \} \cap \mathcal{L}_i(Tail(p)) \neq \emptyset$. There are two possibilities:

	\begin{itemize}
		\item{$q=[p|x/x']$ and $x'$ is an E-$j$-successor of $Tail(p)$. By definition of paths, $Tail(p)$ can not be blocked; hence the $choose-rule$ ensures that $\{C, \neg C \} \cap \mathcal{L}_j(x') \neq \emptyset$. If $x'$ is not blocked, then $x=x'$. As $x=Tail(q)$ then $\{C, \neg C \} \cap \mathcal{L}_j(Tail(q)) \neq \emptyset$. Thus, according to the translation $\{C, \neg C \} \cap \mathcal{L}_j(q) \neq \emptyset$. If $x$ blocks $x'$, by definition of blocking $ \mathcal{L}_j(x') \subseteq \mathcal{L}_j(x)$.  As $x=Tail(q)$, $\{C, \neg C \} \cap \mathcal{L}_j(Tail(q)) \neq \emptyset$, thus $\{C, \neg C \} \cap \mathcal{L}_j(q) \neq \emptyset$.}
		
		\item{$q=[p|z/x]$ and $p=[p'|y/y]$ s.t. $j:x \overset {=} \mapsto i:y$, $y$ is an E-$i$-successor of $Tail(p')$, and either \textit{j:z=j:x} or \textit{j:z} blocks \textit{j:x}.  By definition Tail(p) is not blocked, and the $choose-rule$ ensures that $\{C, \neg C \} \cap \mathcal{L}_i(y) \neq \emptyset$. Also, the $\pi-rule$ and the $\pi-update-rule$ ensure that $\{C, \neg C \} \cap \mathcal{L}_j(x) \neq \emptyset$ and $\mathcal{L}_{-i}(y) \subseteq \mathcal{L}_{j}(x)$, thus $x=\pi_{ij}(y)$.
		Since $z=Tail(q)$, and in any case $\mathcal{L}_j(x) \subseteq \mathcal{L}_j(z)$, it holds that $\{C, \neg C \} \cap \mathcal{L}_j(Tail(q)) \neq \emptyset $, thus $\{C, \neg C \} \cap  \mathcal{L}_j(q) \neq \emptyset$. }		\end{itemize} 
}

\item{Let $(\leq n E.C) \in \mathcal{L}_i(p), E \in \mathcal{E}_{ij}$ and $C$ a j-concept. It has to be proved that $\sharp E^{T}(p,C) = \sharp \{q \in S_j | (p,q) \in a_{ij}(E)$, and $C \in L_j(q)\} \leq n$. 

Let us assume that the $(\leq n E.C)$ property in the tableau is violated. Hence, there is some $p \in S_i$ with $(\leq n E.C) \in \mathcal{L}_i(p)$ and $\sharp E^{T}(p,C) > n$. We show that this implies $\sharp E^{T_{comp}}(Tail(p),C) > n$ in contradiction with the clash-freeness or completeness in the distributed completion graph $T_{comp}$. $E^{T}(p,C)$ contains only paths of the form $q_k=[p|y/y']$ or of the form $q_k=[p|z/y]$, where $y$ is a corresponding individual in another unit, and either \textit{z=y} or \textit{z} blocks \textit{y}. If we show that $Tail'$ is injective on $E^{T}(p,C)$, then  $\sharp E^{T_{comp}}(Tail(p),C) = \sharp Tail'(E^{T}(p,C))) $. To show this, assume that there are two paths $q_1$ and $q_2 \in E^{T}(p,C)$ with $Tail'(q_1)=Tail'(q_2)$. In case $Tail'(q_1)=Tail'(q_2)$ and either it is directly blocked or not, then $Tail(q_1) = Tail(q_2)$.
Also, for each $y' \in E^{T_{comp}}(Tail(p),C)$, $y'$ is either a link E-$j$-successor of $Tail(p)$ or a corresponding node of an E-$i$-successor. In either case $C \in \mathcal{L}_i(y')$. In case $y'$ is the i-projection of another node, then $C \in \mathcal{L}_i(y') \subseteq \mathcal{L}_i(z)$. This implies   $\sharp E^{T_{comp}}(Tail(p),C) > n$.}

\item{Let $(p,q) \in a_{ij}(E_1), E_1, E_2 \in \mathcal{E}_{ij}$ and $E_1 \sqsubseteq^* E_2 $. The fact that  $(p,q) \in a_{ij}(E_2)$ is a consequence of the definition of  E-successor that takes into account the properties hierarchy.}

\item{Finally, if for the the j-part of $\mathcal{L}_{i}(p)$ it holds that $\mathcal{L}_{ij}(p) \neq \emptyset, i \neq j$, then we will prove that there is an path $q=[p|x/x]$, s.t. $(\mathcal{L}_{i}(p)  \cap \mathcal{L}_{j}(q)) \supseteq \mathcal{L}_{-i}(p) \neq \emptyset$. 

Since $\mathcal{L}_{ij}(p) \neq \emptyset, i \neq j$, given that $y=Tail(p)$, it holds that $\mathcal{L}_{ij}(y) \neq \emptyset$. The completeness of the $\pi-rule$ assures that there is a j-node x, s.t.  $x=\pi_{ij}(y)$. Thus, $j:x \overset {=} \mapsto i:y$ and $(\mathcal{L}_{i}(y)  \cap \mathcal{L}_{j}(x)) \supseteq \mathcal{L}_{-i}(y)$. In case $y$ is not blocked, then there is a path $q=[p|z/x]$ s.t. either \textit{j:z=j:x} or \textit{j:z} blocks \textit{j:x}. Since $z=Tail(q)$, $y=Tail(p)$, and  $\mathcal{L}_{j}(x)) \subseteq \mathcal{L}_{j}(z))$, $(\mathcal{L}_{i}(Tail(p))  \cap \mathcal{L}_{j}(Tail(q))) \supseteq (\mathcal{L}_{i}(Tail(p))  \cap \mathcal{L}_{j}(Tail'(q))) \supseteq \mathcal{L}_{-i}(Tail(p))$, thus $(\mathcal{L}_{i}(p)  \cap \mathcal{L}_{j}(q)) = \mathcal{L}_{-i}(p)$}.

\end{itemize}
}

\item{\textbf{Completeness:} If the i-concept $X$ has a distributed tableau w.r.t. $\Sigma$, then the expansion rules of the algorithm can be applied in such a way that yield to a complete clash-free distributed completion graph.

Let T=$\{ \langle S_i, \mathcal{L}_i, (a_{ij}), (\pi_{ij}) \rangle\}, i,j \in \text{I}$ a distributed tableau for $X$, $\Sigma$. We use T to guide the application of the non-deterministic rules. To do this we define a function $\mu$ that maps nodes of the completion graph to elements of the tableau:  
\\
$\mu$ : i-nodes in $T_{comp} \mapsto$ elements in $S_i$
\\
so that for each $x$, $y$ in $T_{comp}$ the following hold:

\begin{enumerate}
\item{For each $\mu(x) \in S_i, \mathcal{L}_i(x) \subseteq \mathcal{L}_i(\mu(x)) $}
\item{If $j:y \overset {=} \mapsto i:x$, then $\mu(y)=\pi_{ij}(\mu(x))$, where $x$ is an i-node and $y$ is a j-node, $i \neq j \in \text{I}$}
\item{If $y$ is an E-neighbor of $x$, $E \in N_{R_i}$, then $(\mu(x), \mu(y)) \in a_{i}(E)$, where $x,y$ are i-nodes, $i \in \text{I}$}
\item{If $y$ is an E-$j$-successor of $x$ or $z$ is an E-$i$-successor of $x$ and $j:y \overset {=} \mapsto i:z$ (thus, $\mu(y) = \pi_{ij}(\mu(z))$), then $(\mu(x), \mu(y)) \in a_{ij}(E)$, where $x$ is an i-node and $y$ is a j-node, $i \neq j \in \text{I}$}
\item{$x \neq y$ implies $\mu(x) \neq \mu(y)$}
\end{enumerate} 

It has to be proved that if a rule is applicable to $T_{comp}$ during the execution of the algorithm, then the application of the rule results to a $T'_{comp}$ and the extension of $\mu$ satisfies the conditions stated above.  The proof for the individual non-deterministic rules is the same as the one reported in $\mathcal{E}$-connections. The interesting cases here concern, the ``interactions'' between roles and links, as well as projections. These cases are as follows:

\begin{itemize}
\item{There is a property $E$ which belongs to $N_{R_i} \cap \mathcal{E}_{ij}$ and either (i) there is an existential or at-least restriction to  the role $E$ (denoted by $\langle Restr \rangle$E) and a value restriction to the link-relation $E$, or (ii) there is an existential or at-least restriction (also denoted by $\langle Restr \rangle$E)  to  the link-relation $E$ and a value restriction to the role $E$.

\begin{itemize}

	\item{In the case (i) above, $\langle Restr \rangle E.C_1 \in \mathcal{L}_i(x)$, where $C_1$ is an i-concept. 
	Let that be $(\exists E.C_1) \in \mathcal{L}_i(x)$  \footnote{We show these cases for the $\exists$ restriction. The proofs for the $\geq$ restriction are similar} where $E \in N_{R_i}$ and $x$ a node in $T_{comp}$. Also, $\forall E.C_2 \in \mathcal{L}_i(x)$, where $C_2$ is a j-concept, and $E \in \mathcal{E}_{ij}$. 
	
	The application of the $\exists - rule$ generates a new i-node $y$ in $T_{comp}$ such that $C_1 \in \mathcal{L}_i(y)$ and $(x,y)$ is labelled with $E$. For this individual rule it has been shown that the resulted completion satisfies the conditions for $\mu$ \cite{Grau04tableau}. Thus, there is a $\mu(y)$ in $S_i$ such that $(\mu(x), \mu(y)) \in a_{i}(E)$. 
	
	Also, the application of the  $\forall - rule$ results in extending the label of $y$ in $T_{comp}$ such that $C_2 \in \mathcal{L}_i(y)$, where $C_2$ is a j-concept and $(\mu(x), \mu(y)) \in a_{i}(E)$. 
	
	The application of the $\pi-rule$ to $y$ will create a $y'$ in $j$ s.t. $j:y' \overset {=} \mapsto i:y$ and  $\mathcal{L}_{ij}(y) \subseteq \mathcal{L}_{-i}(y) \subseteq \mathcal{L}_{j}(y')$. Thus, $C_2 \in \mathcal{L}_j(y')$. 
	
	Given that $\mathcal{L}_{ij}(y) \neq \emptyset$, by the definition of the tableau there is a $t$ such that $\pi_{ij}(\mu(y))=t$.  Thus, $C_2 \in \mathcal{L}_j(t)$. If we replace $t$ with $\mu(y')$, then the conditions of $\mu$ are satisfied.
	 }

	\item{In the case (ii) above, $\langle Restr \rangle E.C_1 \in \mathcal{L}_i(x)$, where $C_1$ is a j-concept. 
	Let $(\exists E.C_1) \in \mathcal{L}_i(x)$  where $E \in \mathcal{E}_{ij}$ and $x$ a node in $T_{comp}$. Also, $\forall E.C_2 \in \mathcal{L}_i(x)$, where $C_2$ is a i-concept, and $E \in N_{R_{i}}$. 
		
	The application of the $\exists - rule$ generates a new j-node $y$ in $T_{comp}$ such that $C_1 \in \mathcal{L}_j(y)$ and $(x,y)$ is labelled with $E$. For this individual rule it has been shown that the resulted completion tree satisfies the conditions for $\mu$. Thus, there is a $\mu(y)$ in $S_j$ such that $(\mu(x), \mu(y)) \in a_{ij}(E)$. 
	
	The application of the $\pi-rule$ to $y$ will create a $y'$ in $i$ s.t. $i:y' \overset {=} \mapsto j:y$ and  $\mathcal{L}_{-j}(y) \subseteq \mathcal{L}_{i}(y')$. Thus, $C_2 \in \mathcal{L}_i(y')$. 
	
	By the definition of the tableau, there is an i-node $t$ such that $\pi_{ji}(\mu(y))=t$, since $\mathcal{L}_{ji}(y) \neq \emptyset$.  Thus, $C_2 \in \mathcal{L}_i(t)$. If we replace $t$ by $\mu(y')$, then the conditions of $\mu$ are satisfied.
	}

\end{itemize}}

\item{Regarding the $\leq-rule$ also in relation to projections, if $(\leq nE.C) \in \mathcal{L}_i(x)$, $E \in \mathcal{E}_{ij}$, then  $(\leq nE.C) \in \mathcal{L}_i(\mu(x))$. Since T is a distributed tableau, it holds that $\sharp E^{T}(\mu(x),C) \leq n$. If the $\leq-rule$ is applicable in the completion graph, then it holds that $\sharp E^{T_{comp}}(x,C) > n$. Thus, there are at least $n+1$ E-successors $y_0,...,y_n$ of $x$ such that $C \in \mathcal{L}_i(y_k)$. There must also be two nodes $y,z \in \{y_0,...,y_n\}$ such that $\mu(y)=\mu(z)$. In this case it  cannot hold that $y \neq z$, because of the conditions imposed to $\mu$. 
Given two j-nodes $y'$ and $z'$, s.t. $j:y' \overset {=} \mapsto i:y$, $j:z' \overset {=} \mapsto i:z$, it also holds that  $\mu(y')=\pi(\mu(y))$ and $\mu(z')=\pi(\mu(z))$.  Since $\mu(y)=\mu(z)$, it is also true by the definition of projection that according to $i$, $\pi(\mu(y))=\pi(\mu(z))$. This implies $\mu(y') = \mu(x')$. Thus, $y' \neq x'$ cannot hold from the subjective point of $i$. Thus, the $\leq-rule$ does not violate any conditions.
}

\item{It must  be noticed that according to the definition of the construction of the completion graph, the label of each individual $x$ in a unit $i$ includes all concept-to-concept correspondences that this unit subjectively holds. Each concept-to-concept correspondence is treated as a subsumption by the unit that holds it (according to the semantics), forcing a disjunction of the type $\textit{i:C}_1 \sqcup \textit{j:C}_2$ in the label of \textit{i:x}. Since T is a combined tableau,  $\{\textit{i:C}_1, \textit{j:C}_2\} \cap \mathcal{L}_i(\mu(x)) \neq \emptyset$. According to the $\sqcup -rule$,  $\{\textit{i:C}_1, \textit{j:C}_2\} \cap \mathcal{L}_i(x) \neq \emptyset$, so that  $\mathcal{L}_i(x) \subseteq \mathcal{L}_i(\mu(x))$. The first case where $x \in C_1^{\mathcal{I}_i}$ is addressed by the local reasoning chunk for the unit $i$. In case $\textit{j:C}_2$ is in $\mathcal{L}_i(\mu(x))$, then by the definition of the distributed tableau,  there is an individual $t$ in the $j$-th unit, such that $t=\pi_{ij}(\mu(x))$ and $\mathcal{L}_{j}(t) \supseteq \mathcal{L}_{-i}(\mu(x)) \supset \{\textit{j:C}_2\}$. In this case, according to the $\pi-rule$ and the $\pi-update-rule$ there is a j-node $y$ in the distributed tableau such that  $j:y \overset {=} \mapsto i:x$ and $\textit{j:C}_2 \in \mathcal{L}_j(y)$. If we replace $t$ by $\mu(y)$, then the conditions of $\mu$ are satisfied.
$\Box$}

\end{itemize} }
\end{enumerate}

\subsection{Subsumption propagation in  $E-\mathcal{SHIQ}$.}

The following paragraphs discuss subsumption propagation between remote ontology units in  $E-\mathcal{SHIQ}$. First we prove that in the case of no-chaining concept $onto$ concept correspondences, $E-\mathcal{SHIQ}$ inherently supports subsumption propagation. The following Theorem considers the generic distributed knowledge base depicted in Figure 5 \cite{CRPITV90P21-30}. 

\begin{figure}[!htb]
\centering
\includegraphics[scale=.5]{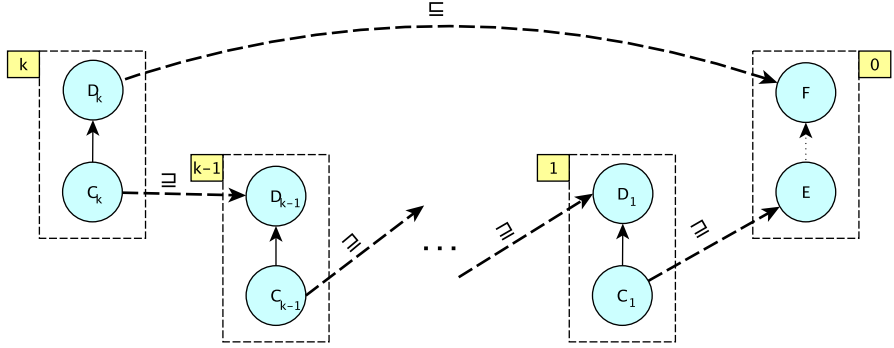}
\caption{A Distributed Knowledge Base. GCI's are indicated by solid arrows, while concept-to-concept correspondences by dashed arrows. Inferences are indicated by dotted arrows.}
\label{fig5}
\end{figure}

\textbf{Theorem} (Subsumption Propagation) 
Given an $E-\mathcal{SHIQ}$ distributed knowledge base $\Sigma$ with index set $I=\{0,1,2,...k\}$, with a distributed TBox with $k+1$ local TBoxes, and with a family of correspondences between units, such that:
\begin{enumerate}
\item{$\Sigma \vDash_d i:C_i \sqsubseteq D_i$ for $1 \leq i \leq n$}
\item{$\textit{i+1:C}_{i+1} \overset{\sqsupseteq} \rightarrow \textit{i:D}_i \in \textbf{C}_{(i+1)i}$ for $1 \leq i \leq k$}
\item{$\textit{1:C}_{1} \overset{\sqsupseteq} \rightarrow \textit{0:E} \in C_{10}$ and $\textit{n:D}_{k} \overset{\sqsubseteq} \rightarrow \textit{0:F} \in \textbf{C}_{n0}$}
\end{enumerate}
then it follows that $\Sigma \vDash_d 0:E \sqsubseteq F$.

\textbf{Proof}.

Let us assume that there is an individual \textit{0:x}  in  $\Delta^{\mathcal{I}_0}$, s.t. $x \in (\textit{0:E}^{\mathcal{I}_0} - \textit{0:F}^{\mathcal{I}_0})$. According to the semantics of concept-to-concept correspondences, \textit{0:x} can not have any corresponding individual in $\textit{k:D}_k^{\mathcal{I}_k}$. There should also be a corresponding \textit{1:x} in  $C_1^{\mathcal{I}_1}$ with $\textit{1:x} \overset{=} \rightarrow \textit{0:x}$.  Therefore, according to unit 0, \textit{0:x} belongs in ($(\textit{0:E}^{\mathcal{I}_0} - \textit{0:F}^{\mathcal{I}_0}) \cap (\Delta^{\mathcal{I}_k} - D_{k}^{\mathcal{I}_k}) \cap C_1^{\mathcal{I}_1}$). Given that domain relations represent equalities, all  individuals corresponding to \textit{0:x} must fulfill the specifications (constraints) for \textit{0:x}.
Regarding \textit{1:x} and given  the local axioms in unit 1, it holds that \textit{1:x} is  in  $D_1^{\mathcal{I}_1}$. 
Therefore, \textit{1:x} belongs in ($(\textit{0:E}^{\mathcal{I}_0} - \textit{0:F}^{\mathcal{I}_0}) \cap (\Delta^{\mathcal{I}_k} - D_{k}^{\mathcal{I}_k}) \cap C_1^{\mathcal{I}_1} \cap D_1^{\mathcal{I}_1}$).


Continuing in the same way along the $into$ concept-to-concept correspondences in the path from unit 0 to the unit $k$, there is a  \textit{k:x}  in  
( $(\textit{0:E}^{\mathcal{I}_0} - \textit{0:F}^{\mathcal{I}_0}) \cap (\Delta^{\mathcal{I}_k} - D_{k}^{\mathcal{I}_k}) \cap (\bigcap_i C_i^{\mathcal{I}_i} )\cap (\bigcap_{i} D_{i}^{\mathcal{I}_j}) )$, for $i=1..k $.  This is clearly a contradiction.$\Box$
\\ \\
Next we discuss the most general case of subsumption propagation among remote ontology units. Such a setting is depicted in figure 6 \cite{CRPITV90P21-30} and it concerns a distributed knowledge base $\Sigma$, with index set $I=\{0,1,2,...k\}$ and a distributed TBox with $n+1$ local TBoxes, and a family of correspondences between units, such that:
\begin{enumerate}
\item{$\Sigma \vDash_d i:C_i \sqsubseteq D_i$ for $1 \leq i \leq n$}
\item{$\textit{i+1:C}_{i+1} \overset{\sqsupseteq} \rightarrow \textit{i:D}_i \in \textbf{C}_{(i+1)i}$ for $1 \leq i \leq k$}
\item{$\textit{i:D}_{i} \overset{\sqsubseteq} \rightarrow \textit{i+1:C}_{i+1} \in \textbf{C}_{i(i+1)}$ for $k \leq i \leq n$}
\item{$\textit{1:C}_{1} \overset{\sqsupseteq} \rightarrow \textit{0:E} \in \textbf{C}_{10}$ and $\textit{n:D}_{n} \overset{\sqsubseteq} \rightarrow \textit{0:F} \in \textbf{C}_{n0}$}
\end{enumerate}
We will show how $E-\mathcal{SHIQ}$ supports $\Sigma \vDash_d 0:E \sqsubseteq F$ and discuss the enhancements of the tableau reasoning algorithm presented.

\begin{figure}[!htb]
\centering
\includegraphics[scale=.5]{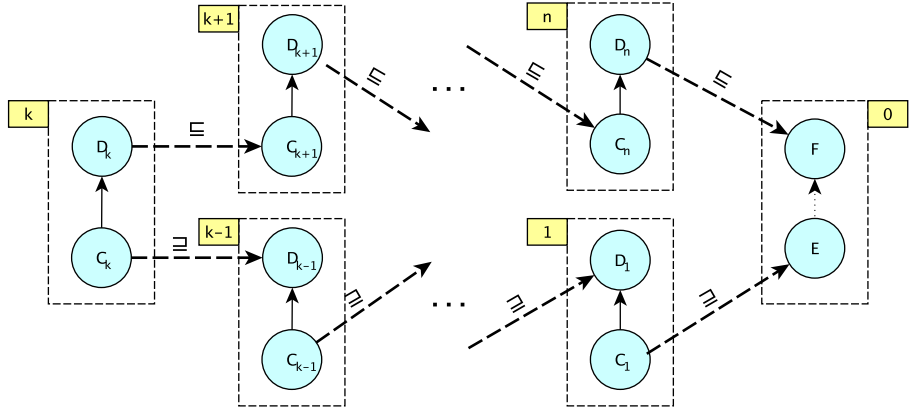}
\caption{A Distributed Knowledge Base. GCI's are indicated by solid arrows, while concept-to-concept correspondences by dashed arrows. Inferences are indicated by dotted arrows.}
\label{fig6}
\end{figure}

Let us assume that there is an individual \textit{0:x}  in  $\Delta^{\mathcal{I}_0}$, s.t. $x \in (\textit{0:E}^{\mathcal{I}_0} - \textit{0:F}^{\mathcal{I}_0})$. According to the semantics of concept-to-concept correspondences \textit{0:x} can not have any corresponding individual in $\textit{n:D}_n^{\mathcal{I}_n}$. There should also be an individual \textit{1:x} in  $C_1^{\mathcal{I}_1}$ with $\textit{1:x} \overset{=} \rightarrow \textit{0:x}$.  Therefore, according to unit 0 (own axioms and its subjective correspondences), \textit{0:x} belongs in ($(\textit{0:E}^{\mathcal{I}_0} - \textit{0:F}^{\mathcal{I}_0}) \cap (\Delta^{\mathcal{I}_n} - D_{n}^{\mathcal{I}_n}) \cap C_1^{\mathcal{I}_1}$). Given that domain relations are one-to-one, corresponding individuals must share the same constraints (i.e. they must belong to the denotation of the same  concepts). 

Given  the local axioms in unit 1, it holds that \textit{1:x} is  in  $D_1^{\mathcal{I}_1}$. But then, given that $\textit{1:x} \overset{=} \rightarrow \textit{0:x}$, according to the subjective point of view of unit $0$, \textit{0:x} should be  in  $D_1^{\mathcal{I}_1}$, as well.
Given also the subjective concept-to-concept correspondences of unit 1, \textit{1:x} belongs in ($(\textit{0:E}^{\mathcal{I}_0} - \textit{0:F}^{\mathcal{I}_0}) \cap (\Delta^{\mathcal{I}_n} - D_{n}^{\mathcal{I}_n}) \cap C_1^{\mathcal{I}_1} \cap C_2^{\mathcal{I}_2} \cap D_1^{\mathcal{I}_1}$): These are also the constraints that  \textit{0:x} must satisfy.

Following the same line of reasoning, according to unit $1$ there is a \textit{2:x}, s.t. $\textit{2:x} \overset{=} \rightarrow \textit{1:x}$. Unit 0 does not assume any relation between the individuals \textit{0:x} and \textit{2:x}. However, given the relation between the \textit{1:x} and \textit{2:x}, both must satisfy a set of constraints (i.e. must be in the denotation of the same concepts): Given the local axioms in these units and their subjective correspondences, both individuals must belong in ($\textit{0:E}^{\mathcal{I}_0} - \textit{0:F}^{\mathcal{I}_0}) \cap (\Delta^{\mathcal{I}_n} - D_{n}^{\mathcal{I}_n}) \cap C_1^{\mathcal{I}_1} \cap C_2^{\mathcal{I}_2} \cap C_3^{\mathcal{I}_3} \cap D_1^{\mathcal{I}_1} \cap D_2^{\mathcal{I}_1} \cap D_3^{\mathcal{I}_1}$).

Even though unit 0 does not know the existence of \textit{2:x}, it does assume that \textit{0:x} must  satisfy all the  constraints of \textit{1:x}.

Thus,  as far as unit 0 is concerned, the constraints must propagate from unit 1 to unit $k$ (downwards), and also in the reverse direction.

Finally, \textit{0:x} is assessed to be in  ( $(\textit{0:E}^{\mathcal{I}_0} - \textit{0:F}^{\mathcal{I}_0}) \cap (\Delta^{\mathcal{I}_n} - D_{n}^{\mathcal{I}_n}) \cap (\bigcap_i C_i^{\mathcal{I}_i} )\cap (\bigcap_{i} D_{i}^{\mathcal{I}_i}) )$, for $i=1..k $.

Exploiting $into$ correspondences from unit $n$ to unit $k$ (i.e. exploiting local axioms and subjective correspondences), \textit{0:x} is assessed to be in  ( $(\textit{0:E}^{\mathcal{I}_0} - \textit{0:F}^{\mathcal{I}_0}) \cap (\Delta^{\mathcal{I}_n} - D_{n}^{\mathcal{I}_n}) \cap (\bigcap_i C_i^{\mathcal{I}_i} )\cap (\bigcap_{i} D_{i}^{\mathcal{I}_j}) \cap (\bigcap_{j} (\Delta^{\mathcal{I}_j} - D_{j}^{\mathcal{I}_j})  )$, for $i=1..k $ and $j=k..n$.
This is clearly a contradiction.

To support this kind of reasoning, we need an additional $\pi-update-rule$ that will propagate constraints (i.e. update the labels of tableau nodes) in the reverse order along projection paths (i.e. paths of the form $(\textit{k:x} \rightarrow ... \rightarrow \textit{1:x} \rightarrow \textit{0:x})$, s.t. $i:x=\pi(\textit{(i-1):x})$, where $i=1,...k$), than in the order where nodes have been projected.
This additional tableau expansion rule is shown in table 4.  This rule requires that if $i:x$ is not blocked and there is a node $j:x'$ such that $j:x' \overset {=} \mapsto i:x$ and $\bigcup_{k \in N(j)-\{i\}} \mathcal{L}_{jk}(x') \neq \emptyset$, then  $\mathcal{L}_{i}(x) \cap \mathcal{L}_{j}(x') = \mathcal{L}_{-i}(x) \cup \mathcal{L}_{-j}(x') $

\begin{table}
\centering
\large
\caption{Enhanced labels' update rule}
\resizebox{1.0\textwidth}{!}{
\begin{tabular}{|p{1.85cm}|p{19.8cm} |} \hline
$\pi$\textit{-update-rule+} & if $i:x$ is not blocked and there is a node $j:x'$ such that $j:x' \overset {=} \mapsto i:x$ and $\bigcup_{k \in N(j)-\{i\}} \mathcal{L}_{jk}(x') \neq \emptyset$, then set $\mathcal{L}_{i}(x)$ equal to $\mathcal{L}_{i}(x) \cup (\bigcup_{k \in N(j)-\{j\}} \mathcal{L}_{jk}(x'))$. \\
\hline
\end{tabular}}
\label{expansionRules}
\end{table}

The $\pi-update-rule+$ can be applied when each of the peers have constructed a local clash-free and complete distributed completion graph. The update of  labels of nodes using the $\pi-update-rule+$ in this case results to adding new concepts to the labels of nodes, assuring that no further application of expansions rules is required. The addition of  new elements in nodes' labels may result to the occurrence of clashes. It must be pointed out that by applying this rule, the property 11 of tableaus need no change, since  the rule requires that $\mathcal{L}_{i}(x) \cap \mathcal{L}_{j}(x') = \mathcal{L}_{-i}(x) \cup \mathcal{L}_{-j}(x') $. Since $\mathcal{L}_{-i}(x) \cup \mathcal{L}_{-j}(x') \supseteq \mathcal{L}_{-i}(x) $, it holds that $\mathcal{L}_{i}(x) \cap \mathcal{L}_{j}(x') \supseteq \mathcal{L}_{-i}(x) $ \footnote{The new rule affects only the proof of the converse of the satisfiability theorem, since the label for a node \textit{i:x} must include not only the concepts in the labels of nodes projected to this node (as it is specified), but also (due to the new rule) the concepts in the labels of the nodes to which \textit{i:x}  is projected. Then, it can easily be proved that the property 11 of tableaus holds in this case as well.}.

\section{Implementation of the $E-\mathcal{SHIQ}$ distributed reasoner}

Given a set of indices $I$, a distributed knowledge base $\Sigma$ with $\|I\|$ ontology units, and a set of $\|I \|$  peers such that each peer is responsible for a specific ontology unit, the $E-\mathcal{SHIQ}$ distributed reasoner combines the local reasoning chunks for each of these peers towards constructing a distributed tableau. 

Each peer exploits its own ontology unit, the concept-to-concept correspondences and link relations with neighboring units. Each unit also holds assertional knowledge specifying concept and role assertions, as well as assertions concerning link-relations, and individual correspondences with neighboring units. This information is stored in standard OWL files and a modified version of C-OWL \cite{conf/semweb/BouquetGHSS03} file, that in addition to correspondences specifies link-relation restrictions and assertions. Subsequently, when we refer to the peer $i$, we  also refer to the ontology unit $i$, and vice-versa. 

\begin{figure}
\center
\includegraphics[scale=.55]{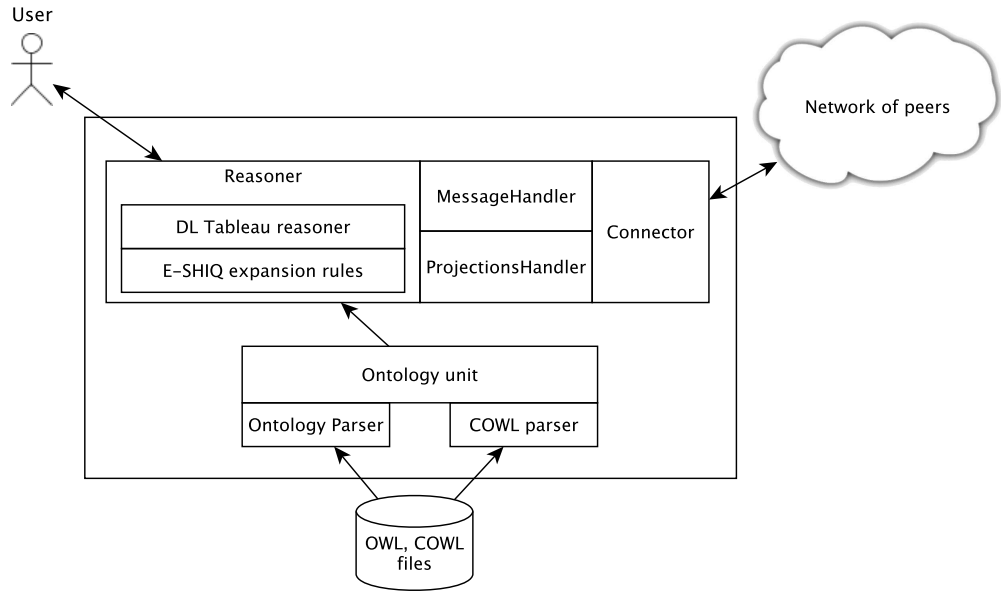}
\caption{The overall architecture of a peer}
\label{architecture}
\end{figure}

The overall architecture of a reasoning peer is shown in figure \ref{architecture}. The core component of the peer is the reasoner which implements the $E-\mathcal{SHIQ}$ tableau algorithm. This is an extension of the Pellet reasoning engine, although any other tableau reasoner would fit the purpose. The connector component communicates messages to/from other peers. Messages are mainly of two types: $Projection-request$ messages and $projection-response$ messages.  The connector runs on a separate thread per peer, enabling asynchronous communication between them, via TCP/IP sockets. For the deserialization of the unit's knowledge base, each peer uses an ontology and COWL parser. A reasoning peer has been implemented to operate  over both HTTP and local file access.


\subsection*{C-OWL extension}

The C-OWL parser deserializes the concept-to-concept correspondences that hold between concepts in different units, as well as, link relations, link assertions and individual correspondences specified in the C-OWL file. Specifications in C-OWL extend those specified in \cite{conf/semweb/BouquetGHSS03}  
with support for link relations, assertions and transitivity axioms, as defined in section \ref{combining}.

Specifically, we can distinguish the contents of a C-OWL file into two major parts. The first part
contains the individual and concept-to-concept correspondences, and the second the link-relation restrictions and assertions. The correspondences are grouped under the element
\texttt{Mapping}. Each \texttt{Mapping} element specifies a correspondence to a specific unit. Of course, we can specify
a number of mappings in a C-OWL file, and a number of correspondences per mapping element. 

Considering the example in  figure \ref{fig4}, the concept-to-concept correspondences between the concepts  \texttt{Conference} and \texttt{MedicalConference} will be specified in a C-OWL file for \texttt{unit1} as follows:

\noindent
\begin{Verbatim}[fontsize=\small]
<cowl:Mapping>
  <cowl:sourceOntology>
    <owl:Ontology rdf:about="http://localhost/figure4/unit2.owl"/>
  </cowl:sourceOntology>
  <cowl:targetOntology>
    <owl:Ontology rdf:about="http://localhost/figure4/unit1.owl"/>
  </cowl:targetOntology>
  <cowl:bridgeRule>
    <cowl:Onto><cowl:source>
      <owl:Class rdf:about="http://localhost/figure4/unit2.owl#Conference"/>
    </cowl:source><cowl:target>
      <owl:Class rdf:about="http://localhost/figure4/unit1.owl#MedicalConference"/>
    </cowl:target></cowl:Onto>
  </cowl:bridgeRule>
</cowl:Mapping>
\end{Verbatim}

The elements \texttt{cowl:sourceOntology} and \texttt{cowl:targetOntology} set the source and target ontology URIs for this \texttt{cowl:Mapping} element. The $onto$ correspondence is specified in the \texttt{cowl:bridgeRule} element, by the child element \texttt{cowl:Onto}. The elements \texttt{cowl:source} and \texttt{cowl:target}, specify the concepts for this correspondence. Similarly, we can specify individual and $into$ correspondences, using the tags \texttt{cowl:IndividualCorrespondence} and \texttt{cowl:Into}, respectively.

The restrictions on links follow the structure of standard OWL role restrictions. For the link relation $presentedAt$  of \texttt{unit1} in figure \ref{fig4}, the C-OWL file for this unit  contains the following specifications:

\noindent
\begin{Verbatim}[fontsize=\small]
<cowl:Linking>
  <cowl:LinkProperty rdf:resource="http://localhost/figure4/unit1.owl#presentedAt"/>
</cowl:Linking>
<owl:Class rdf:about="http://localhost/figure4/unit1.owl#Article">
  <owl:equivalentClass>
    <owl:Restriction>
      <owl:onProperty rdf:resource="http://localhost/figure4/unit1.owl#presentedAt"/>
        <owl:allValuesFrom rdf:resource="http://localhost/figure4/unit4.owl#Event"/>
    </owl:Restriction>
  </owl:equivalentClass>
</owl:Class>
\end{Verbatim}

The element \texttt{cowl:Linking} states that \texttt{\&unit1;presentedAt} is a link relation and further includes restrictions for this property. We can specify any number of restrictions in the same C-OWL file and for the same link relation. The restriction illustrated, defines that every \texttt{\&unit1;Article} is \texttt{\&unit1;presentedAt} an \texttt{\&unit4;Event}. 

The C-OWL extension proposed has important differences compared to the OWL extension proposed in \cite{GrauPS09}. 

By extending  C-OWL with specifications for link relations, we specify how the knowledge in a unit is combined - together with concept-to-concept correspondences - with knowledge in other units. Thus, the local knowledge is clearly separated from correspondences and link relations to elements of other units. As an extension to this fact, a peer using this unit can anytime reason in a completely isolated way, using only local knowledge, without considering link relations and correspondences. Doing so, it considers that all  concepts in other units are equivalent to the \textit{$top$-concept}, considering them to be interpreted by a full-hole interpretation. This is important, since peers are supported to identify whether  inconsistencies result from local knowledge, or whether it is an effect of combing local knowledge with knowledge in other units.

Also, the extension proposed in \cite{GrauPS09}, assumes that distinct units are interpreted over disjoint domains. As a consequence of this, a resource declared as a local i-concept, cannot be used also as the range of an  ij-link-relation, $i,j \in I$. Supporting punning in $E-\mathcal{SHIQ}$, an ij-link relation can be used as an i-role as well, and thus be restricted to i-concepts, as well as to j-concepts.

\subsection{The Reasoning Process}

The reasoning process in each of the peers can be regarded to run in two phases: The \textit{initialization} and the \textit{communication} phase. These two phases may be intertwined. Nevertheless, the current implementation considers these two phases to be distinct, running one after the other. Although this approach does not take full advantage of the potential parallelization of the overall process, there are certain advantages discussed in the paragraphs that follow.

The initialization phase includes the deserialization of the knowledge stored in the OWL and C-OWL files and the construction of the local completion graph, testing about the consistency of peer's  own local knowledge (i.e. of its ontology unit). A peer with consistent local knowledge participates in the distributed reasoning process by evaluating projection-requests towards computing an overall tableau, jointly with other peers, while a peer with inconsistent knowledge considers a hole interpretation for its own unit and informs the others for this: The consequence of this is that neighboring peers consider the ontology elements of that peer (i.e. concepts participating in link restrictions and/or concept correspondences) to be equivalent to the top concept. 

The advantage of having a distinct initialization phase per peer is that initialization processes run in parallel (i.e. for each peer) without requiring any further communication with neighbors, while peers with inconsistent knowledge may not  be considered during the communication phase, saving communication cost and time. 
Furthermore, after the initialization phase, each peer with consistent knowledge preserves the local completion graph for later use. This will be further explained in the paragraphs that follow. A further advantage of serializing the two phases, is that expansion rules do not apply to the labels of nodes after the construction of a complete completion graph. Therefore, projections may include only the concept names in nodes' labels. This simplifies communication between peers, and also takes full advantage of parallelizing the local reasoning chunks.  

Having finished the initialization phase, peers posess their local completion graphs, and know which of their neighbors hold inconsistent knowledge. During the communication phase, peers exchange $projection-request$ and $projection-response$ messages. A projection-request message directed from a peer $i$ to a peer $j$ asks for the projection of a (source) node corresponding to an individual \textit{i:x} to the local reasoning chunk of $j$. Such a projection-request may specify the corresponding (target) individual of $i:x$ (e.g. \textit{j:x}) in case peer $i$ subjectively holds such a correspondence between individuals, or it may ask the creation of a new node in $j$ according to definition 6. The projection-request message also contains the $\mathcal{L}_{-i}$ fragment of the label of \textit{i:x}, according to definition 6. 
Please recall that given the label of a node in peer $i$, projection-request messages are sent from $i$ to any other neighbor peer $j \in I$ with $\mathcal{L}_{ij} \neq \emptyset$.
A projection-response message from peer $j$ to peer $i$, is the response to a projection-request message sent to  $j$ from   $i$. Such a response may either indicate a clash, or it may specify a (possibly empty) set of concept names to be added in the source node's label (i.e. the node corresponding to \textit{i:x}), according to the $\pi-update-rule+$.

Given the large number of messages exchanged between peers,  for the effectiveness of the communication process, each peer  packages all projection-request messages to be sent to any of its neighbor peers to a single message. The same happens for the projection response messages sent back. Of course this can not be the case when the initialization and communication phases are intertwined. 

Each peer that receives a package of projection-request messages (asking for a set of projections to be made) will make a working-copy of its local completion graph and do the following: (a) For each requested projection it will consider either opening a new node in the working copy of its local completion graph or, in case the projection specifies a specific corresponding (target) individual for which there is a node in the constructed completion graph, it will update the label of that node; and (b) it will check whether any of these additions result to a clash, or to further updates in the labels of nodes in originating peer's local completion graph. These tasks run on a new thread in the projection-request message receiver, so as to allow other projection-request messages made from any other peer to be processed concurrently.   

It must be pointed out that given a projection-request from a peer $k$ to the peer $i$, then $i$ will make a working copy of its own completion graph and serve all the requests from $k$ there.  Having said this we have to recall that projections concern subjective correspondences between individuals in different units: Thus projections from $k$ received by $i$, are considered independently from projections originated from any other peer $l \neq k$ during the reasoning process in $i$. 


Due to projection-request messages received by peer $i$, this peer may also "trigger" further projection requests to any neighbor $j$. These projection requests from $i$ are packaged together with other requests that are due to $i$'s own knowledge (i.e. from nodes in the local completion graph of $i$). Subsequently we refer to these projection requests as requests that have been $triggered$ by $k$. 
Packaging all these requests to $j$  is necessary given that peers have to reason taking into account peers' coupled knowledge. Thus, nodes that have already been projected from $i$ to $j$ due to $i$ own knowledge, they are re-projected to $j$ so as these to be processed in conjunction with nodes that are due to knowledge from $k$. Of course this pattern of projections happens along any path of peers. This has  important consequences to the efficiency of any reasoning task: Projection requests may be repeated several times, and for each single projection request from $k$ to $i$, the peer $i$ repeats numerous other projection requests to $j$. If this continues along a path, then there can be avalanches of projection requests. This phenomenon is further discussed in the experimental results section.

Since each peer  makes a working copy of its local completion graph for each set of projection-requests received from any of its neighbors, we have a considerable cost in memory resources, but the gain in performance is high given that peers do not need to rebuild their local completion graph in order to serve each projection-request.

If one of the neighbors responds to a projection-request message with  a clash, then the projection-response message will force the source node (i.e. the corresponding node in the completion graph of the sender) to close, and the tableau algorithm in the  sender of the request will seek for alternative branches in the local completion graph. 
If on the other hand, the projection-response message indicates an update of source node's label, the  message contains a set of additional concept names to be added in the label of the source node.

For each projection-response message the receiver peer decides whether a clash occurs in any of the source nodes whose label has been updated. It must be noticed that, given that updates of corresponding nodes' labels concern only additions of concept names, there is no need of re-applying any of the expansion rules, thus, termination of the algorithm is not affected.  Nevertheless, as already stated, the update of any node label may trigger further projection-requests to any of the neighboring peers. 

Given that a completion graph node may be projected to more than one neighboring peers, it is possible that some of these peers may respond earlier than others to this request with a clash. To save computational resources, in case a peer sends a request for a projection to a set of neighbors (projection-request receivers), and it receives a clash for that projection from any subset of the projection-request receivers, it will inform the rest of its neighbors about this result, asking them to stop the local reasoning process for this request. Obviously, this may lead to a reaction of ``stop-requests'' among peers, restoring an important amount of computational resources.

Finally, given that a peer may issue the same projection-request several times to the same neighbors, we further optimize the distributed reasoning process using what we have called a \textit{projection cache} for each of the peers. Using the projection cache a peer can store the projection-request messages it sends to its neighbors and the corresponding projection-response messages it receives from them. Specifically, each record in the cache is a tuple of the form
$\langle \textbf{PrRq}, P, \textbf{PrR} \rangle$, where \textbf{PrRq} is a packaged projection-request message sent to peer $P$, and \textbf{PrR} is the packaged projection-response message received from $P$ (containing the response for each individual projection request in \textbf{PrRq}). 
We do not implement a garbage collector for projection cache records, since this would be an overhead for the reasoner's performance. When a packaged projection-request is to be sent, the peer first queries its cache to find if this message has already been sent. If an exact-match is found, the projection response is retrieved from the cache. Otherwise, the message is sent and the cache is updated with this message and with the response received. It must be noticed that in the current implementation the package of all requested projections must match with a record in the projection cache. Further sophisticated mechanisms for matchmaking can be implemented, given of course that sophistication will not incur further excess requirements on computational resources.

\section{The Impact of Projections in the Reasoning Process: Experimental Results}

While peers  reason jointly to combine their knowledge, it can easily be seen that  since the initialization phases happen concurrently and  locally to each of the peers - and this is also the case for the computation and further exploitation of the projection-response messages - an $E-\mathcal{SHIQ}$ reasoning process in comparison to a centralized reasoning process (i.e. a process that exploits a single ontology, whose knowledge entails any axiom entailed by the distributed knowledge base and vise-versa), incurs an additional communication cost between peers, due to the projection messages required for combing the local completion graphs.

This section aims to demonstrate and discuss interesting aspects concerning the behavior of the implemented $E-\mathcal{SHIQ}$ distributed reasoner towards the implementation of a distributed registry of services in a democratized grid or cloud setting where agents order (i.e. offer / request) resources. Towards this target we are using the Semantic Information System Ontology (SIS ontology), briefly presented in section 7.1.  While any other ontology could be used for our purposes, it would require its partition to an $E-\mathcal{SHIQ}$ distributed knowledge base. Having said that, we must point out that the comparison of the $E-\mathcal{SHIQ}$ distributed reasoner with other distributed reasoners can not be done in a direct way, given that such reasoners operate under different principles and assumptions, requiring knowledge to be specified in other representation frameworks.  

For experimentation purposes we have partitioned different versions of the SIS ontology manually. Each ontology version is a populated SIS ontology, specifying a snapshot of the SIS registry at a particular time point. This is further detailed in the section that follows. While by using any sophisticated partitioning algorithm we may achieve a ``better'' distribution of the axioms and of the individuals to  different units, each SIS version has been partitioned meticulously to distinct units so as to study the impact of projection messages in the distributed reasoning process in the following cases:

- Having units of different sizes in the reasoning process. The size of each unit may be determined by the set of known axioms, the set of known individuals, or a combination of them. Here we consider both numbers (i.e. the number of axioms and the number of individuals). This is due to an idiosyncracy of the SIS ontology, which makes the case of this ontology very interesting. We will delve into this issue in the section that follows. The size of units is important also in combination to  the subjectiveness of correspondences, i.e. to the direction of projection-request messages: In case correspondences and link relations are from the smaller to the larger units, then there will be a small number of projection-request messages, while the number of such messages will be much larger in case correspondences and links are from larger units. Consider for instance two units $i$ and $j$ of different magnitude. In case all correspondences and link relations are from the smaller unit $i$ to the larger unit $j$, then the number of projections will be significantly less than those in the case where correspondences and links were from $j$. While this is obvious, we need to study the cost incurred by projection messages in these cases, as well as the impact of having the projection cache. All comparisons are made against a centralized reasoning process.  

- Having different configurations of units' connectivity, also in combination to  the direction of projection-request messages: Having a unit $i$ that receives/sends projection-requests from/to other  units $j$, $k$ and $m$, then  unit $i$ acts as a projection-messages multiplier between these units: Each projection-request message from peer  $j$ to  $i$ may result to numerous other projection-requests to peers $j,k$ and $m$. Thus, if $i$ receives $M$ messages from any other peer, these (in the worst case) may trigger $3*M$ messages to  its neighbor peers (including $j$), plus $3*N$ messages concerning projection-requests due to own knowledge: In total $3*(N+M)$ messages. In cases with where knowledge is distributed among units in an unbalanced way, $M$ can be order(s) of magnitude larger than $N$. In case knowledge is distributed in a balanced way among multiple units, then $N$ and $M$ are of the same order of magnitude, and can be significantly smaller than  $M$ in the ``un-balanced'' case. In this case, we expect a significantly  smaller number of projection-requests and responses among peers.
Correspondences and link relations among units may also lead to cycles of projections in cases where there are cyclic ``connections'' between units \footnote{a cyclic connection between units $i$ and $j$ may be due either to concept and/or individual correspondences from $i$ to $j$ and from $j$ to $i$ (maybe through paths), or to combinations between concept and/or individual correspondences from $j$ to $i$ and ij-link relations.}, and to "avalanche" effects due to the triggering of projection-request messages along paths in the network of peers. These phenomena incur a great communication cost to the overall reasoning process. We need to test whether the  caching of projections has  any significant effect towards reducing this cost.

To further justify the experimental settings chosen we have to make the following remarks:

- Having units of significantly different sizes presents a worst case scenario, especially in the cases where concept and individual correspondences are symmetric  between units: I.e. in cases where for each correspondence from the subjective point of view of a unit $i$ to a unit $j$, there is a correspondence stating the same relation between ontology entities from the subjective point of view of unit $j$ to $i$. The effects of differences in the size of  units, compared to cases where knowledge is distributed evenly to units, concern the parallelization of the reasoning process, as well as the number of projection-request messages sent.  
As far as the parallelization of the reasoning process is concerned, while larger units impose a much larger processing load than smaller units, the latter have to delay their processing until they receive responses to their projection requests. This ``slack time'' of peers with smaller units is much larger than the corresponding time required when knowledge in distributed in units in a balanced way.  
As far as the number of projection-requests is concerned, the effect of having units acting as multipliers of such messages between larger units, means that any large unit will receive - in the worst case- a large number of projection requests, which is equal to the sum of the number of projection requests emanating from each of the other  large units. This, in combination to the local processing load imposed by larger units, result to a rather unbalanced load between peers, further increasing the ``slack time'' of peers with smaller units. As already pointed out, in any balanced case, although the ``multiplication effect'' will occur, the local processing load can be more balanced, and there will be a smaller amount of messages to each of the peers, due to the distribution of knowledge.

- The number of individual correspondences any unit possesses does not affect the communication cost. In contrast to that, the number of individuals known in each unit, in relation to the concept correspondences and link relations to other units, affect the communication cost: This is so, since any instance of a concept will be projected to another unit, if there is a correspondence or a link relation to the other unit. This happens independently of the existence of individual correspondences between units. As already said, in case such an individual correspondence is known, then the completion graph node for this individual will be projected to the node of the corresponding individual. If there is not any such individual correspondence, the  projection-request receiver will be  asked to open a new node in its completion graph. In our experiments, peers do not possess any  correspondences between individuals. 

- Distinguishing between the cases where knowledge  in different units is coupled with concepts' correspondences only, from the cases where knowledge is coupled  with link relations only, is not important for measuring the efficacy of the distributed reasoner, given that during reasoning both types of constructors result to projection requests from one unit to the other.



\subsection{The SIS ontology}
The Grid4All semantic information system (SIS) \cite{VourosPKVTVKA08} provides a matchmaking and selection service for agents (either software or human) willing to offer or request resources or services in a democratized Grid or Cloud environment. Each agent participates in the system as a provider, as a consumer or both. The system supports agents to discover markets that trade requested/offered resources and also satisfy specific requirements concerning availability, cost,  quality etc. We refer to offers and requests as $orders$, while any type of order may initiate a market.  Markets and resources are directly registered to the Semantic Information System (SIS) ontology. To retrieve markets, SIS must perform a matchmaking process, matching registered orders to agents' queries.  

More specifically, the SIS ontology represents knowledge about agents, markets, and the different types of resources available in a Grid or Cloud  environment. In such an environment, a resource is a tradable entity that is offered by a resource provider, or requested by a resource consumer. Hardware resources are dealt at a logical rather than at a physical level. This is done, because a consumer is less concerned about the specific properties of a piece of hardware (i.e. the controller technology of a hard disk), and is more concerned about the services available by the resource (i.e. the capacity of a hard disk). 

Each resource has a unique identifier and it is related to a specific service, through which it is available in the overall environment. A resource can be described as atomic, composite or aggregated. A hardware resource is always hosted in a machine, and a machine always hosts at least one atomic resource. A composite or aggregated resource consists of atomic resources in one or more machines. The difference between composite and aggregated resources, is that a composite resource comprises at least two heterogeneous resources (i.e. a computational and a storage resource), while an aggregated resource comprises homogeneous resources. Other properties, such as the location, quality, quantity and performance of each resource may also be specified.

In this article we focus on the aggregated and composite resources available in the registry, and specifically clusters and compute nodes. A compute node comprises exactly one computational resource and any number of storage resources. A computational resource is also a composite resource, constructed by at least one CPU, a volatile memory and an operating system, all installed on at least one machine. On the other hand, a cluster is an aggregated resource, as it comprises a set of compute nodes.


There are several reasons for which we have selected the SIS ontology for performing experiments with the $E-\mathcal{SHIQ}$ distributed reasoner, instead of any other benchmarking ontology. The first is our motivation towards this research: Having a centralized registry operating with this ontology is a bottleneck to SIS. Second, any other ontology can not be used for benchmarking purposes as it is: We would need to partition it to a distributed $E-\mathcal{SHIQ}$  knowledge base, as we did with the SIS ontology. 

Delving into the details of the SIS ontology, and towards more crucial arguments towards using this ontology in our study, each registered order  (i.e. offer or request), expands both the terminological (TBox) and assertional (ABox) parts of the SIS ontology: This results to an ontology that has both, a large number of GCI and also a large number of assertions in its ABox. Having said this, we need to point out that a request is formed as a concept specifying the restrictions an offered resource must match, while on the other hand, an offer is formed both, as an instance specifying the specific resource traded, and as a concept specifying the  restrictions any matching requested resource must satisfy. Doing so, the retrieval of instances per order-query (either request or offer) is done by means of a classification task: The query is a concept which is classified in the ontology. All instances of its sub-concepts match the corresponding order. This idiosyncrasy of the SIS ontology has been motivated by our attitude towards supporting the matchmaking task by means of the classification of concepts in the largest possible extent. This makes the use of the SIS ontology during reasoning interesting, given the large number of axioms and individuals created, as the number of registered orders increases.

In addition to those, given that a large number of agents can participate in the grid/cloud and that each agent can  set a large number of orders, this entails the requirements (a) to serve all  agents' orders in a timely fashion, and (b) to reason with large snapshots of the  SIS registry. In the experimental settings we use different snapshots of the SIS registry, to test the distributed reasoner for a variety of registry sizes, varying the number and type of registered resources.
Among the tradable resources in the SIS registry, we perform experiments using orders concerning compute node and cluster resources, due to  the complexity of their specifications. 

The above-mentioned facts, in combination to the expressivity of the language used, makes the SIS ontology itself a bottleneck for the scalability of the Grid4All. Indeed, this is the case when the SIS registry is supported by a monolithic ontology and in case the matchmaking reasoning services are to be provided by a single (centralized) reasoner. This makes SIS a very interesting case for studying the behavior of the $E-\mathcal{SHIQ}$ distributed reasoner.  

\begin{figure}[!htb]
\center
\includegraphics[width=\textwidth]{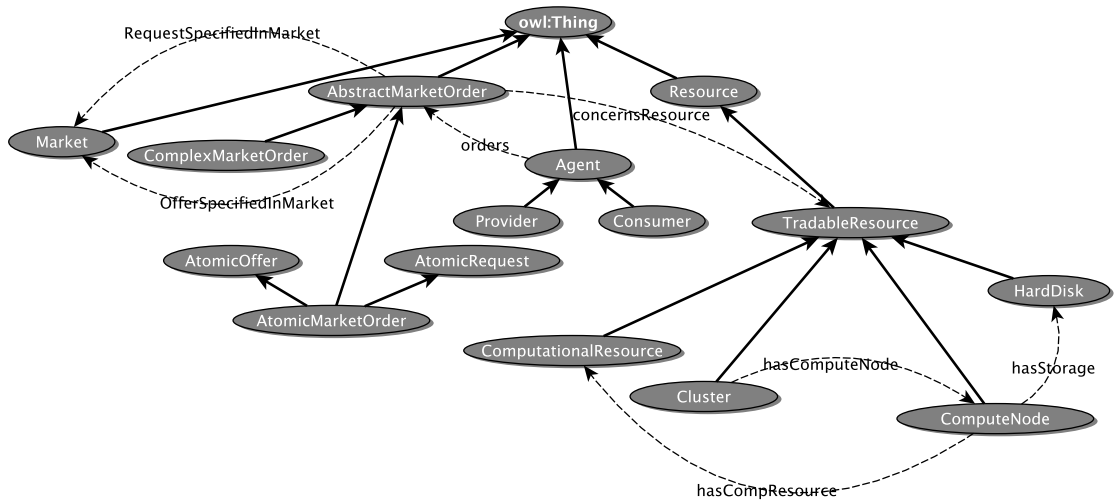}
\caption{The concepts and roles of the SIS registry: Solid arrowed lines show subsumption relations and dashed lines show roles. }
\label{SIS}
\end{figure}

The basic concepts and the roles in  this ontology are illustrated in figure \ref{SIS}.

\subsection{Set up of experimental cases} 
We have evaluated the $E-\mathcal{SHIQ}$ distributed reasoner over a variety of SIS registry versions, extracted  from the Grid4All system. 
Each registry version is serialized in an  OWL file, resulting to a set of ontologies of 300, 600, and 1200 compute node (denoted by $cn$) and 50, 100, and 200 cluster (denoted by $c$) orders. We denote these ontologies with the number of instances and their type: E.g. 300cn is the ontology with 300 compute nodes and  50c the ontology with 50 clusters\footnote{available at: http://ai-group.ds.unipi.gr/ai-group/SISontologies.html}. 

Each ontology uses 28 object properties and 47 disjointness axioms. All ontologies are in the $\mathcal{ALCHIQ}$ fragment of Description Logics. This is the expressivity of the SIS ontology: Transitivity axioms do not exist in this case, although the reasoner supports transitive roles and link-relations.



\begin{table}
\centering
\large
\caption{Concepts, Individuals and axioms per unit}
\resizebox{1.0\textwidth}{!}{
\begin{tabular}{|c|c|c|c|c|c|c|c|c|c|c|c|c|c|} 
\hline
\multirow{2}{*}{}	&\multirow{2}{*}{unit}	&\multicolumn{2}{|c|}{300cn}	&\multicolumn{2}{|c|}{600cn}	&	\multicolumn{2}{|c|}{1200cn}	&	\multicolumn{2}{|c|}{50c}	&	\multicolumn{2}{|c|}{100c}	&	\multicolumn{2}{|c|}{200c}	\\ \cline{3-14}
	&	&2-unit	&4-unit	&2-unit	&4-unit	&2-unit	&4-unit	&2-unit	&4-unit	&2-unit	&4-unit	&2-unit	&4-unit\\ \hline
	&\# 1	&325	&125	&625	&224	&1225	&424	&395	&147	&735	&261	&1441	&494\\
Concepts	&\# 2	&11	&11	&11	&11	&11	&11	&61	&61	&110	&110	&211	&211\\
	&\# 3	&	&126	&	&225	&	&426	&	&149	&	&263	&	&499\\
	&\# 4	&	&124	&	&226	&	&425	&	&149	&	&261	&	&498\\ \hline
	&\# 1	&4963	&1655	&9540	&3181	&19002	&6335	&5332	&1778	&10407	&3469	&20324	&6777\\
	&\# 2	&300	&300	&600	&600	&1200	&1200	&285	&285	&554	&554	&1108	&1108\\
Individuals	&\# 3	&	&1653	&	&3180	&	&6334	&	&1778	&	&3469	&	&6774\\
	&\# 4	&	&1655	&	&3179	&	&6333	&	&1776	&	&3469	&	&6773\\ \hline
	
&\# 1	&944	&344	&1845	&642	&3643	&1242	&1155	&411	&2174	&753	&4293	&1452\\
Axioms	&\# 2	&14	&14	&14	&14	&14	&14	&64	&64	&113	&113	&214	&214\\
	&\# 3	&	&348	&	&645	&	&1247	&	&417	&	&759	&	&1467\\
	&\# 4	&	&342	&	&648	&	&1244	&	&417	&	&752	&	&1464\\ \hline
\end{tabular}}
\label{distribution}
\end{table}

Each ontology has been partitioned in a number of units that are connected in certain ways. We have implemented distributed knowledge bases of 2 and 4 units\footnote{available at: http://ai-group.ds.unipi.gr/ai-group/ESHIQontologies.html}. All units are in the $\mathcal{ALCHIQ}$ fragment of Description Logics. We have not used any specific modularization algorithm, since this is out of the scope of our current work, while the objective was to partition ontologies so as to be able to study the additional cost incurred by the communication between reasoning peers, as described in the introductory part of this section. 

\begin{figure}[h]
\begin{center}$
\begin{array}{ccc}
\includegraphics[width=0.33\textwidth]{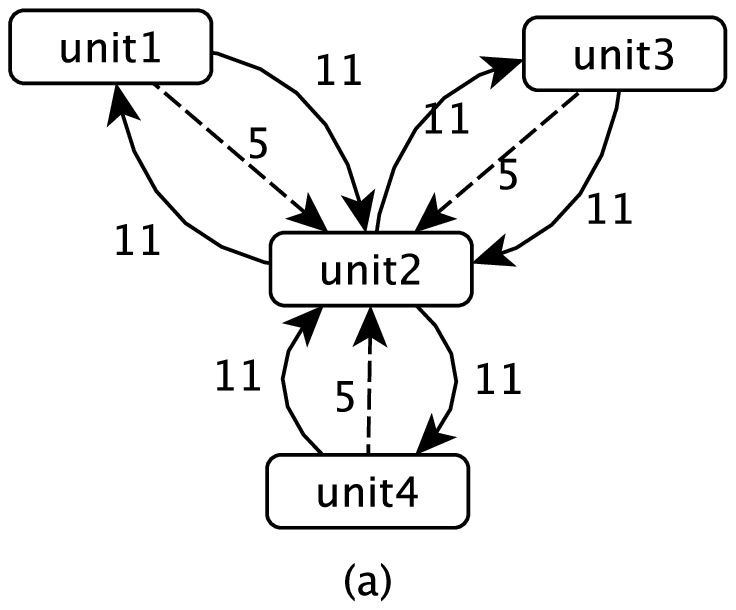}&\includegraphics[width=0.33\textwidth]{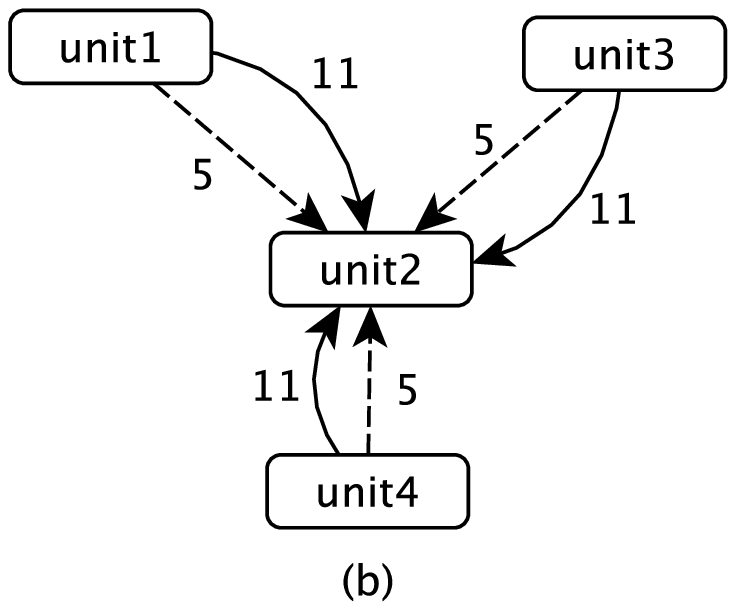}&\includegraphics[width=0.33\textwidth]{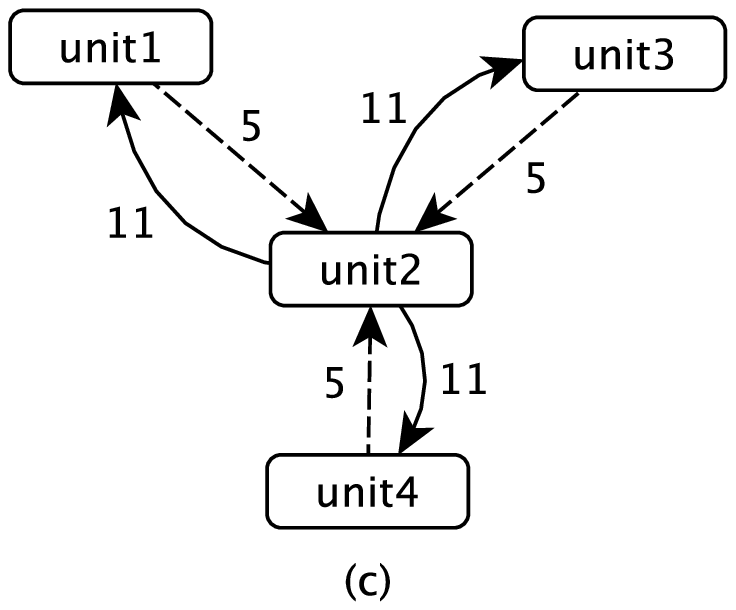}
\end{array}$
\end{center}
\caption{4-unit networks: (a) BD, (b) BS and (c) SB correspondences respectively}
\label{nets}
\end{figure}

Networks of 4 units have been formed according to the topology shown in figure \ref{nets}, where continuous edges denote concept-to-concept correspondences and the dashed edges denote link relations. The direction of the edges denotes the unit for which the correspondences or link-relation specifications  hold. Thus, projection requests are sent to the opposite direction than that of the  arrows. Units 1, 3 and 4 share knowledge for compute nodes, storage and computational resources, while unit 2 holds knowledge about clusters, agents and generic axioms about atomic resources. For this reason unit 2 always holds the lower number of axioms than other units in the network, while it holds the lower number of individuals, especially for the cases where the registry is being populated by compute nodes. The numbers at each edge denote the number of correspondences and links between units: This shows in an abstract way the number of concepts shared, or subsumed  between units. 

As can be noticed in figure \ref{nets}, we study the behavior of the distributed reasoner in cases where  (a) correspondences between units are symmetric (bi-directional), in the sense that subjective views of connected units concerning correspondences coincide. These cases are denoted as ``BD''. (b) Correspondences exist from the point of view of the smaller unit (i.e. from unit 2).  These cases are denoted by ``BS'', and (c) Cases where correspondences are from the point of view of the larger units (i.e. from units 1, 3, 4) to the smaller unit. These cases are denoted by ``SB''. Of course link relations have a specific direction in all cases. It must be emphasized that the same cases are studied for settings of 2 units. While units 1,3,4 share the same specifications both at the terminological and assertional levels, we have not set correspondences between them, in order to have a clear picture of the impact of units' size to the communication overhead. In any case, these correspondences do not add any further information to our understanding of the distributed reasoner behavior given that long paths and cycles still exist in the BD and SB settings: While in the case of networks of type BS, there are no projection requests to unit 2, in networks of type BD or type SB, any projection request to unit 2 may trigger further propagation requests from unit 2 to any other unit.

The partitioning of knowledge in SIS registries is done in two steps. First, we distribute the concept names of an empty SIS registry (i.e. of a registry with no concepts and individuals concerning specific orders) to the units. This is also  done for the axioms where these concepts are involved. The only criterion at this step is that unit 2 will possess mainly knowledge about clusters. Next, we proceed with the population of the units with orders available in each version of the SIS registry. For each individual $x$ in the registry, we identify its types, as well as the role assertions in which it is involved, together with the related individuals. Then, the units that contain the corresponding concept(s) in their signature are identified. Assertions concerning that individual are added to the unit having the smaller ABox. In case of ties, we do a random selection among units. Any role assertions for the specific individual will be added in the unit, either as a role or a link-relation assertion. An assertion is added in the unit as link relation assertion if the related individuals are in different units, or as a role assertion if the individuals are in the same unit. If one of the individuals in any assertion has not been placed in a unit, the assertion is placed in a queue until the home unit of each individual has been decided.

The distribution of the concepts, axioms and individuals in units is summarized in table \ref{distribution}.  It must be noticed that unit 2 has much smaller TBox and ABox from all other units, while it has a central position in any network: This is the worst case described earlier, where large units  in the periphery  trigger projection requests to other larger units via unit 2.



\subsection{Experimental Results}

Experiments concern networks of 2 and 4 peers, where, as already said, each peer is responsible for a single ontology unit. 

\begin{table}
\center
\caption{Classification times (secs) }
\resizebox{\textwidth}{!}{
\begin{tabular}{ | bc | ^c | ^c | ^r | ^r | ^r | ^r | ^r | ^r | ^r | ^r | ^r | ^r | ^r | ^r |}
  \hline
 \multicolumn{1}{|c|}{\multirow{2}{*}{cache}} 	&\multicolumn{1}{|c|}{\multirow{2}{*}{units}}& \multicolumn{1}{|c|}{\multirow{2}{*}{type}} &\multicolumn{2}{|c|}{300cn}	&\multicolumn{2}{|c|}{600cn}	&\multicolumn{2}{|c|}{1200cn}&\multicolumn{2}{|c|}{50c}&\multicolumn{2}{|c|}{100c}&\multicolumn{2}{|c|}{200c}\\ \cline{4-15}
	&	&	&in sequence	&parallel	&in sequence	&parallel	&in sequence	&parallel	&in sequence	&parallel	&in sequence	&parallel	&in sequence	&parallel\\ \hline 
\rowstyle{\bfseries}&	&BD	&527.2	&505.3	&1681.8	&1624.2	&7820.7	&7654.1	&567.3	&526.8	&2366	&2242.8	&10411.5	&9867.2\\
\rowstyle{\bfseries}enabled	&2	&BS	&211.1	&210.8	&546.1	&545.6	&2441.4	&2440.4	&171.1	&170.8	&796.6	&796	&3530.9	&3529.7\\
\rowstyle{\bfseries}	&	&SB	&337.3	&320.7	&864.7	&832.7	&3819.7	&3719	&334.7	&307.5	&1384.4	&1289	&6203	&5761.8\\ \hline
	&	&BD	&543.2	&523.2	&2048.4	&1994.8	&7789.5	&7623.3	&1329.6	&1290	&8099.5	&7649.6	&61170.8	&60646.2\\
disabled	&2	&BS	&200	&199.7	&528.1	&527.7	&2603.8	&2602.7	&183.4	&183.1	&854.9	&854.4	&4112.9	&4111.7\\
	&	&SB	&337.3	&321.2	&846.5	&817.8	&3929.1	&3835.3	&1089.7	&1063.7	&7100.1	&7008.7	&59158.8	&58700.4\\ \hline
\rowstyle{\bfseries}&	&BD	&519.6	&172.3	&1229.9	&398.7	&4616	&1589.2	&968.1	&330.7	&3481	&1210.2	&14600.2	&5087\\
\rowstyle{\bfseries}enabled	&4	&BS	&9.3	&9	&17.5	&17	&69.1	&68.3	&9.4	&9.1	&29.1	&28.6	&85.2	&84.2\\
\rowstyle{\bfseries}	&	&SB	&57.1	&39.9	&91.4	&56.2	&363.6	&252.3	&413.2	&165.5	&1631.3	&639.7	&7333.5	&2857.6\\ \hline
&	&BD	&2248.8	&753.7	&10320.7	&3441.5	&82645.8	&27564.8	&4019.1	&1258.8	&24661.7	&6981	&216133.9	&64735\\
disabled	&4	&BS	&8.8	&8.5	&17.3	&16.9	&64.4	&63.6	&25.7	&25.4	&94.7	&94.2	&349.5	&348.6\\
	&	&SB	&54.3	&38.9	&94.8	&56.9	&345.2	&246.3	&3813.1	&1196.5	&22922.6	&6977.5	&190775.2	&56915.9\\ \hline
\multicolumn{3}{|c|}{Pellet}	&712.4	&712.4	&2708.9	&2708.9	&11440.4	&11440.4	&1230.9	&1230.9	&5340.3	&5340.3	&28981.8	&28981.8\\
  \hline
\end{tabular}}
\label{classificationSummaryCache}
\end{table}

Aiming to study the communication cost among peers, experimental results compare the time that the $E-\mathcal{SHIQ}$ distributed reasoner manages to classify the different versions of the SIS registry, against the time needed by its centralized counterpart, i.e. Pellet. Thus, each peer performs the initialization phase, as already described, the communication phase that is necessary to complete the consistency check of the distributed knowledge base in conjunction to the other peers, and finally, the communication phase for the classification task. The whole task is initiated and executed by each peer concurrently to the others.

Experiments ran on a 8-core MacOS server (Intel(R) Xeon(R) E5620 2.4 GHz) with 6 GB memory, on Darwin 10.8.0 and Sun Java 1.7. Communication of peers in each experiment is performed using TCP/IP sockets on the same machine.  

For each experiment we measure the total time needed by all peers to complete the classification task: This is what is indicated as the ``in sequence'' case in the tables that follow, given that this is the time that the network of peers needs to complete the classification task, when peers run in sequence. In contrast to the  ``in sequence'' time we also show the time that the network of peers needs to complete the classification task, when all peers start the classification task simultaneously: This is the larger time recorded by a peer to complete the whole task, and it is indicated as the ``parallel'' case.

Finally, for each experimental case we record the  time required by the  peers to complete the whole task in case they do not use the projection cache (indicated as \textit{cache disabled} cases) and  in the case they do use the projection cache (indicated as \textit{cache enabled} cases). 

Results are summarized in the table \ref{classificationSummaryCache}. The 1st column of this table indicates whether the projection cache is being enabled or disabled, and the 2nd column shows the number of peers in the network. Each network is of type BD, SB, or BS, as it is indicated in the 3rd column of the table. Finally, each aggregated column, from the  4th to the last one, concerns a specific version of the SIS registry. For each ontology version we indicate both, the time needed by the network of peers to complete the classification task in the "in sequence" and in the parallel cases. 
The times needed by the Pellet to complete the classification task for each version of the SIS registry are shown in the last row of the table. 

\begin{table}
\center
\caption{Classification Time Gain form 2 to 4 Ontology Units}
\resizebox{0.75\textwidth}{!}{
\begin{tabular}{ | c | r | r | r | r | r | r | r | r | r | }
  \hline
	&\multicolumn{3}{|c|}{BD}&\multicolumn{3}{|c|}{BS} &\multicolumn{3}{|c|}{SB}\\ \cline{2-10}
	&2-unit	&4-unit	&\multicolumn{1}{|c|}{gain(\%)}	&2-unit	&4-unit	&\multicolumn{1}{|c|}{gain(\%)} &2-unit	&4-unit	&\multicolumn{1}{|c|}{gain(\%)} \\ \hline
300cn	&505.3	&172.3	&65.89	&210.8	&9.0	&95.72	&320.7	&39.9	&87.55\\ \hline
600cn	&1624.2	&398.7	&75.45	&545.6	&17.0	&96.88	&832.7	&56.2	&93.25\\ \hline
1200cn	&7654.1	&1589.2	&79.24	&2440.4	&68.3	&97.20	&3719	&252.3	&93.22\\ \hline
50c	&526.8	&330.7	&37.22	&170.8	&9.1	&94.66	&307.5	&165.5	&46.18\\ \hline
100c	&2242.8	&1210.2	&46.04	&796.0	&28.6	&96.40	&1289	&639.7	&50.37\\ \hline
200c	&9867.2	&5087.0	&48.45	&3529.7	&84.2	&97.61	&5761.8	&2857.6	&50.40\\
  \hline
\end{tabular}}
\label{Gain}
\end{table}

As results in table \ref{classificationSummaryCache} show, the distributed reasoner is more efficient than Pellet when the projection cache is enabled by the peers. Indeed,  results show that when the projection cache is disabled, the time needed - both in the "in sequence" and in the parallel cases-  is larger than that required by the Pellet reasoner: This shows that the communication cost due to projections is large. Even in the parallel cases, the time required by the network to complete the classification task is larger than the time required by the Pellet to classify the SIS registry. Nevertheless, we must point out that the (positive or negative) differences in the recorded time increase when the number of units increase (i.e. as the distribution of knowledge increases), and especially in cases where long paths and cycles of projections occur due to correspondences (BD cases), or due to correspondences and link relations (SB cases). Specifically, the times recorded by the distributed reasoner in networks of 4 peers, when the cache is enabled (in contrast to the cases where the cache is disabled), are much lower than those recorded by the Pellet, even in cases where cycles of projection-requests exist, and despite the existence of large units. Notably, this is the case even if peers work in sequence.


\begin{table}
\center
\caption{Classification time (secs) for registries with compute node orders}
\resizebox{\textwidth}{!}{
\begin{tabular}{ | bc | ^c | ^r | ^r | ^r | ^r | ^r | ^r | ^r | ^r | ^r | ^r | ^r | ^r | ^r | ^r | ^r | ^r | ^r | ^r |}
  \hline
\multicolumn{2}{|c}{}&\multicolumn{6}{|c|}{300cn}&\multicolumn{6}{|c|}{600cn} &\multicolumn{6}{|c|}{1200cn}\\ \hline 
\multicolumn{1}{|c|}{cache}&\multicolumn{1}{|c|}{unit}&\multicolumn{2}{|c|}{BD} &\multicolumn{2}{|c|}{BS} &\multicolumn{2}{|c|}{SB} &\multicolumn{2}{|c|}{BD} &\multicolumn{2}{|c|}{BS} &\multicolumn{2}{|c|}{SB} &\multicolumn{2}{|c|}{BD} &\multicolumn{2}{|c|}{BS} &\multicolumn{2}{|c|}{SB}\\ \hline
\rowstyle{\bfseries}\multirow{7}{*}{\begin{sideways}enabled\end{sideways}}	&\# 1	&21.9	&151.3	&0.3	&0.1	&16.6	&5.6	&57.7	&348.1	&0.5	&0.1	&32	&11.7	&166.6	&1263.9	&1	&0.3	&100.7	&37.4\\
\rowstyle{\bfseries}	&\# 2	&76.5	&51.9	&2.6	&2.5	&30.8	&18.3	&315	&126.6	&4.4	&4.2	&53.1	&21.8	&1356.5	&517.4	&10.9	&8.5	&226.7	&93.2\\
\rowstyle{\bfseries}	&\# 3	&	&146.3	&	&0.1	&	&5.9	&	&363.6	&	&0.1	&	&11.6	&	&1435.9	&	&0.3	&	&37.2\\
\rowstyle{\bfseries}	&\# 4	&	&149	&	&0.1	&	&5.7	&	&356.6	&	&0.1	&	&11.8	&	&1245.6	&	&0.3	&	&36.8\\ \cline{2-20}
\rowstyle{\bfseries}	&consistency check	&428.7	&21.1	&208.2	&6.5	&289.9	&21.6	&1309.2	&35.1	&541.2	&12.9	&779.6	&34.4	&6297.6	&153.2	&2429.6	&59.8	&3492.3	&159\\
\rowstyle{\bfseries}	&total (in sequence)	&527.2	&519.6	&211.1	&9.3	&337.3	&57.1	&1681.8	&1229.9	&546.1	&17.5	&864.7	&91.4	&7820.7	&4616	&2441.4	&69.1	&3819.7	&363.6\\
\rowstyle{\bfseries}	&total (parallel)	&505.3	&172.3	&210.8	&9	&320.7	&39.9	&1624.2	&398.7	&545.6	&17	&832.7	&56.2	&7654.1	&1589.2	&2440.4	&68.3	&3719	&252.3\\ \hline
\multirow{7}{*}{\begin{sideways}disabled\end{sideways}}	&\# 1	&20	&731.6	&0.3	&0.1	&16.1	&5	&53.7	&3363.6	&0.4	&0.1	&28.7	&12.1	&166.2	&27219.7	&1.1	&0.3	&93.8	&32.6\\
	&\# 2	&76.9	&52.5	&2.5	&2.3	&30.6	&17.7	&290.8	&125.8	&4.2	&3.8	&52.3	&21	&1356	&526.1	&9.4	&8.1	&225.9	&92.1\\
	&\# 3	&	&716.8	&	&0.1	&	&5.2	&	&3406.7	&	&0.1	&	&12.8	&	&27412.8	&	&0.3	&	&33.5\\
	&\# 4	&	&725.8	&	&0.1	&	&5.2	&	&3389.9	&	&0.1	&	&13.1	&	&27335.2	&	&0.3	&	&32.8\\ \cline{2-20}
	&consistency check	&446.3	&22.1	&197.3	&6.2	&290.5	&21.2	&1703.9	&34.7	&523.5	&13.1	&765.5	&35.9	&6267.3	&152	&2593.3	&55.4	&3609.3	&154.2\\
	&total (in sequence)	&543.2	&2248.8	&200	&8.8	&337.3	&54.3	&2048.4	&10320.7	&528.1	&17.3	&846.5	&94.8	&7789.5	&82645.8	&2603.8	&64.4	&3929.1	&345.2\\
	&total (parallel)	&523.2	&753.7	&199.7	&8.5	&321.2	&38.9	&1994.8	&3441.5	&527.7	&16.9	&817.8	&56.9	&7623.3	&27564.8	&2602.7	&63.6	&3835.3	&246.3\\
  \hline
\end{tabular}}
\label{classificationTimeCN}
\end{table}

To further illustrate the difference in the efficacy of the distributed reasoner as the number of units increase from 2 to 4, table \ref{Gain} shows the gain in time required for the classification task. The times concern the parallel cases recorded by the $E-\mathcal{SHIQ}$ distributed reasoner  in any registry version, when the cache is enabled, and for networks of 2 and 4 peers.  Specifically, for each of BD, SB, BS cases we estimate the gain in time by means of the formula $gain=\frac{t_2-t_4}{t_2}*100\%$, where $t_2, t_4$ are the classification times recorded by  networks of 2 and 4 peers, respectively, in any case. This table shows that in cases where correspondences are uni-directional from the smaller to the larger units, the time gain achieved is high in all cases, compared to the time gain achieved when correspondences and link relations cause cyclic projection requests and larger paths of projections to occur, in cases BD and SB. Gor these networks, and in cases where registries are populated by clusters, the time gain achieved is lower compared to the cases where the registries are populated by compute nodes. This is so, due to the large number of projection-requests triggered among units. This is further discussed below.

\begin{table}
\center
\caption{Classification time (secs) for registries with cluster orders}
\resizebox{\textwidth}{!}{
\begin{tabular}{ | bc | ^c | ^r | ^r | ^r | ^r | ^r | ^r | ^r | ^r | ^r | ^r | ^r | ^r | ^r | ^r | ^r | ^r | ^r | ^r |}
  \hline
\multicolumn{2}{|c}{}&\multicolumn{6}{|c|}{50c}&\multicolumn{6}{|c|}{100c} &\multicolumn{6}{|c|}{200c}\\ \hline
\multicolumn{1}{|c|}{cache}&\multicolumn{1}{|c|}{unit}&\multicolumn{2}{|c|}{BD} &\multicolumn{2}{|c|}{BS} &\multicolumn{2}{|c|}{SB} &\multicolumn{2}{|c|}{BD} &\multicolumn{2}{|c|}{BS} &\multicolumn{2}{|c|}{SB} &\multicolumn{2}{|c|}{BD} &\multicolumn{2}{|c|}{BS} &\multicolumn{2}{|c|}{SB}\\ \hline
\rowstyle{\bfseries}\multirow{7}{*}{\begin{sideways}enabled\end{sideways}}	&\# 1	&40.5	&296.7	&0.3	&0.1	&27.2	&119.4	&123.2	&1004.6	&0.6	&0.2	&95.3	&467.2	&544.3	&4355.3	&1.1	&0.3	&441.3	&2134.3\\
\rowstyle{\bfseries}	&\# 2	&142.7	&43.6	&3.6	&2.7	&65.3	&8.6	&533.8	&160.9	&7.7	&5.4	&247.1	&29.8	&2341.8	&681.6	&15.6	&10.8	&1032.1	&116.7\\
\rowstyle{\bfseries}	&\# 3	&	&297	&	&0.1	&	&119.7	&	&1105.3	&	&0.2	&	&494.6	&	&4476.4	&	&0.3	&	&2224.9\\
\rowstyle{\bfseries}	&\# 4	&	&306.8	&	&0.1	&	&139.8	&	&1123	&	&0.2	&	&551.7	&	&4736.2	&	&0.3	&	&2505\\ \cline{2-20}
\rowstyle{\bfseries}	&consistency check	&384.1	&23.9	&167.2	&6.4	&242.2	&25.7	&1708.9	&87.2	&788.3	&23.2	&1041.9	&88	&7525.4	&350.8	&3514.2	&73.4	&4729.7	&352.6\\
\rowstyle{\bfseries}	&total (in sequence)	&567.3	&968.1	&171.1	&9.4	&334.7	&413.2	&2366	&3481	&796.6	&29.1	&1384.4	&1631.3	&10411.5	&14600.2	&3530.9	&85.2	&6203	&7333.5\\
\rowstyle{\bfseries}	&total (parallel)	&526.8	&330.7	&170.8	&9.1	&307.5	&165.5	&2242.8	&1210.2	&796	&28.6	&1289	&639.7	&9867.2	&5087	&3529.7	&84.2	&5761.8	&2857.6\\ \hline
\multirow{7}{*}{\begin{sideways}disabled\end{sideways}}	&\# 1	&39.6	&1192.5	&0.3	&0.1	&26	&1124	&125.8	&7145.2	&0.5	&0.2	&91.5	&6624.5	&524.6	&62924.9	&1.2	&0.3	&458.4	&55560\\
	&\# 2	&887.1	&420.1	&20.6	&18.8	&825.7	&387	&6252.6	&2835.4	&81.7	&70.2	&5928.5	&2546.8	&53031.9	&25094.4	&368	&272.8	&53557.3	&22311.5\\
	&\# 3	&	&1147.8	&	&0.1	&	&1105.5	&	&7220	&	&0.2	&	&6773.8	&	&63379.6	&	&0.3	&	&55987.7\\
	&\# 4	&	&1235.1	&	&0.1	&	&1172.4	&	&7374.9	&	&0.2	&	&6894.8	&	&64094.7	&	&0.3	&	&56572.3\\ \cline{2-20}
	&consistency check	&402.9	&23.7	&162.5	&6.6	&238	&24.2	&1721.1	&86.2	&772.7	&24	&1080.2	&82.6	&7614.3	&640.3	&3743.7	&75.8	&5143.1	&343.6\\
	&total (in sequence)	&1329.6	&4019.1	&183.4	&25.7	&1089.7	&3813.1	&8099.5	&24661.7	&854.9	&94.7	&7100.1	&22922.6	&61170.8	&216133.9	&4112.9	&349.5	&59158.8	&190775.2\\
	&total (parallel)	&1290	&1258.8	&183.1	&25.4	&1063.7	&1196.5	&7649.6	&6981	&854.4	&94.2	&7008.7	&6977.5	&60646.2	&64735	&4111.7	&348.6	&58700.4	&56915.9\\
  \hline
\end{tabular}}
\label{classificationTimeC}
\end{table}

Tables \ref{comm1} and \ref{comm2}  illustrate the projection request messages triggered by units  for each experimental case, depending on whether peers have enabled or disabled the projection cache. Each entry in the table records the total number of projection-requests $triggered$ by each of the peers: Please recall that each peer sends ``own'' projection requests to its neighbors, which trigger further projections to their neighbors, and so on. All these projections triggered by a peer are summed in the corresponding entry for this peer\footnote{It is very difficult to show in a succinct way how projection-requests propagate along paths in the network, thus we aggregate all projections triggered by a peer to the entry for this peer.}. The table shows the projection requests that concern the classification task, only. We observe that for the BD and SB cases, and for the registries populated with clusters, the total number of projection requests triggered by the larger units is very large. This is due to the fact that unit 2, acting as a multiplier of projection requests, and for each package of projection requests that it receives from any of its neighbors, will package  requests triggered due to these projections plus own projection requests. These packaged requests are sent to all neighbors. This is not the case for registries populated with compute nodes, since unit 2 does not hold correspondences that trigger further projections for compute nodes. The increase of individuals in case units are populated with clusters result to a  significant increase in the triggered projections along paths in the network. This significantly larger number of projections explain the decrease in gain observed in table \ref{Gain} for these cases.

\begin{table}
\center
\caption{Projections triggered by unit and its neighbors}
\resizebox{\textwidth}{!}{
\begin{tabular}{ | bc | ^r | ^r | ^r | ^r | ^r | ^r | ^r | ^r | ^r | ^r | ^r | ^r | ^r | ^r | ^r | ^r | ^r | ^r | ^r | }
  \hline
\multirow{2}{*}{\begin{sideways}cache\end{sideways}} & \multirow{2}{*}{unit} 
&\multicolumn{6}{|c|}{300cn}&\multicolumn{6}{|c|}{600cn} &\multicolumn{6}{|c|}{1200cn}\\ \cline{3-20}
& &\multicolumn{2}{|c|}{BD} &\multicolumn{2}{|c|}{BS} &\multicolumn{2}{|c|}{SB} &\multicolumn{2}{|c|}{BD} &\multicolumn{2}{|c|}{BS} &\multicolumn{2}{|c|}{SB} &\multicolumn{2}{|c|}{BD} &\multicolumn{2}{|c|}{BS} &\multicolumn{2}{|c|}{SB}\\ \cline{3-20}
& &2-unit	&4-unit	&2-unit	&4-unit	&2-unit	&4-unit	&2-unit	&4-unit	&2-unit	&4-unit	&2-unit	&4-unit	&2-unit	&4-unit	&2-unit	&4-unit	&2-unit	&4-unit\\ \hline
\rowstyle{\bfseries}\multirow{4}{*}{\begin{sideways}enabled\end{sideways}}	&\# 1	&2341	&890	&0	&0	&2341	&762	&4481	&1643	&0	&0	&4481	&1519	&8909	&3099	&0	&0	&8909	&2975\\
\rowstyle{\bfseries}	&\# 2	&7	&21	&7	&18	&2	&6	&7	&18	&7	&21	&2	&6	&7	&21	&7	&21	&2	&6\\
\rowstyle{\bfseries}	&\# 3	&	&932	&	&0	&	&808	&	&1592	&	&0	&	&1456	&	&3133	&	&0	&	&2977\\
\rowstyle{\bfseries}	&\# 4	&	&917	&	&0	&	&793	&	&1652	&	&0	&	&1528	&	&3103	&	&0	&	&2979\\ \hline
\multirow{4}{*}{\begin{sideways}disabled\end{sideways}}	&\# 1	&2343	&2090	&0	&0	&2343	&764	&4483	&4163	&0	&0	&4483	&1521	&8911	&8131	&0	&0	&8911	&2977\\
	&\# 2	&7	&21	&7	&21	&2	&6	&7	&21	&7	&21	&2	&6	&7	&21	&7	&21	&2	&6\\
	&\# 3	&	&2226	&	&0	&	&810	&	&3972	&	&0	&	&1458	&	&8133	&	&0	&	&2979\\
	&\# 4	&	&2185	&	&0	&	&795	&	&4186	&	&0	&	&1530	&	&8141	&	&0	&	&2981\\
  \hline
\end{tabular}}
\label{comm1}
\end{table}

Delving more into the behavior of individual peers, tables \ref{classificationTimeCN} and \ref{classificationTimeC} record the  time required by each of the peers to complete its part in the overall classification task. The tables distinguish between registries, depending on the types of individuals populating each registry: Compute nodes (table 8) or clusters (table 9). Each entry per peer shows the time required by the  peer for the whole task, while the time required by the peers to check the consistency of their distributed knowledge base is recorded in the rows labeled ``consistency check''. As it is shown, registries populated with clusters, due to the larger complexity of specifications and the number of projection requests triggered, require greater time in all tasks. Nevertheless, in both cases, the results show the following phenomena: (a) Having the projection cache enabled, peers require less time compared to the time required by Pellet and of course the time required when the cache is disabled. (b) The positive and negative differences in the recorded time are much greater in all cases with 4 peers (compared to the time required by Pellet), even for the cases where  cycles and longer paths of projection-requests may occur (i.e in the BD and SB cases).



\begin{table}
\center
\caption{Projections triggered by unit and its neighbors}
\resizebox{\textwidth}{!}{
\begin{tabular}{ | bc | ^r | ^r | ^r | ^r | ^r | ^r | ^r | ^r | ^r | ^r | ^r | ^r | ^r | ^r | ^r | ^r | ^r | ^r | ^r | }
  \hline
\multirow{2}{*}{\begin{sideways}cache\end{sideways}} & \multirow{2}{*}{unit} 
&\multicolumn{6}{|c|}{50c}&\multicolumn{6}{|c|}{100c} &\multicolumn{6}{|c|}{200c}\\ \cline{3-20}
& &\multicolumn{2}{|c|}{BD} &\multicolumn{2}{|c|}{BS} &\multicolumn{2}{|c|}{SB} &\multicolumn{2}{|c|}{BD} &\multicolumn{2}{|c|}{BS} &\multicolumn{2}{|c|}{SB} &\multicolumn{2}{|c|}{BD} &\multicolumn{2}{|c|}{BS} &\multicolumn{2}{|c|}{SB}\\ \cline{3-20}
& &2-unit	&4-unit	&2-unit	&4-unit	&2-unit	&4-unit	&2-unit	&4-unit	&2-unit	&4-unit	&2-unit	&4-unit	&2-unit	&4-unit	&2-unit	&4-unit	&2-unit	&4-unit\\ \hline
\rowstyle{\bfseries}\multirow{4}{*}{\begin{sideways}enabled\end{sideways}}	&\# 1	&2787	&4627	&0	&0	&2787	&902	&5418	&8674	&0	&0	&5418	&1820	&10581	&18085	&0	&0	&10581	&3561\\
\rowstyle{\bfseries}	&\# 2	&19	&21	&19	&21	&14	&6	&19	&21	&19	&21	&14	&6	&19	&21	&19	&21	&14	&6\\
\rowstyle{\bfseries}	&\# 3	&	&4709	&	&0	&	&985	&	&9083	&	&0	&	&1831	&	&18092	&	&0	&	&3556\\
\rowstyle{\bfseries}	&\# 4	&	&4645	&	&0	&	&921	&	&9440	&	&0	&	&1788	&	&17219	&	&0	&	&3486\\ \hline
\multirow{4}{*}{\begin{sideways}disabled\end{sideways}}	&\# 1	&2789	&29269	&0	&0	&2789	&27705	&5420	&103203	&0	&0	&5420	&100031	&10583	&394549	&0	&0	&10583	&388363\\
	&\# 2	&192	&321	&192	&321	&187	&306	&361	&615	&361	&615	&356	&600	&715	&1221	&715	&1221	&710	&1206\\
	&\# 3	&	&29911	&	&0	&	&28187	&	&104021	&	&0	&	&100833	&	&398524	&	&0	&	&392358\\
	&\# 4	&	&29719	&	&0	&	&28123	&	&103104	&	&0	&	&99998	&	&397516	&	&0	&	&391488\\
  \hline
\end{tabular}}
\label{comm2}
\end{table}

\subsection{Concluding remarks on experimental results}

Overall the concluding remarks on the experimental results are as follows: 

For the cases of BS networks, the correspondences hold from the point of view of the smaller unit (i.e. unit 2). The time recorded for the larger units in these cases is very small. Obviously, the reason is that the large units have no correspondences or link relations that may trigger further projections, thus large units compute their local concept taxonomy during the initialization phase and end their task when they have served the few projection requests from unit 2. This is indeed the case, given that no messages are transmitted from any of the peers exploiting units 1, 3, 4 to the peer for unit 2 (as it is also shown in table \ref{comm2}). On the other hand, during the classification there is a small delay in peer 2, due to the projection messages it sends to its neighbors. However these projection messages are  evaluated almost instantly by the neighboring peers.

For the cases of SB networks, the correspondences hold from the point of view of larger units (i.e. unit 1, 3, and 4). For the classification task of SIS registries with compute node orders, we must notice that   peer 2 needs more time than its neighbors. The reason is that although it has no correspondences towards its neighbors, it still depends on them because of the link relations. Therefore, peer 2 transmits projection-request messages to its neighbors in this setting, as well. In cases with 4 units and cluster orders, we have many projection requests triggered between units. This fact affects the overall performance, since peer 2 will reply to a peer, only when it receives the projection responses from other peers. The delay is significantly reduced when the projection cache is enabled. 

For the case of BD networks, correspondences hold for the viewpoint of both, small and large units. This means that we have more correspondences in the BD cases, compared to the BS and SB cases. This affects the classification time in these cases, which is higher than the time recorded to any of the corresponding BS and SB cases. More correspondences, result to more projection-requests that are triggered during the reasoning tasks. This also affects the time spent for the consistency check of the distributed knowledge base, which is also higher than the corresponding time in the BS and SB cases. 

Finally, when the cache of projections is disabled, the classification time recorded for the  BD and SB cases is always higher than the  time recorded by Pellet. This happens due to the triggering of projection request messages along paths in the network, resulting to ``avalanches'' of projection request messages. 
This fact is a strong evidence that the cache of projections is necessary to preserve scalability.

\section{Concluding Remarks and Future Work}

Aiming to provide a rich representation framework for combining and 
reasoning with distinct ontologies in open, heterogeneous and inherently distributed settings, this article proposes the  $E-\mathcal{SHIQ}$ representation framework. In these settings we can expect that different ontologies may be combined in many 
different, subtle ways, mostly by means of correspondences and domain-specific relations between concepts and individuals, while peers retain subjective beliefs on how their knowledge is coupled with that of others. Our aim is to support peers to reason jointly with knowledge distributed in their ontologies, by combining local reasoning chunks. 

Peers - each responsible for an ontology unit - combining their knowledge with the $E-\mathcal{SHIQ}$ framework, participate in a collaborative distributed tableau algorithm towards constructing a distributed completion graph: Doing so, each peer possesses a part of the overall graph, and graphs in different peers are combined by means of projections of graph nodes. No peer possesses the overall graph, and the collaboration between peers is realized via the maintenance of corresponding nodes' labels  from one chunk of the completion graph to another, via projection requests and projection responses. 
Emphasis has been given to preserving the subjectivity and locality of units' knowledge: This enables a peer-to-peer distributed reasoning process. The article shows that $E-\mathcal{SHIQ}$ inherently supports subsumption propagation through concept-to-concept correspondences.

The proposed $E-\mathcal{SHIQ}$ tableau algorithm can be used to extend any tableau reasoner. The implementation presented in this article concerns extending Pellet and includes a variety of optimization methods, to further enhance the performance of the reasoning process. One such optimization, is that units group their projection-request messages into a single packaged request that is sent to a neighbor peer. Actually, for the $E-\mathcal{SHIQ}$ distributed reasoner presented, the communication phase follows the initialization phase (as these are defined in section 6.1), allowing the gathering of all possible projection-requests from any peer to its neighbors. Further experimentation is necessary towards interleaving the construction of the local completion graphs and communicating projection-requests and projection-responses under different conditions: For instance, we may expect that sending a small number of projection-requests may decrease the waiting time of peers (i.e. the time waiting for responses). However, this may not be the case when requests are sent to peers with large units, who have not constructed their local completion graph. 

In the current implementation of the distributed reasoner, for each projection-request message received by a peer, this peer replicates its local completion graph, in order to process the requested projections w.r.t. to own (local and subjective) knowledge and the subjective knowledge of peers that triggered the received requests. The preservation of the local completion graph reduces the cost of processing the projected nodes, because each unit does not need to reconstruct  the local completion graph from scratch. However, the replication of the completion graph for each request received from any neighbor, may result to a significant amount of required resources per peer. To achieve a good balance between the processing time and the resources required, further research is necessary: This may be handled by a task manager in each peer which will further organize the requests and the available resources, for better performance. Also, we need to recall that replications of completion graphs are not preserved (i.e. each replica is destructed once the projection requests are served). 

The implemented distributed reasoner also employs a projection cache per peer, reducing the duplication of projection-requests sent from one peer to another. Experimental results have shown that this caching functionality is necessary to increase the efficiency of the distributed reasoner:  Further research is necessary towards more sophisticated caching techniques (e.g. via abstraction methods), as well as towards more  advanced matchmaking techniques between projection-request messages and cached projections.

Although the experimental results have shown that the distributed reasoning process proposed is more efficient than a centralized reasoning process (i.e. one that processes a semantically equivalent monolithic ontology), they have also uncovered some limitations of the implemented reasoner. Specifically, we have seen that making the distributed reasoning process as much parallel as possible, is not that easy: There will always be the case where some peer will have to wait for the results of projection requests it has made to other peers. 
As it is well known, and it is also shown by our experiments, to achieve a higher degree of parallelism and reduce the waiting time of peers, knowledge has to be distributed to ontology units evenly: In this case the processing load is also distributed to the peers evenly. Towards achieving this goal in settings with autonomous peers, where each peer evolves its knowledge independently from the others, self-organization mechanisms are necessary. This however may contradict two basic requirements: (a) knowledge in a unit is local and is not available to other units, (b) peers are autonomous and hold subjective knowledge concerning how their knowledge is combined with the knowledge of others.
However a self-organization approach may also be necessary towards reducing the length of paths in any network of peers. This will reduce the number of projection-requests triggered along any path in the network of peers.

An important line of research towards making $E-\mathcal{SHIQ}$  usable in real-world settings were large ontologies exist, concerns ontology modularization. Specifically, we need partitioning algorithms that will construct  $E-\mathcal{SHIQ}$ knowledge bases, towards enabling distributed reasoners to process this knowledge efficently. Further research concerns constructing such a modularization algorithms, preserving properties such as locality of knowledge, even distribution of knowledge between modules, and further investigating issues concerning connectivity of modules towards making the reasoning process as much efficient as possible.

\acks{

This research project is being supported by the project ``IRAKLITOS II'' of the O.P.E.L.L. 2007 - 2013 of the NSRF (2007 - 2013), co-funded by the European Union and National Resources of Greece.
}


\vskip 0.2in
\bibliography{references}
\bibliographystyle{theapa}

\end{document}